\documentclass[10pt,twocolumn,letterpaper]{article}

\usepackage[pagenumbers]{cvpr} 

\newcommand{\best}[1]{\textbf{#1}}

\usepackage{multirow}

\newcommand{\videoInput}{\mathbf{I}}
\newcommand{\matmap}{\mathbf{mat}}
\newcommand{\diffusionModel}{\mathbf{f}}
\newcommand{\diffusionModelParams}{\theta}
\newcommand{\diffusionModelFn}{\diffusionModel_{\diffusionModelParams}}
\newcommand{\vaeEncoder}{\mathcal{E}}
\newcommand{\vaeDecoder}{\mathcal{D}}
\newcommand{\typeEmb}{\mathbf{c}_{\text{prompt}}}
\newcommand{\dataDistribution}{p_{\text{data}}}
\newcommand{\diffusionNoise}{\mathbb{\epsilon}}
\newcommand{\vaepbr}{\mathrm{VAE}_\mathrm{pbr}}

\definecolor{myorange}{rgb}{1, 0.85, 0.7}
\definecolor{myred}{rgb}{1, 0.7, 0.7}

\newcommand{\reducedstrut}{\vrule width 0pt height 1.05\ht\strutbox depth 1.0\dp\strutbox\relax}
\newcommand{\sota}[1]{
  \begingroup
  \setlength{\fboxsep}{0pt}
  \colorbox{myred}{\reducedstrut#1\/}
  \endgroup
}
\newcommand{\subsota}[1]{
  \begingroup
  \setlength{\fboxsep}{0pt}
  \colorbox{myorange}{\reducedstrut#1\/}
  \endgroup
}

\definecolor{cvprblue}{rgb}{0.21,0.49,0.74}
\usepackage[pagebackref,breaklinks,colorlinks,allcolors=cvprblue]{hyperref}
\usepackage{xcolor}

\def\paperID{} 
\def\confName{CVPR}
\def\confYear{2026}

\title{VideoMatGen: PBR Materials through Joint Generative Modeling}

\author{Jon Hasselgren\\
NVIDIA
\and
Zheng Zeng\\
NVIDIA
\and
Milo\v{s} Ha\v{s}an \\
NVIDIA
\and 
Jacob Munkberg \\
NVIDIA\\
}

\begin{document}

\newcommand{\figTeaser}{
\begin{figure*}
    \centering
    \includegraphics[width=\textwidth]{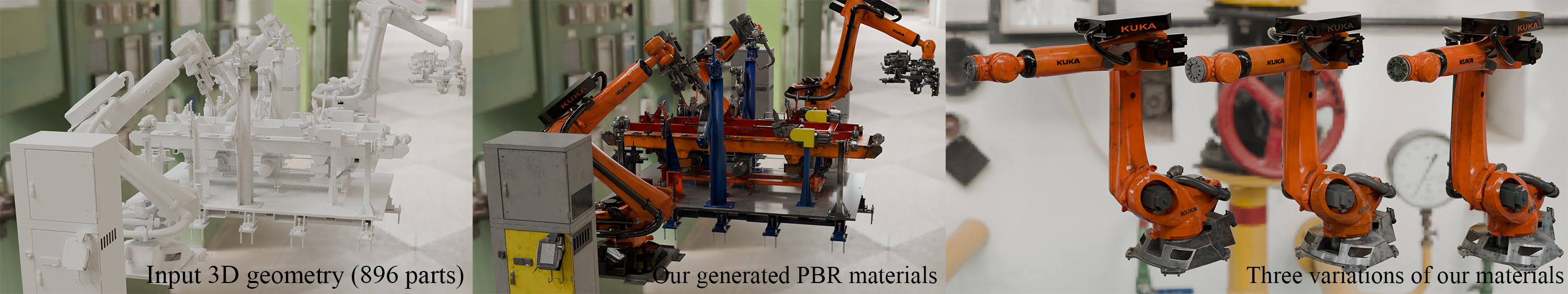}
    \caption{Given 3D models and text prompts, we generate unique high quality PBR materials for each 
    3D part using a finetuned video diffusion model. Our generated materials are directly applicable in 
    any content creation applications. Here we show a Physical AI training application, applying the generated materials to a virtual factory setting. On the right, we show three variations of generated materials (from the same detailed text prompts and different random seeds) for an industrial robot asset with 19 parts.}
    \label{fig:teaser}
\end{figure*}
}


\newcommand{\figSystem}{
\begin{figure*}
    \centering
    \includegraphics[width=0.99\textwidth]{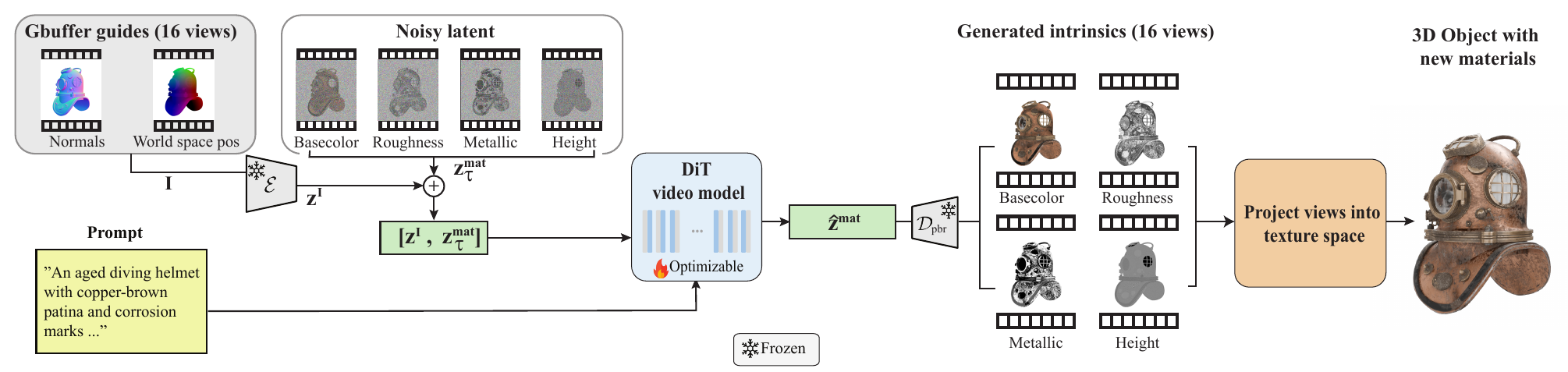}
    \vspace*{-2mm}
    \caption{Our method starts from a known 3D model and a text prompt. We first render videos of normal maps and 
    world space positions. Next, these conditions are encoded into latent space, using a pretrained encoder, $\mathcal{E}$, to produce latent conditions, $\textbf{z}^{\videoInput}$. These are concatenated with noisy latents, $\textbf{z}_\tau^{\matmap}$, representing material modalities, along the channel dimension.
    The latents and text prompt are then passed to our finetuned video model, which generates 
    a denoised latent, $\hat{\textbf{z}}^{\matmap}$. The denoised latent is decoded into 
    videos of the intrinsic material channels: base color, roughness, metallicity, and height,
    using a custom VAE decoder $\mathcal{D}_{\mathrm{pbr}}$ which decodes all material properties jointly.
    Finally, we project the generated views into texture space to extract high quality, standard PBR materials.}
    \label{fig:system}
\end{figure*}
}


\newcommand{\figSeed}{
\begin{figure}
    \centering
    \setlength{\tabcolsep}{1pt}
    \begin{tabular}{ccc}
       \includegraphics[width=0.32\columnwidth]{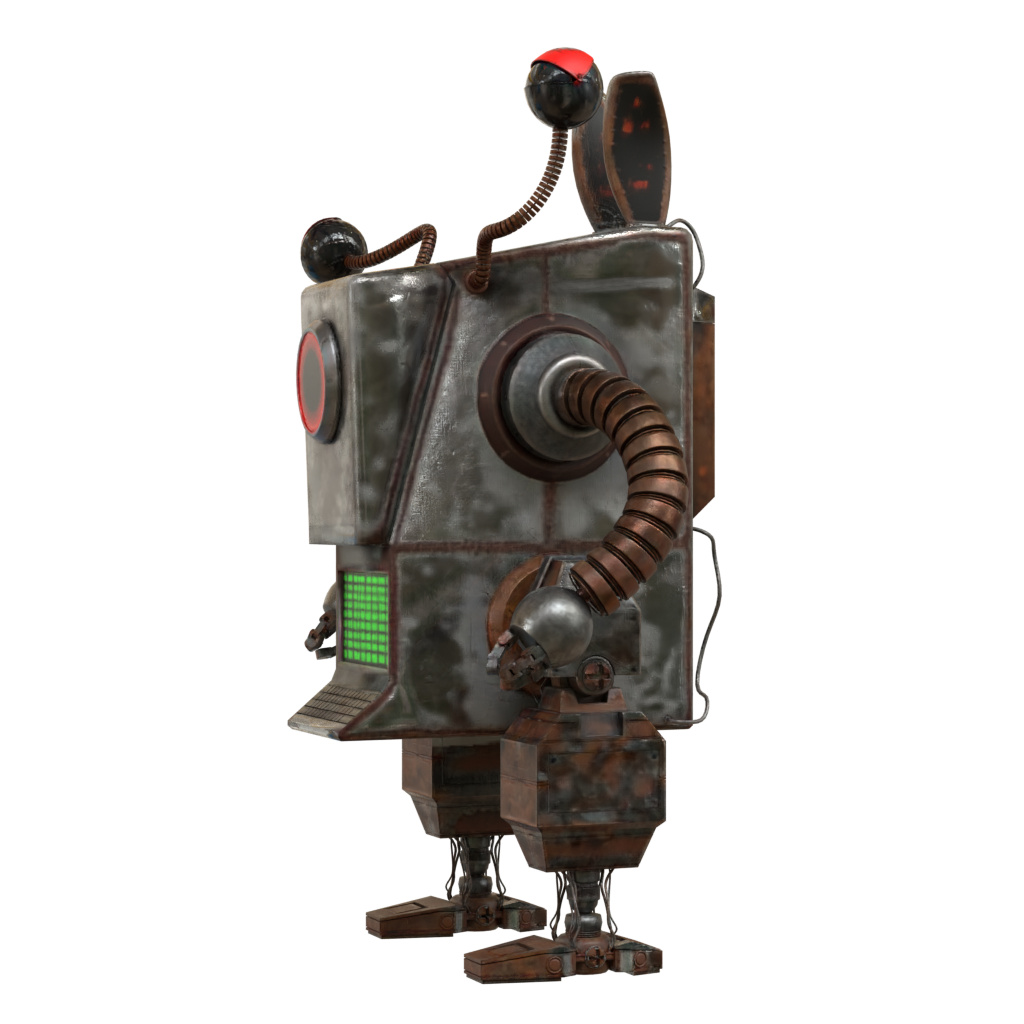} &
       \includegraphics[width=0.32\columnwidth]{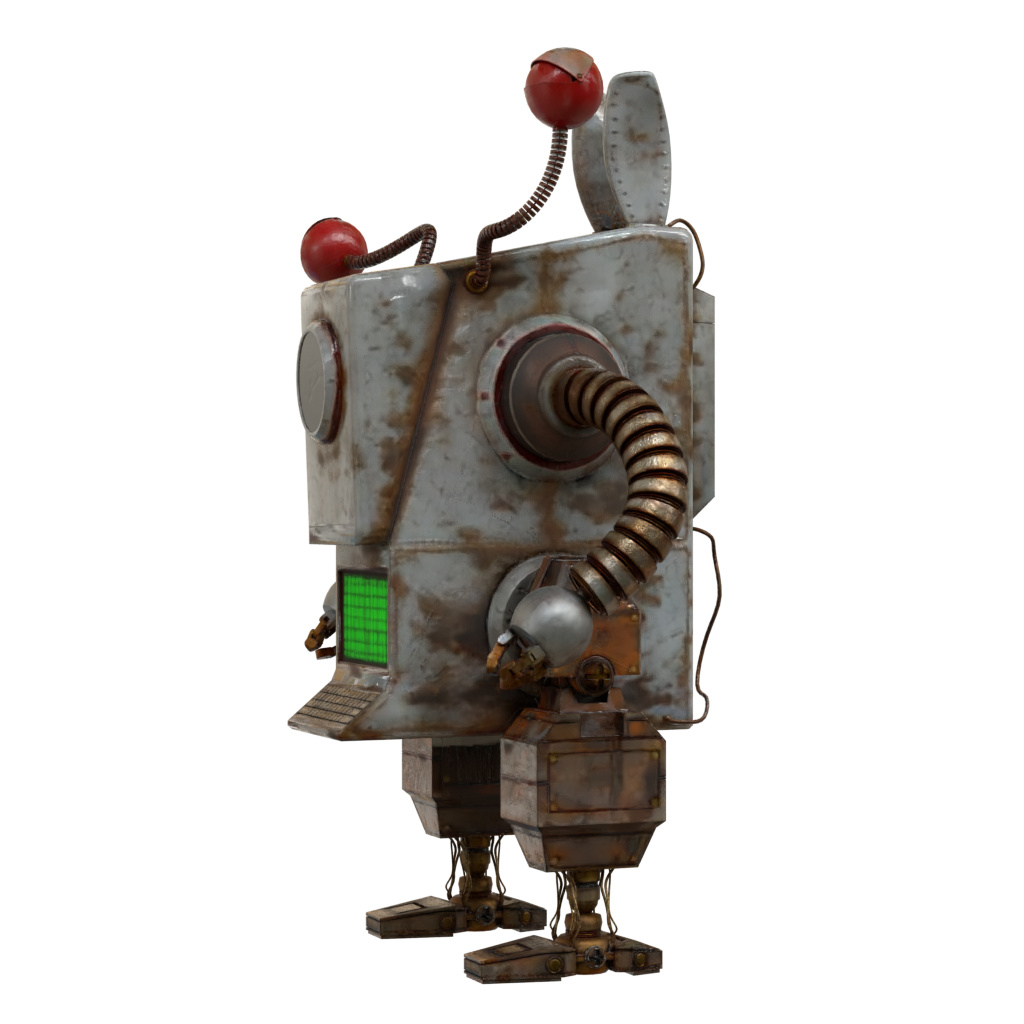} &
       \includegraphics[width=0.32\columnwidth]{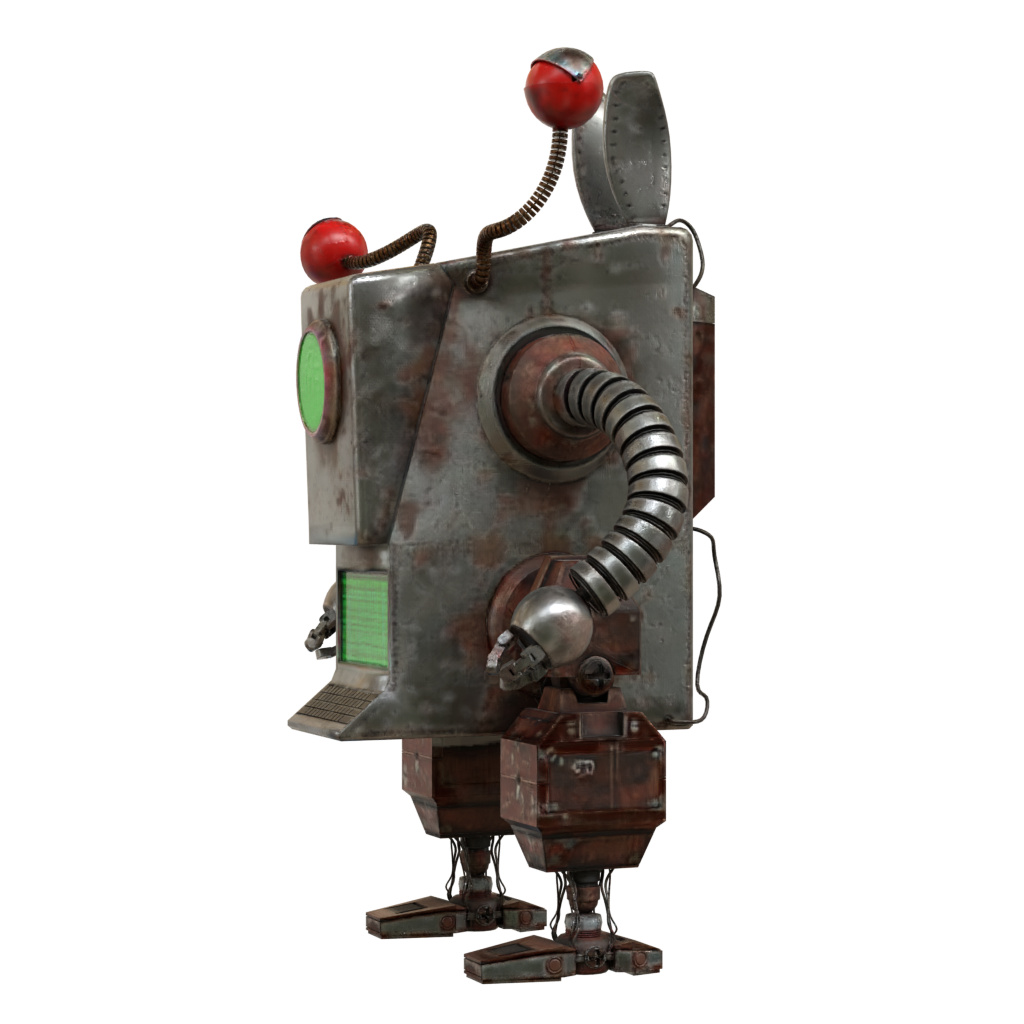} \\
       \includegraphics[width=0.32\columnwidth]{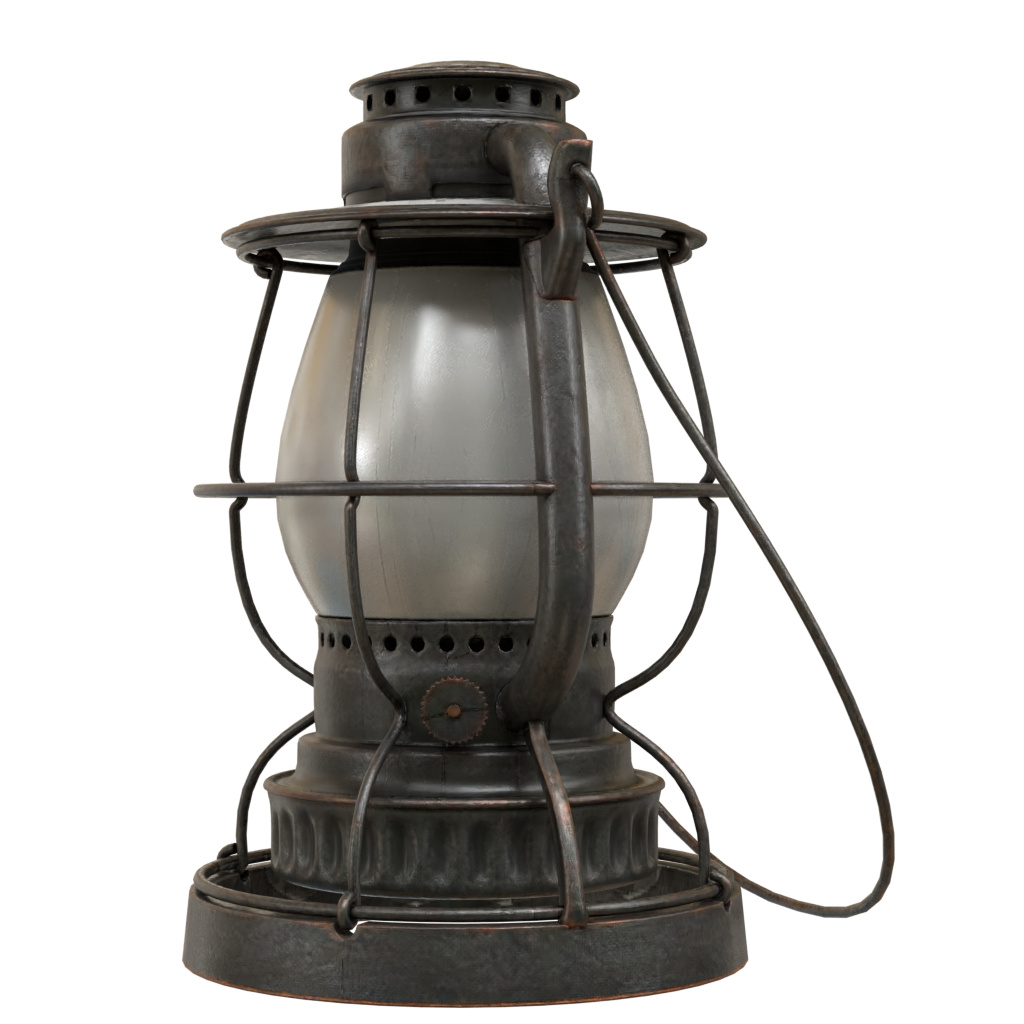} &
       \includegraphics[width=0.32\columnwidth]{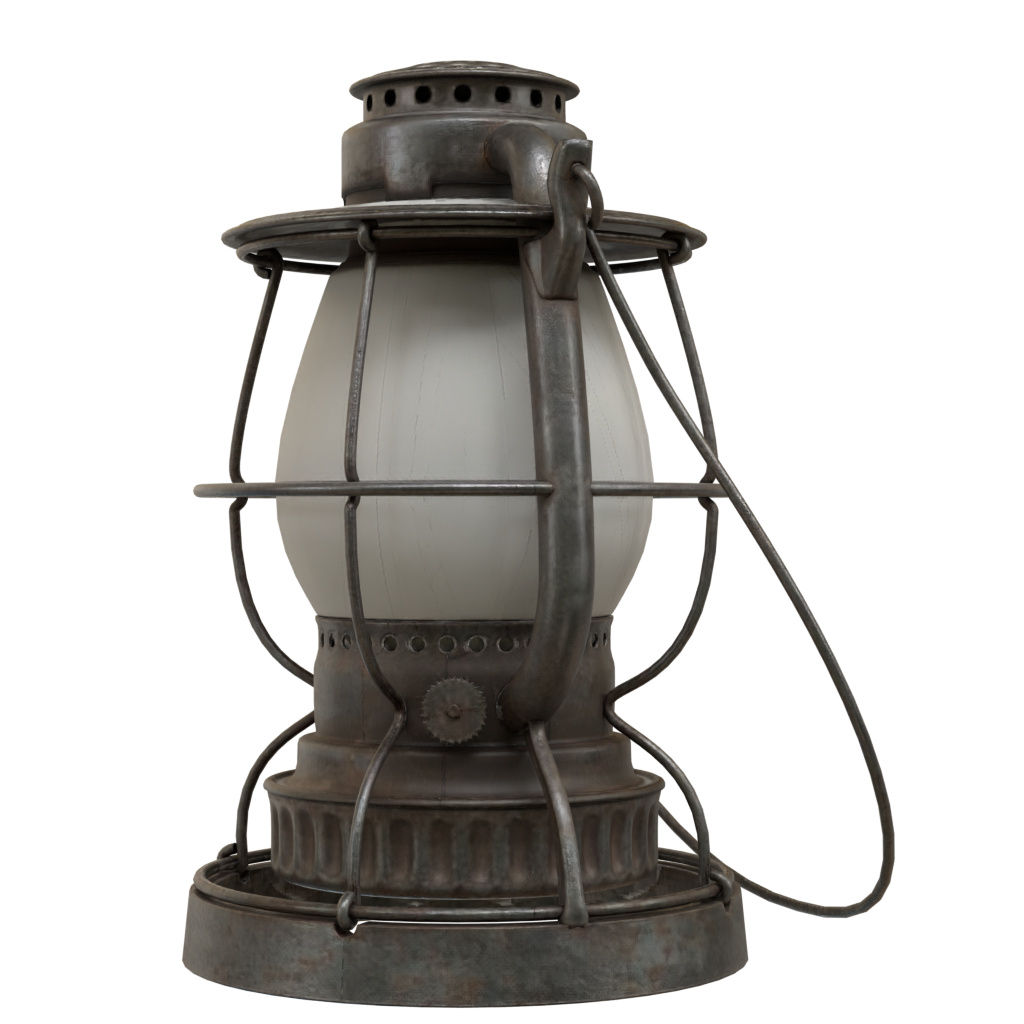} &
       \includegraphics[width=0.32\columnwidth]{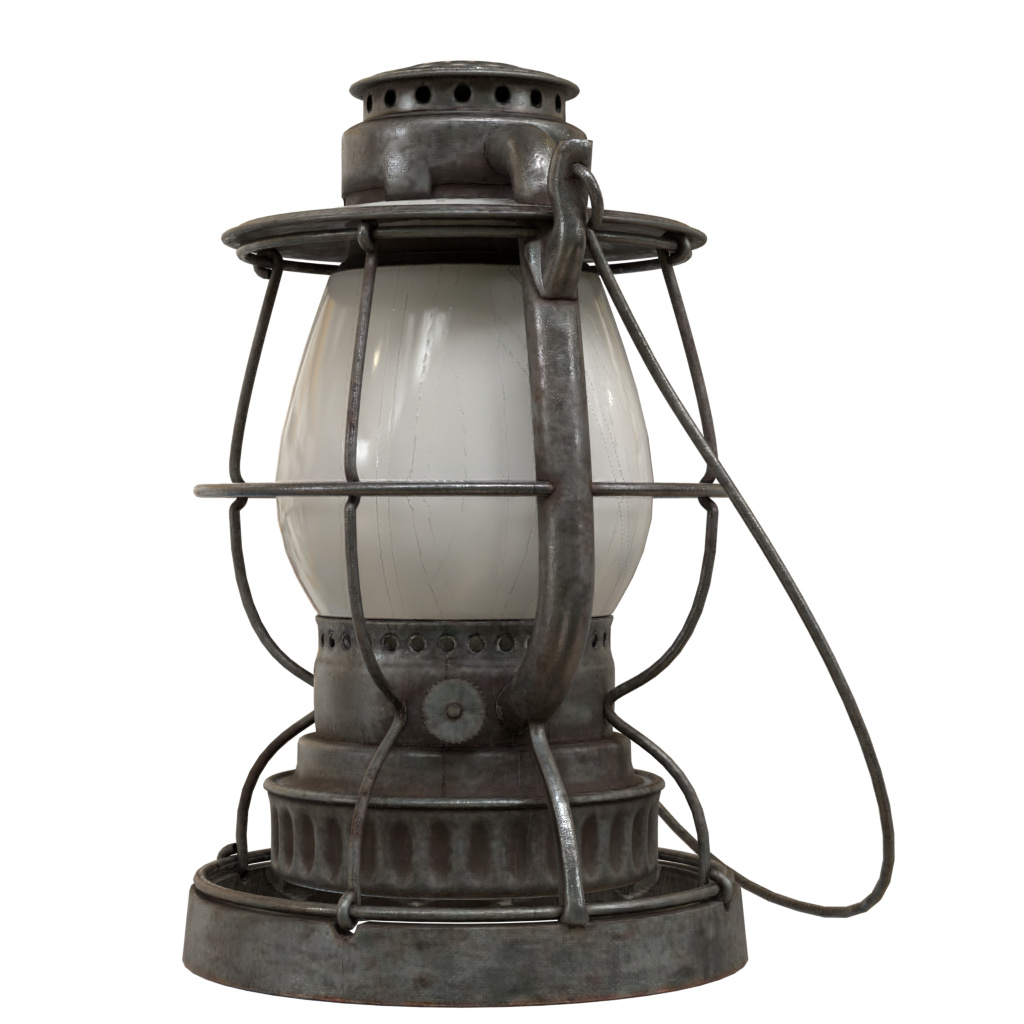} \\
    \end{tabular}
    \caption{
        We generate three materials from the same text prompt (see supplemental), each with a unique random seed. 
        This results in subtle variations of materials for the two examples. 
    }
    \label{fig:seed}
 \end{figure}
}


\newcommand{\figRelighting}{
\begin{figure}
    \footnotesize
    \centering
    \setlength{\tabcolsep}{1pt}
    \begin{tabular}{cccc}
        \rotatebox[origin=c]{90}{Hunyuan(I)} &
        \raisebox{-0.5\height}{\includegraphics[width=0.3\columnwidth]{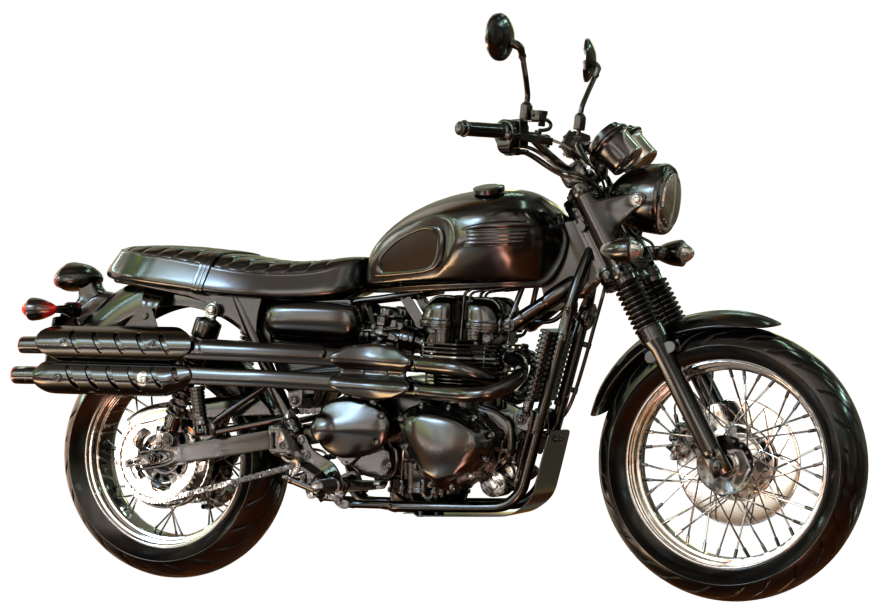}} &
        \raisebox{-0.5\height}{\includegraphics[width=0.3\columnwidth]{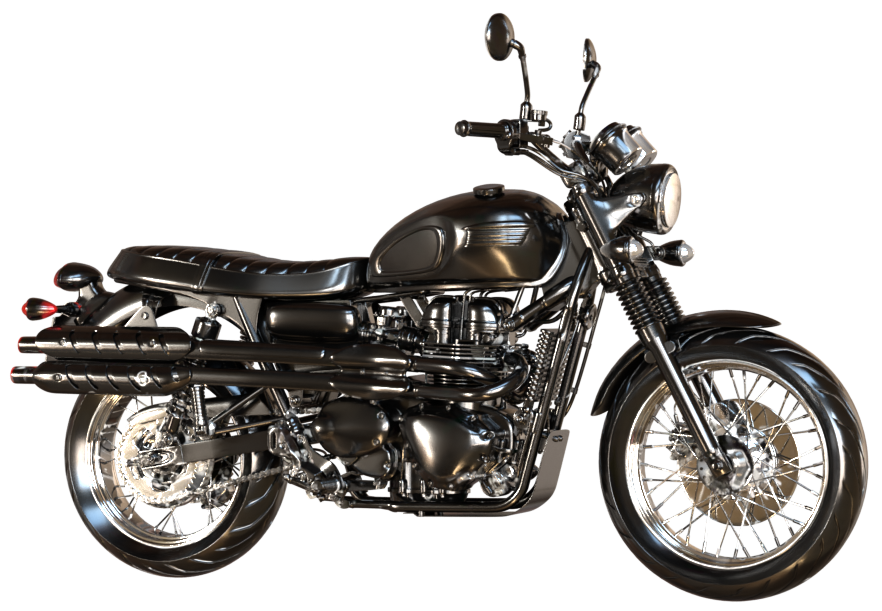}} &
        \raisebox{-0.5\height}{\includegraphics[width=0.3\columnwidth]{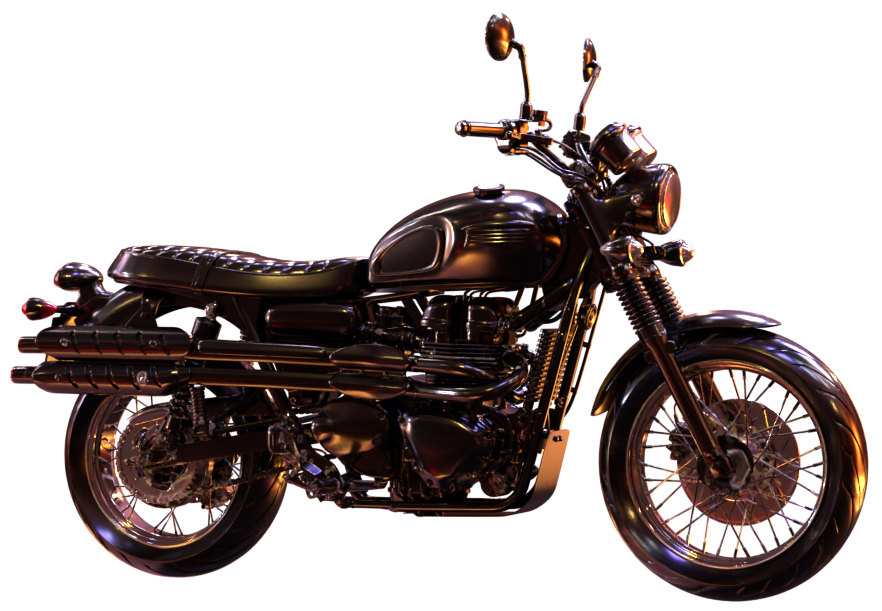}} \\
        \rotatebox[origin=c]{90}{Videomat} &
        \raisebox{-0.5\height}{\includegraphics[width=0.3\columnwidth]{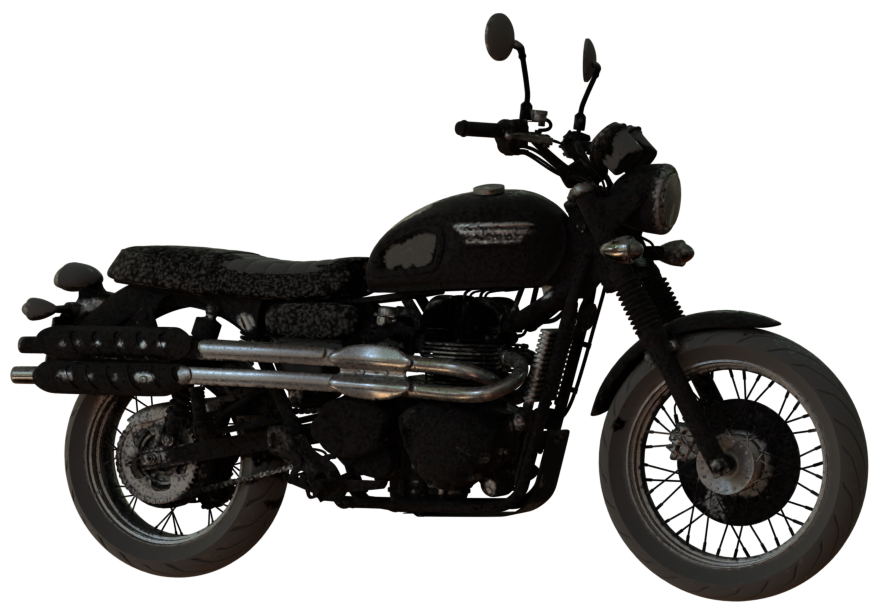}} &
        \raisebox{-0.5\height}{\includegraphics[width=0.3\columnwidth]{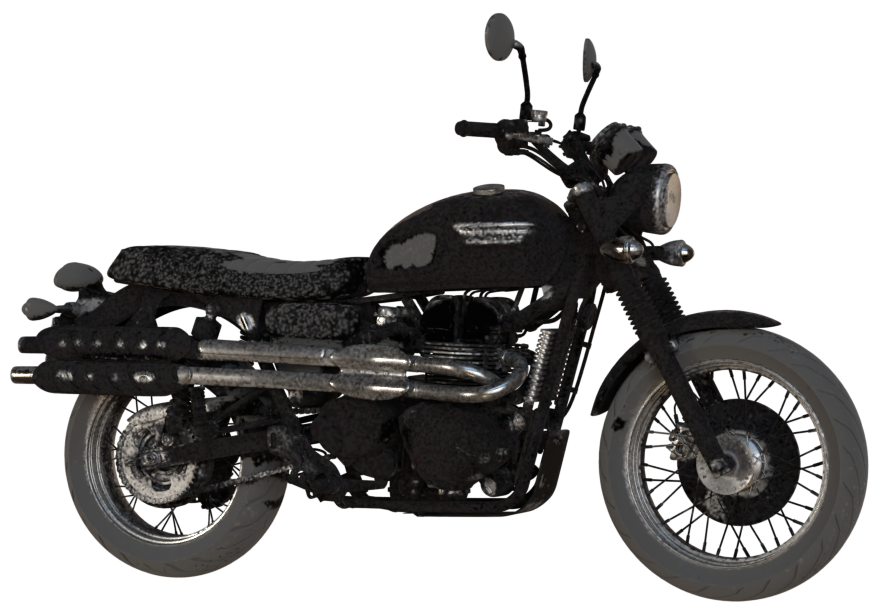}} &
        \raisebox{-0.5\height}{\includegraphics[width=0.3\columnwidth]{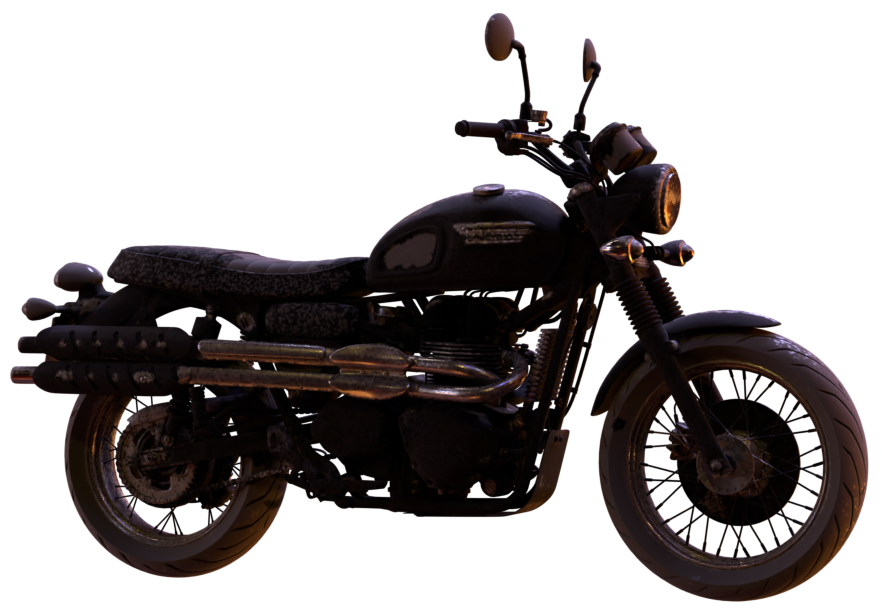}} \\
        \rotatebox[origin=c]{90}{Our} &
        \raisebox{-0.5\height}{\includegraphics[width=0.3\columnwidth]{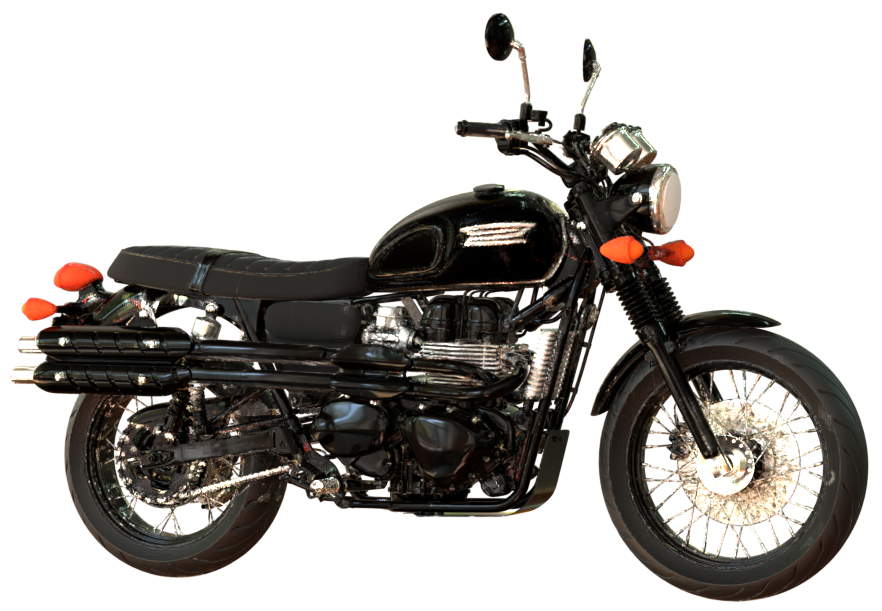}} &
        \raisebox{-0.5\height}{\includegraphics[width=0.3\columnwidth]{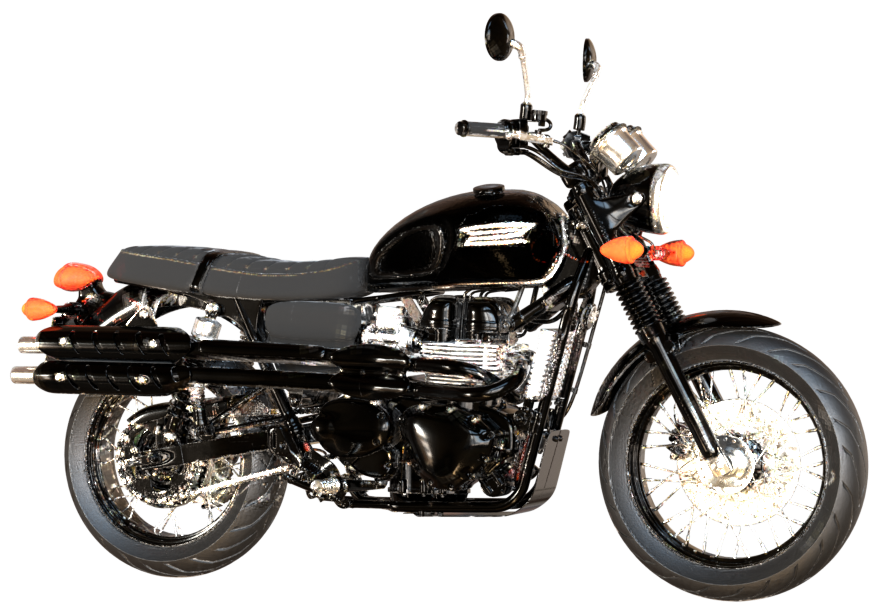}} &
        \raisebox{-0.5\height}{\includegraphics[width=0.3\columnwidth]{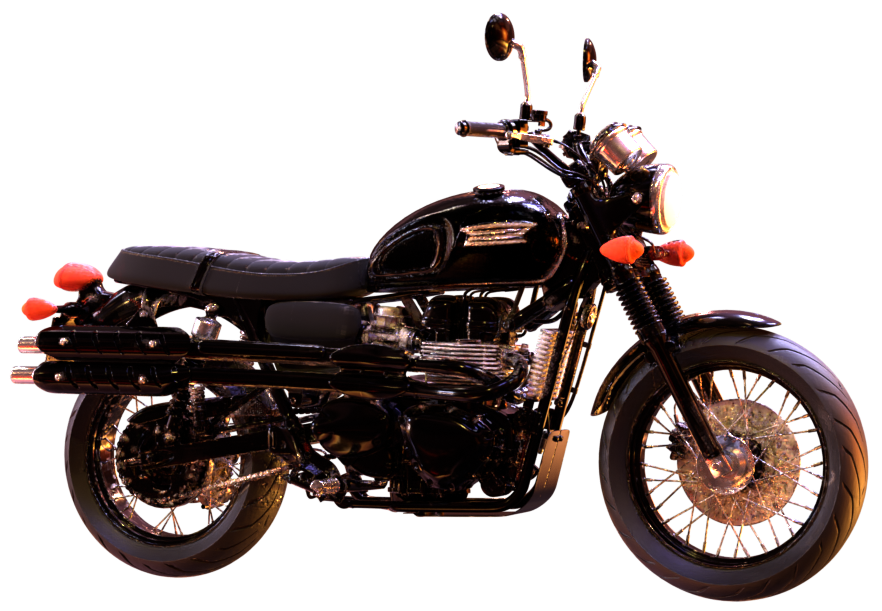}} \\
        &
        \includegraphics[width=0.1\columnwidth]{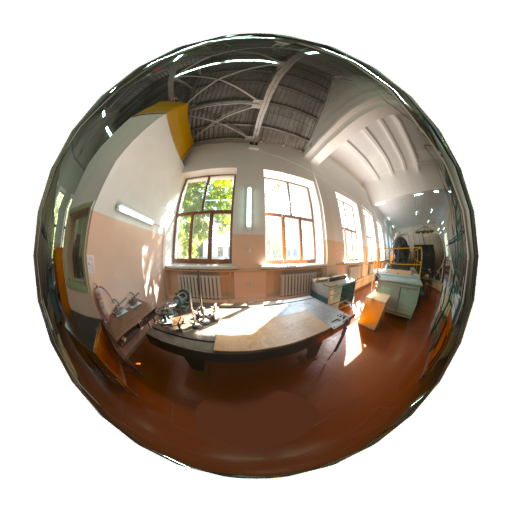} &
        \includegraphics[width=0.1\columnwidth]{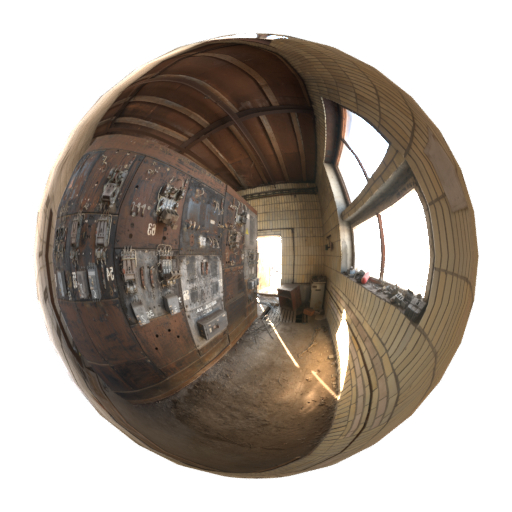} &
        \includegraphics[width=0.1\columnwidth]{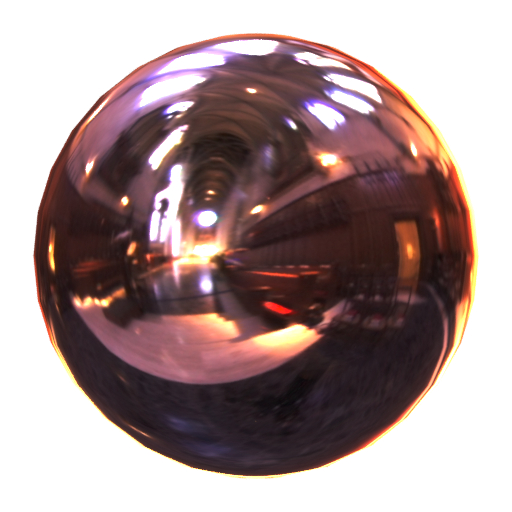} \\
        & Aerodynamics workshop & Distribution board & Grace cathedral
    \end{tabular}
    \caption{
        We show relit results, using three HDR probes~\cite{polyhaven}, of the generated materials for Hunyuan3D-Paint (image-guided), VideoMat, and our method (both text-guided). Our generated materials produce convincing details in three different lighting scenarios.
    }
    \label{fig:relighting}
 \end{figure}
}


\newcommand{\figMainQualityResults}{
\begin{figure*}
    \centering
    \small
    \setlength{\tabcolsep}{1pt}
    \begin{tabular}{cccccccccccc}
        & Geometry & & Relit & Base color & Roughness & Metallicity & & Relit & Base color & Roughness & Metallicity \\
        \multirow{2}{*}{\rotatebox{90}{\makebox[0.16\textwidth]{\centering \textsc{Diver}}}} &
        \multirow{2}{*}{\includegraphics[width=0.16\textwidth]{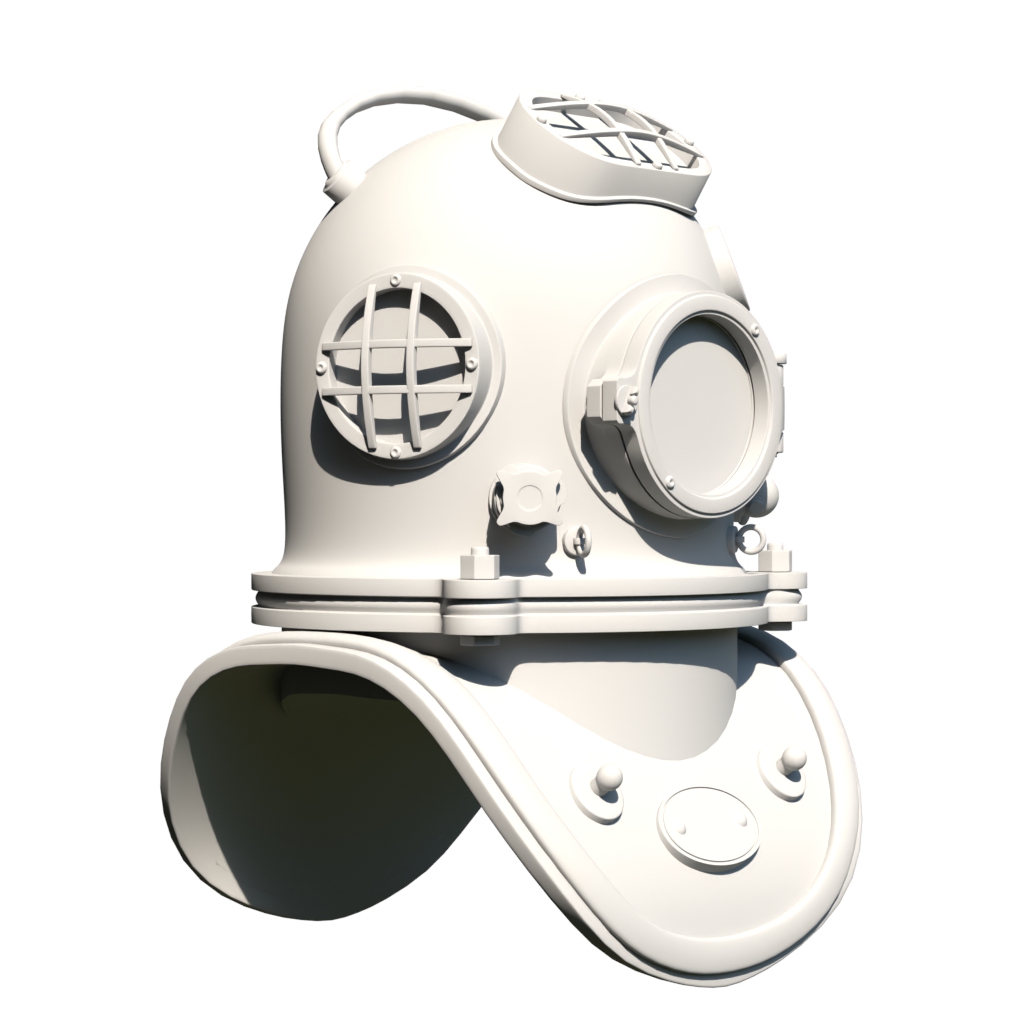}} & 
        \rotatebox[origin=c]{90}{Hunyuan(I)} &
        \raisebox{-0.5\height}{\includegraphics[width=0.093\textwidth]{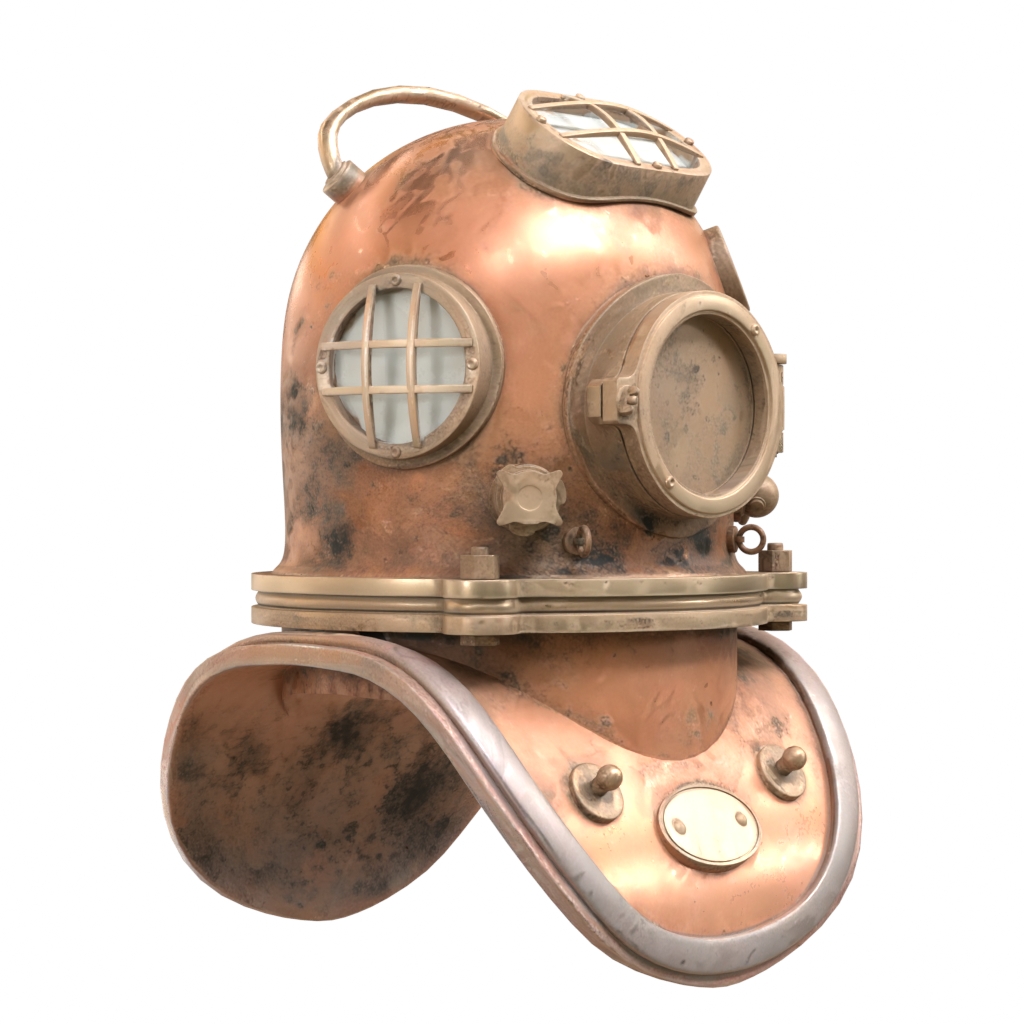}} &
        \raisebox{-0.5\height}{\includegraphics[width=0.093\textwidth]{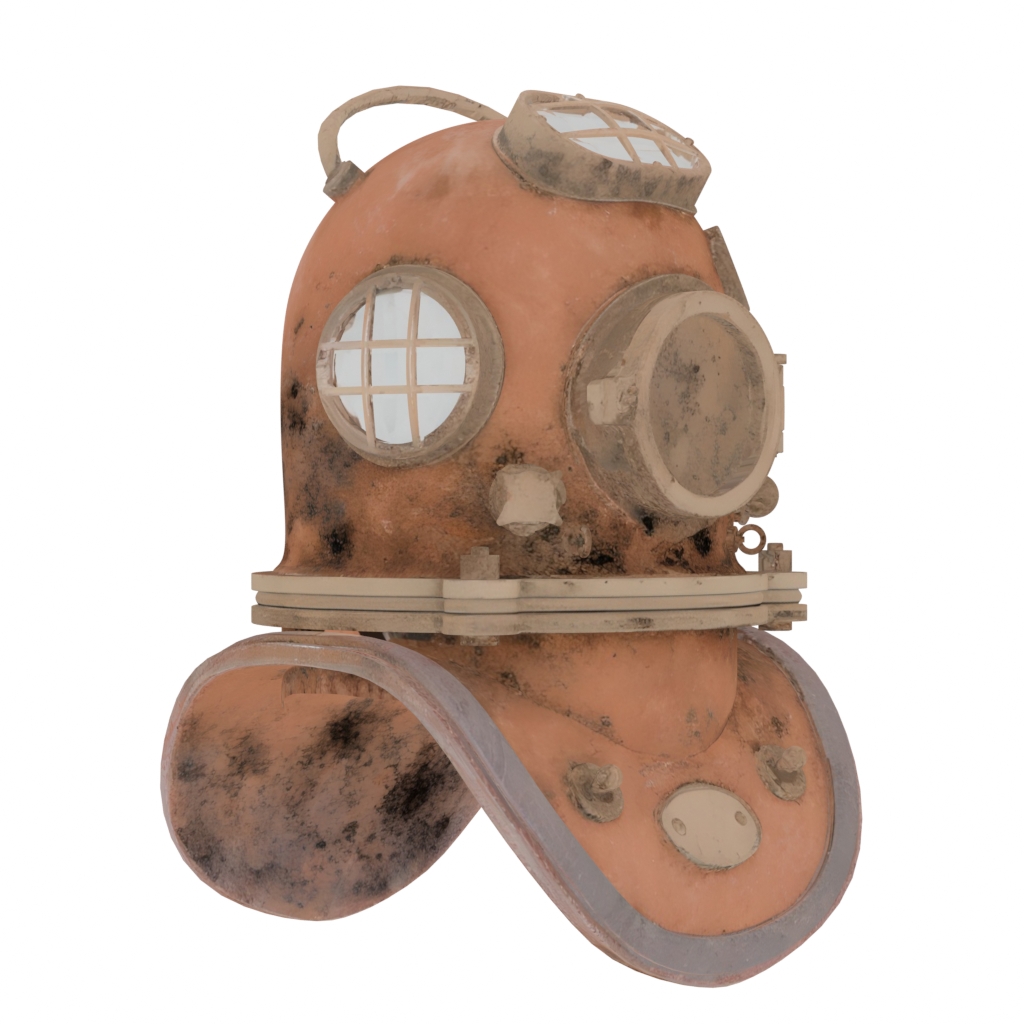}} &
        \raisebox{-0.5\height}{\includegraphics[width=0.093\textwidth]{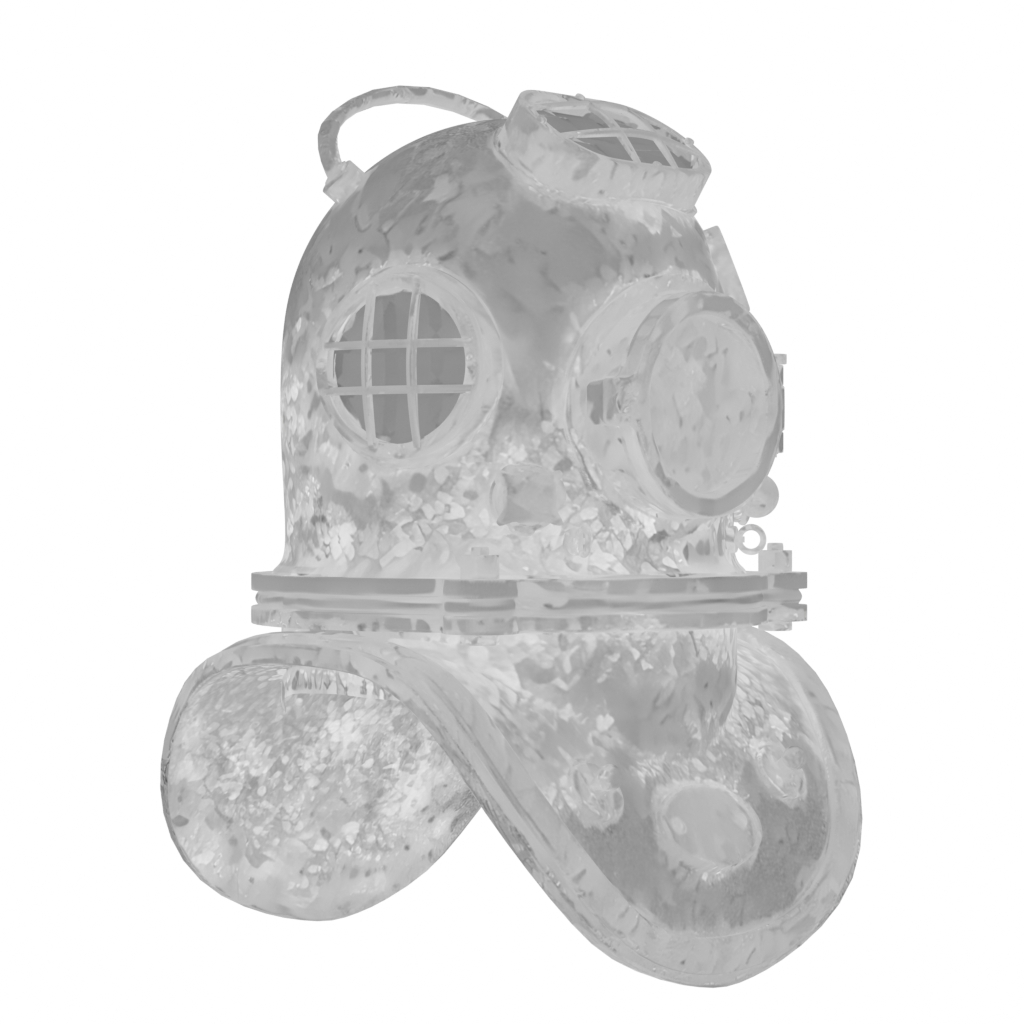}} &
        \raisebox{-0.5\height}{\includegraphics[width=0.093\textwidth]{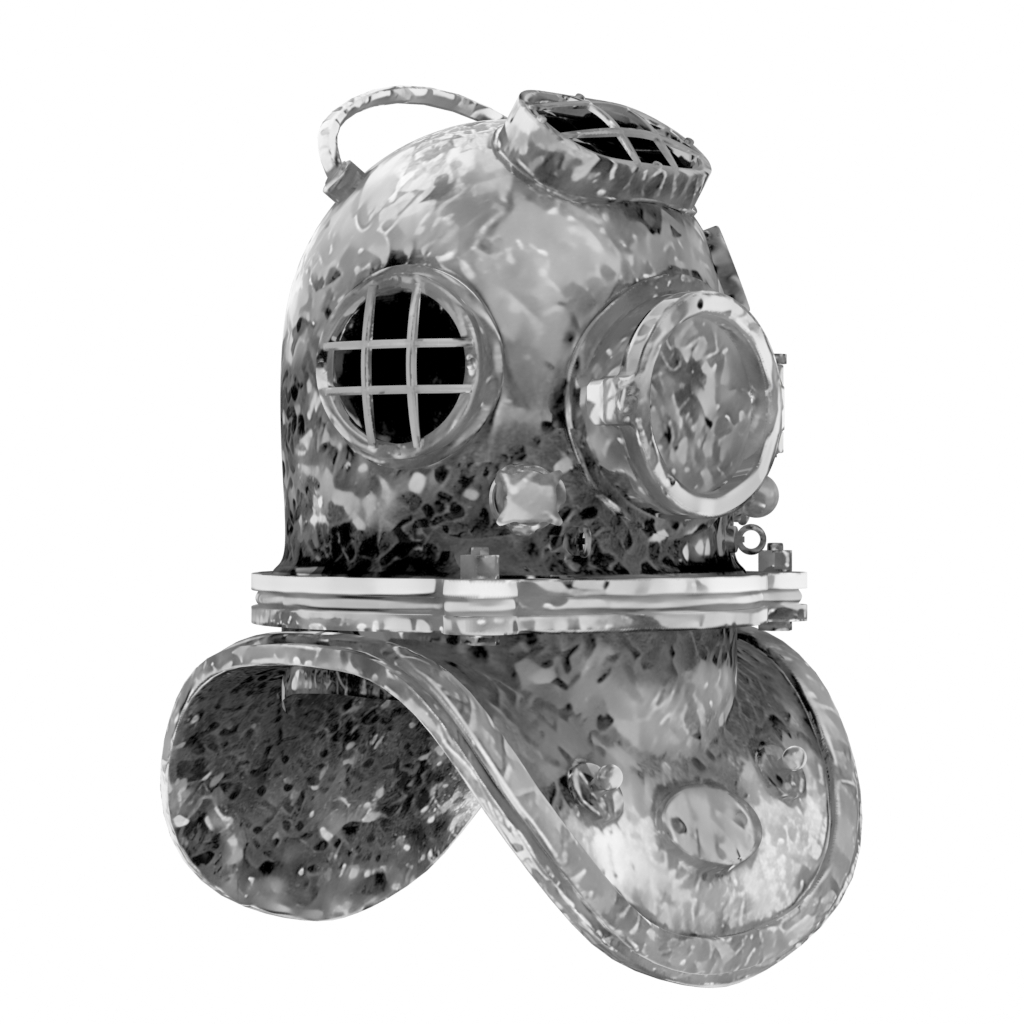}} &
        \rotatebox[origin=c]{90}{Hunyuan(T)} &
        \raisebox{-0.5\height}{\includegraphics[width=0.093\textwidth]{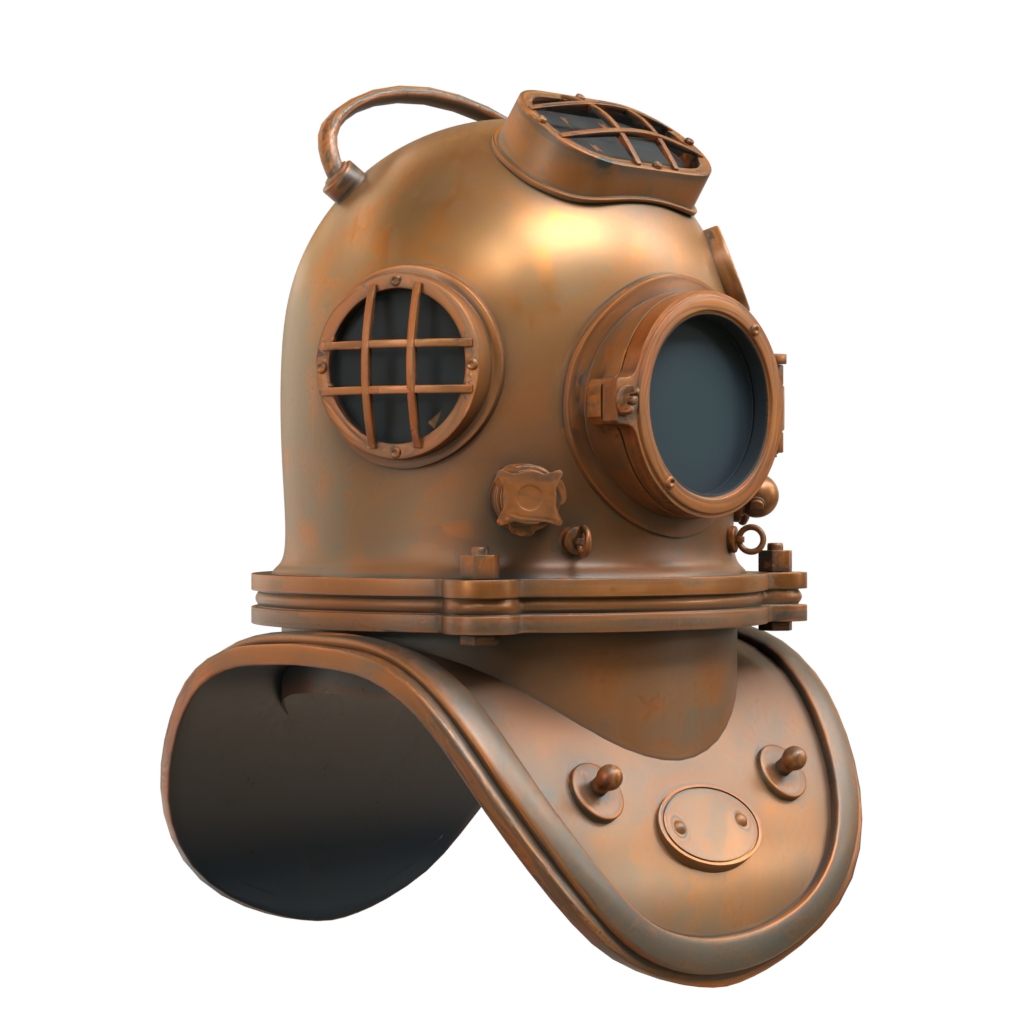}} &
        \raisebox{-0.5\height}{\includegraphics[width=0.093\textwidth]{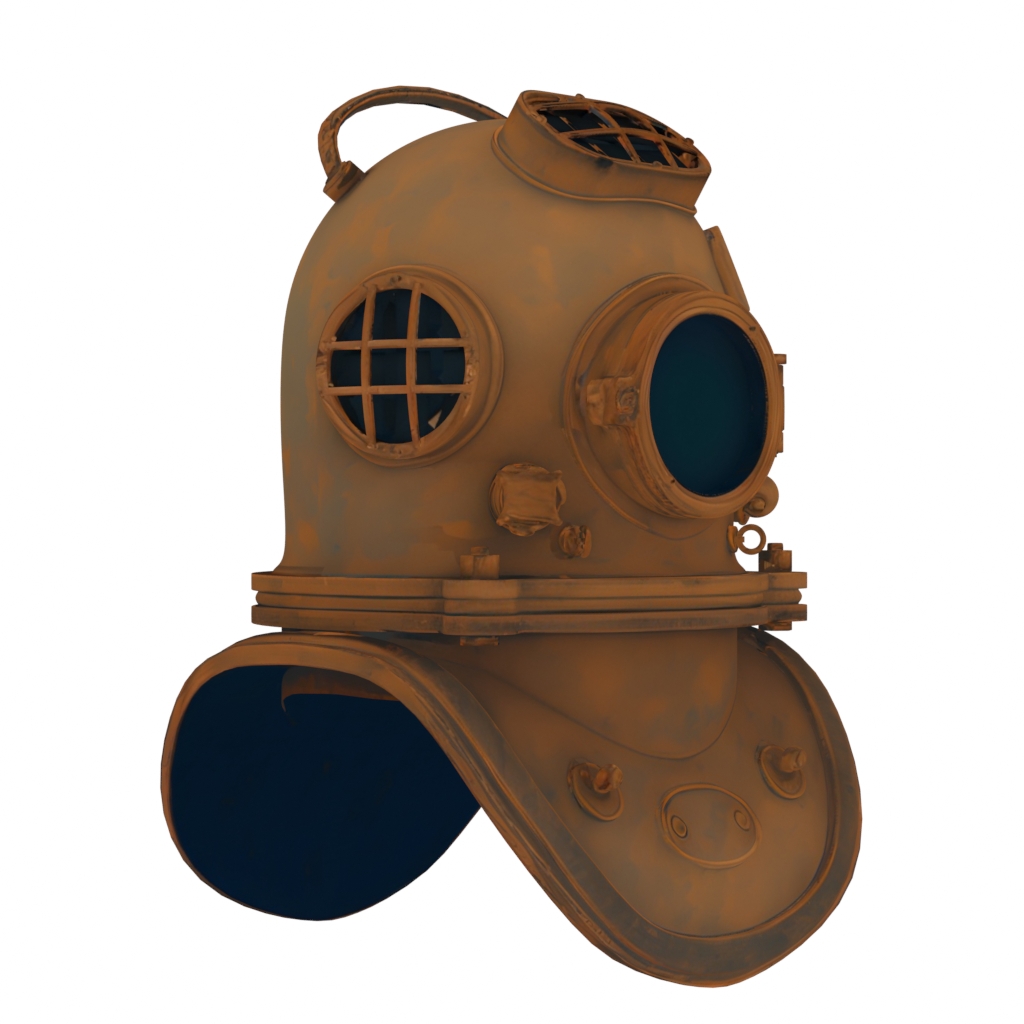}} &
        \raisebox{-0.5\height}{\includegraphics[width=0.093\textwidth]{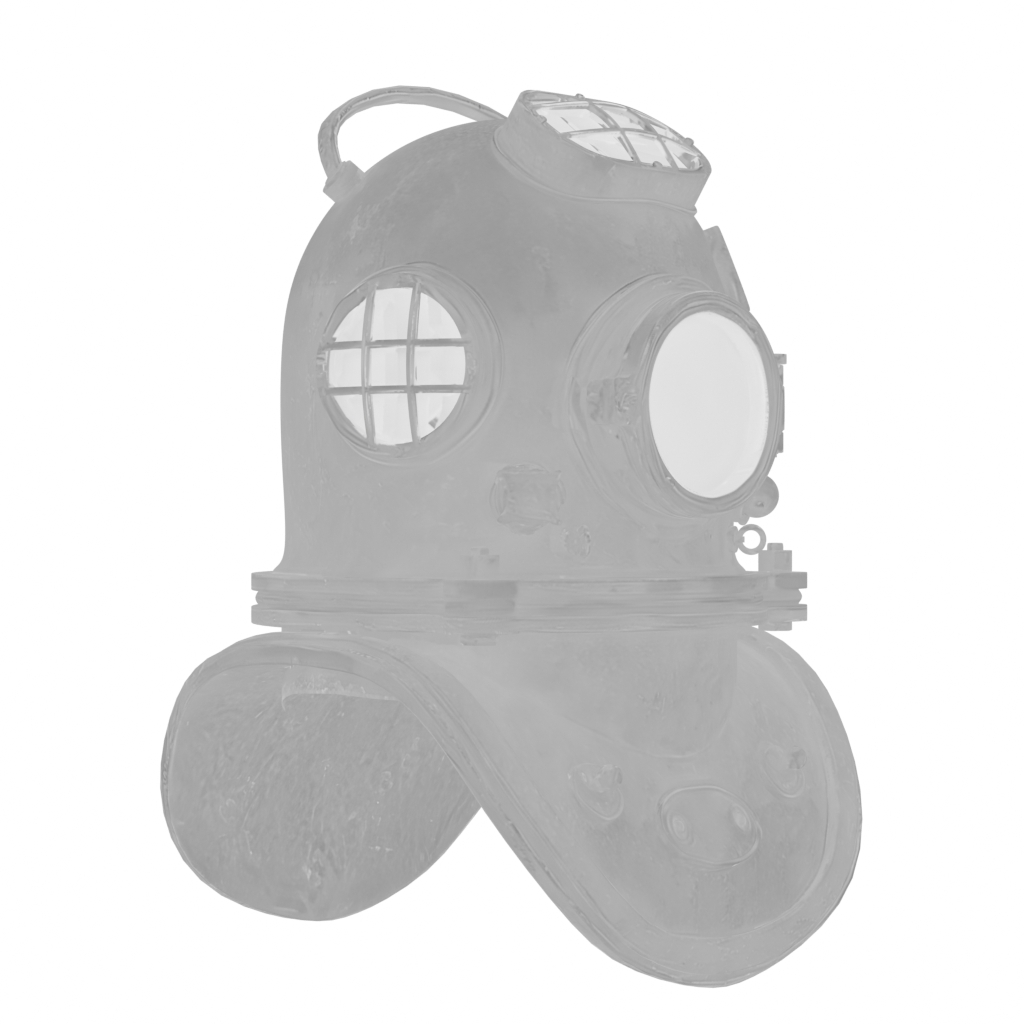}} &
        \raisebox{-0.5\height}{\includegraphics[width=0.093\textwidth]{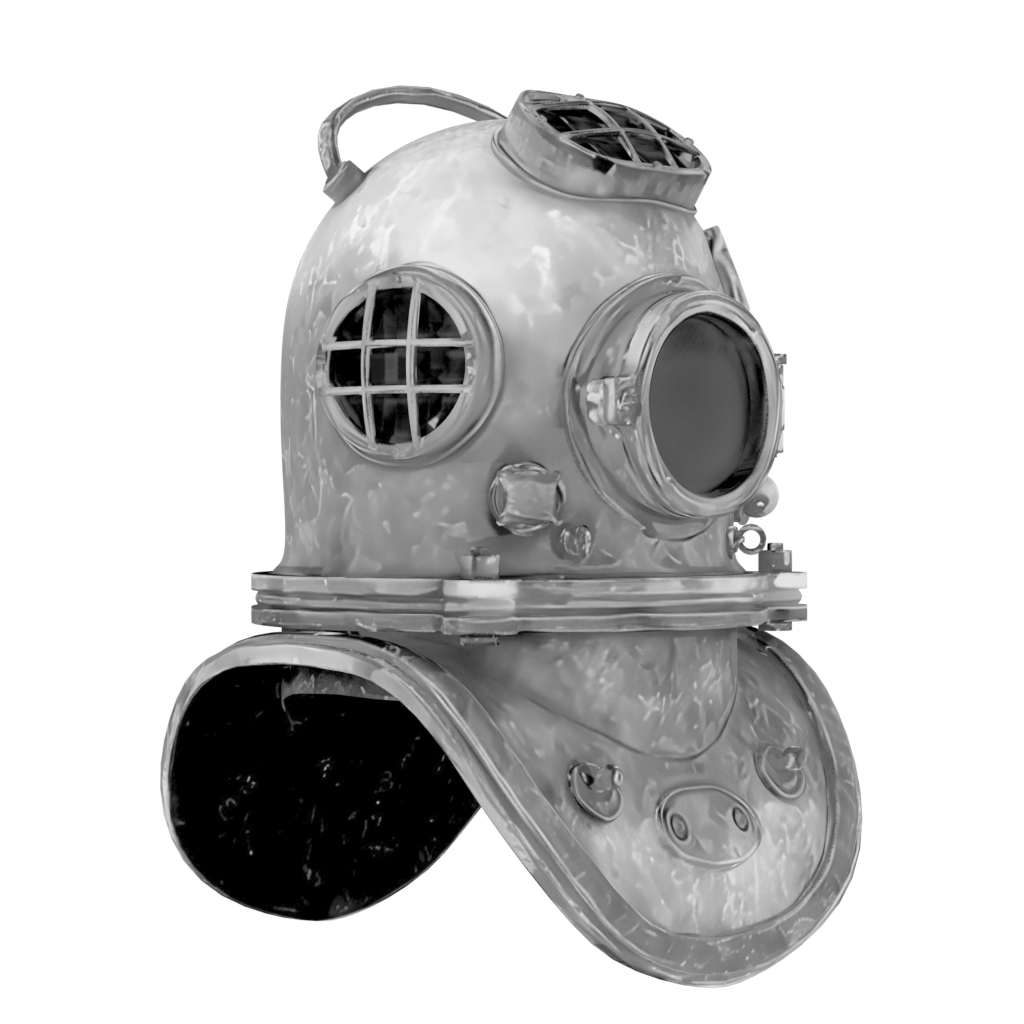}} \\
        & & 
        \rotatebox[origin=c]{90}{VideoMat (T)} &
        \raisebox{-0.5\height}{\includegraphics[width=0.093\textwidth]{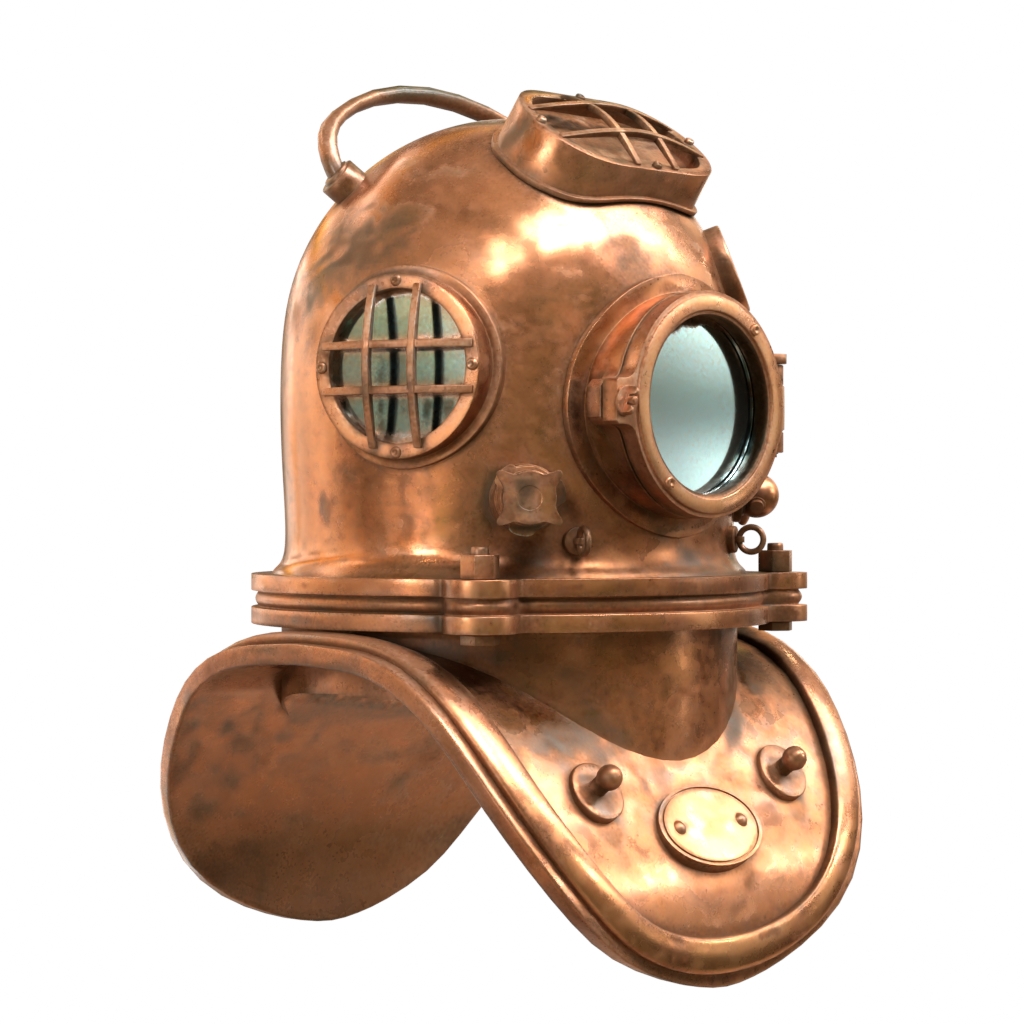}} &
        \raisebox{-0.5\height}{\includegraphics[width=0.093\textwidth]{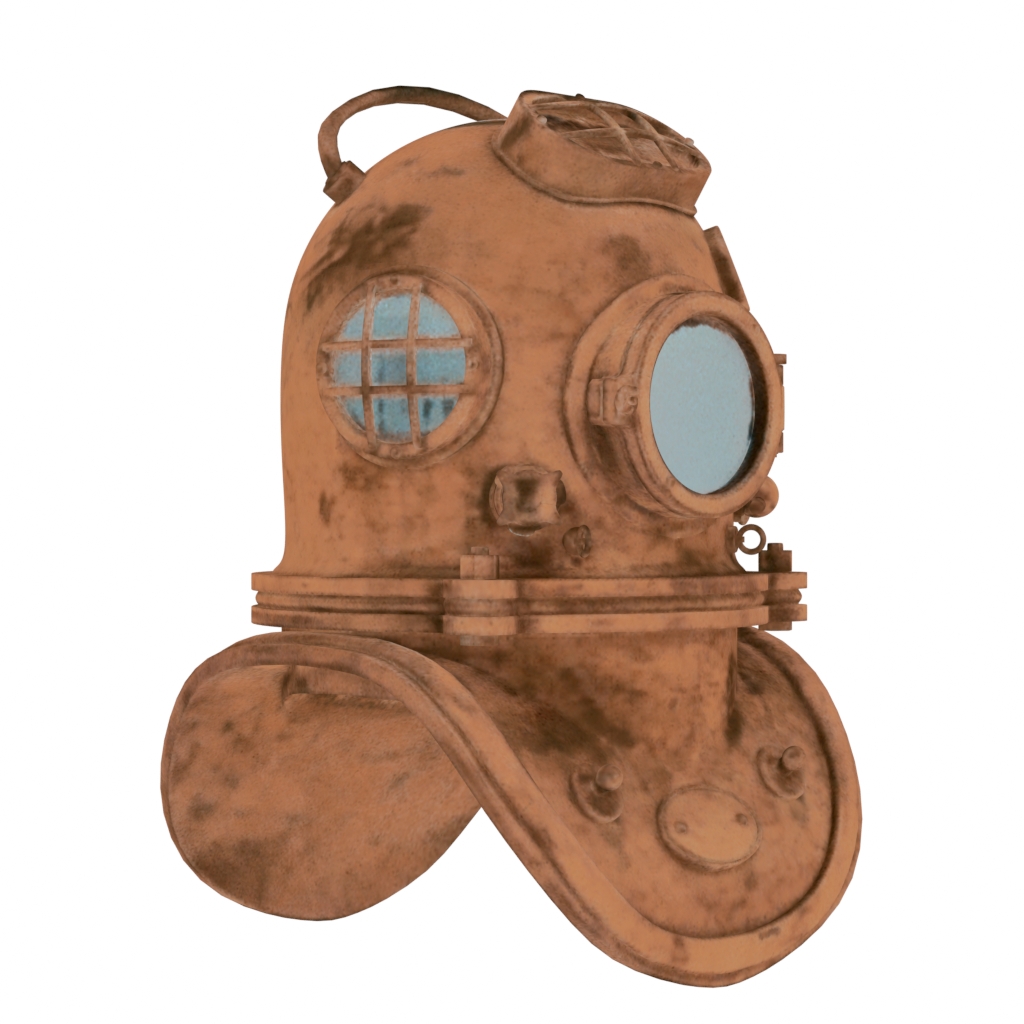}} &
        \raisebox{-0.5\height}{\includegraphics[width=0.093\textwidth]{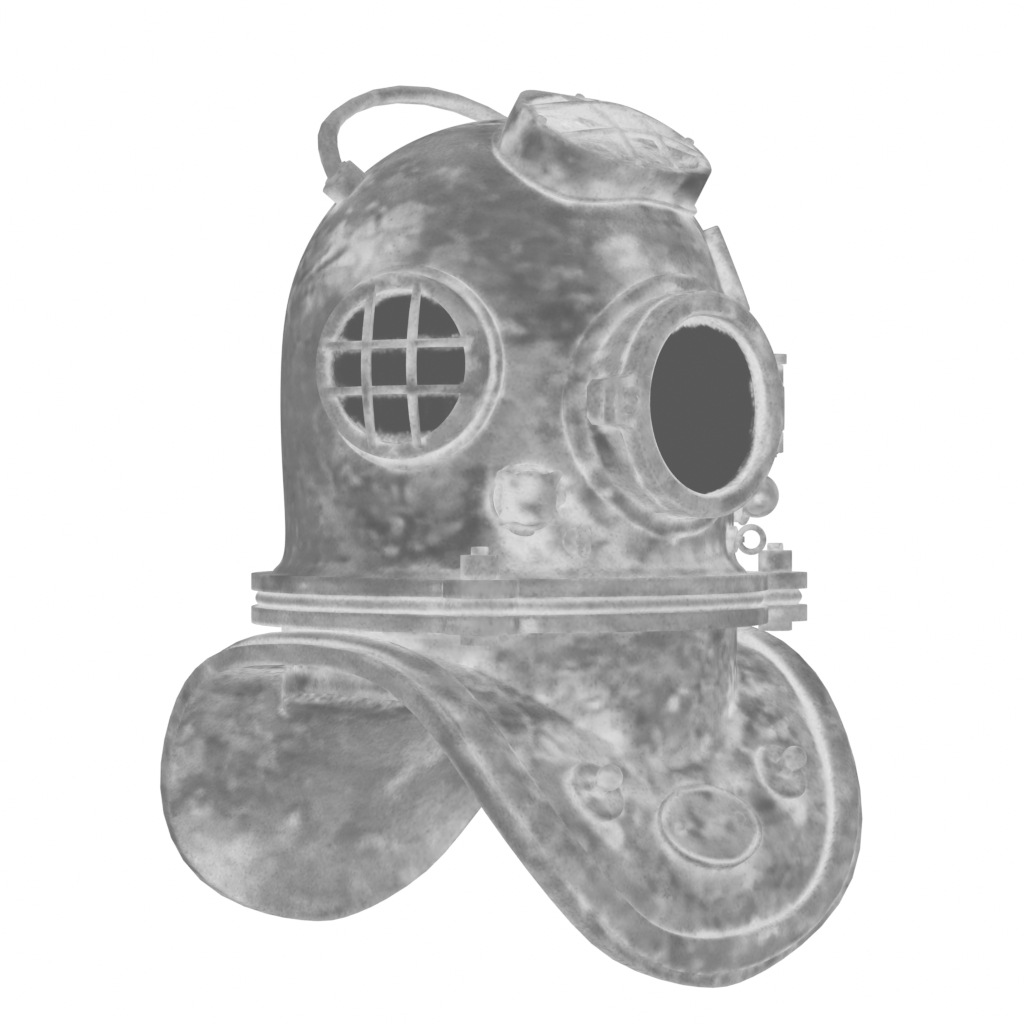}} &
        \raisebox{-0.5\height}{\includegraphics[width=0.093\textwidth]{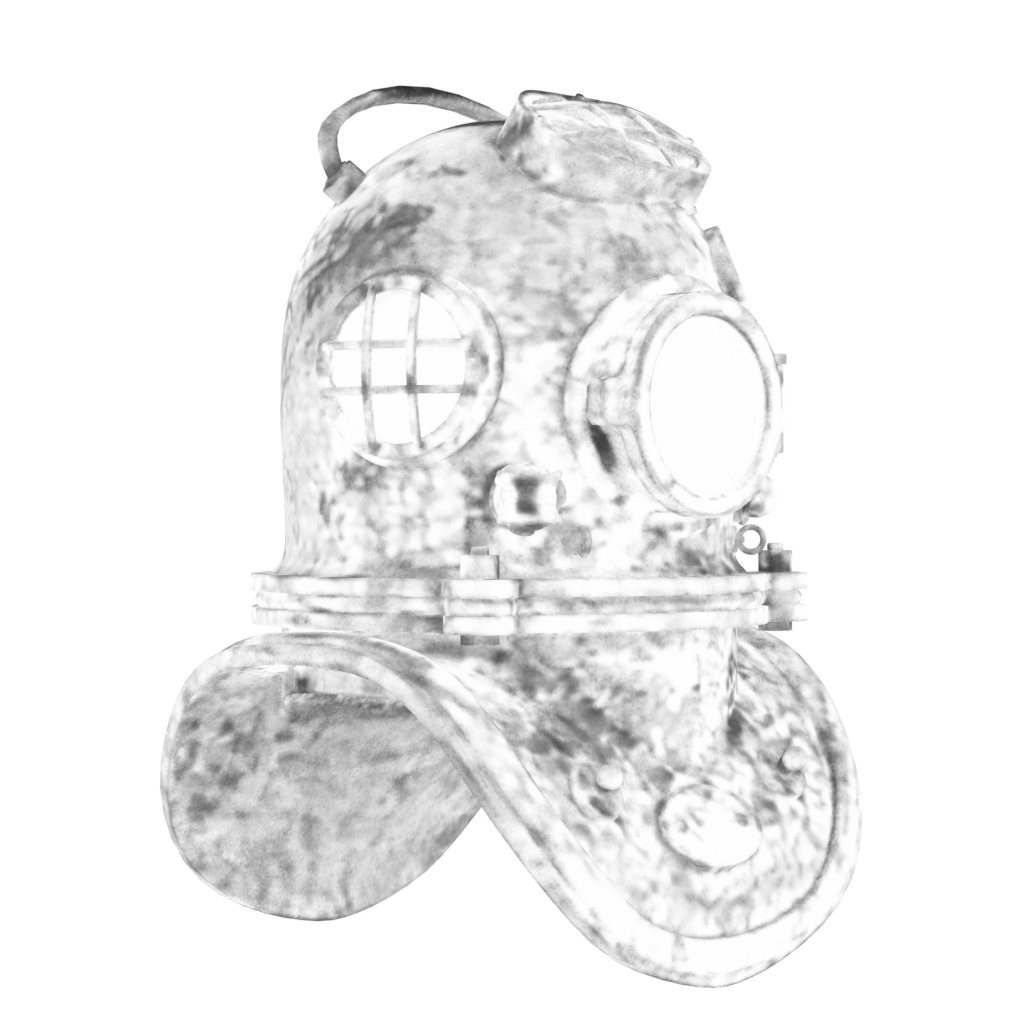}} &
        \rotatebox[origin=c]{90}{\bf Ours (T)} &
        \raisebox{-0.5\height}{\includegraphics[width=0.093\textwidth]{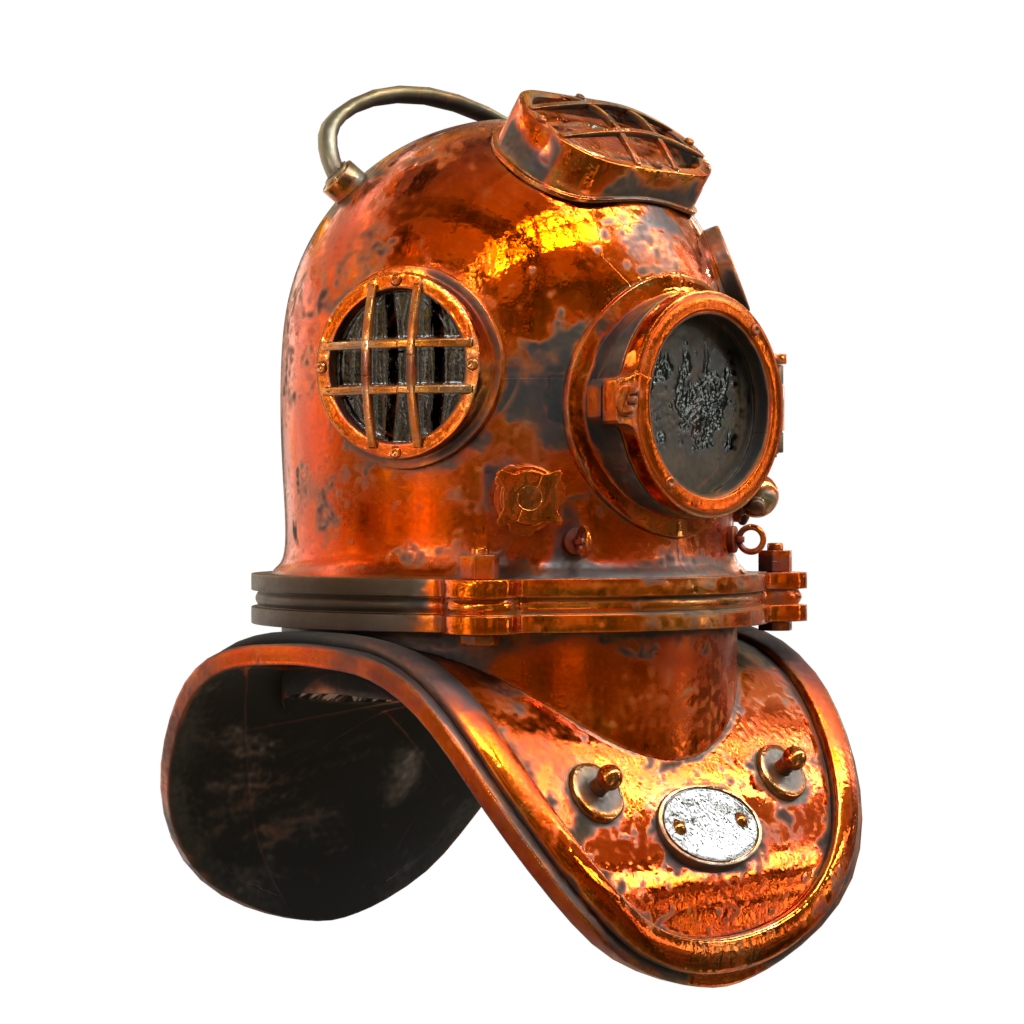}} &
        \raisebox{-0.5\height}{\includegraphics[width=0.093\textwidth]{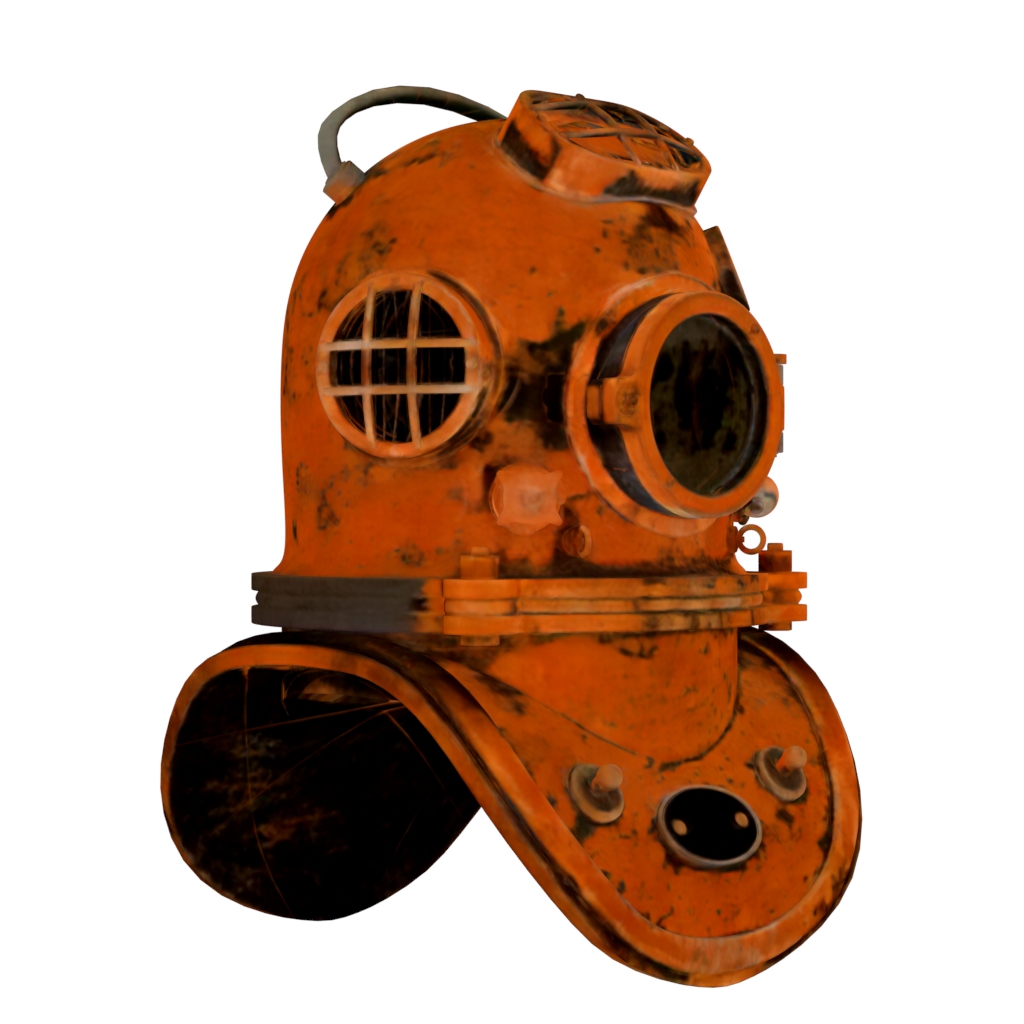}} &
        \raisebox{-0.5\height}{\includegraphics[width=0.093\textwidth]{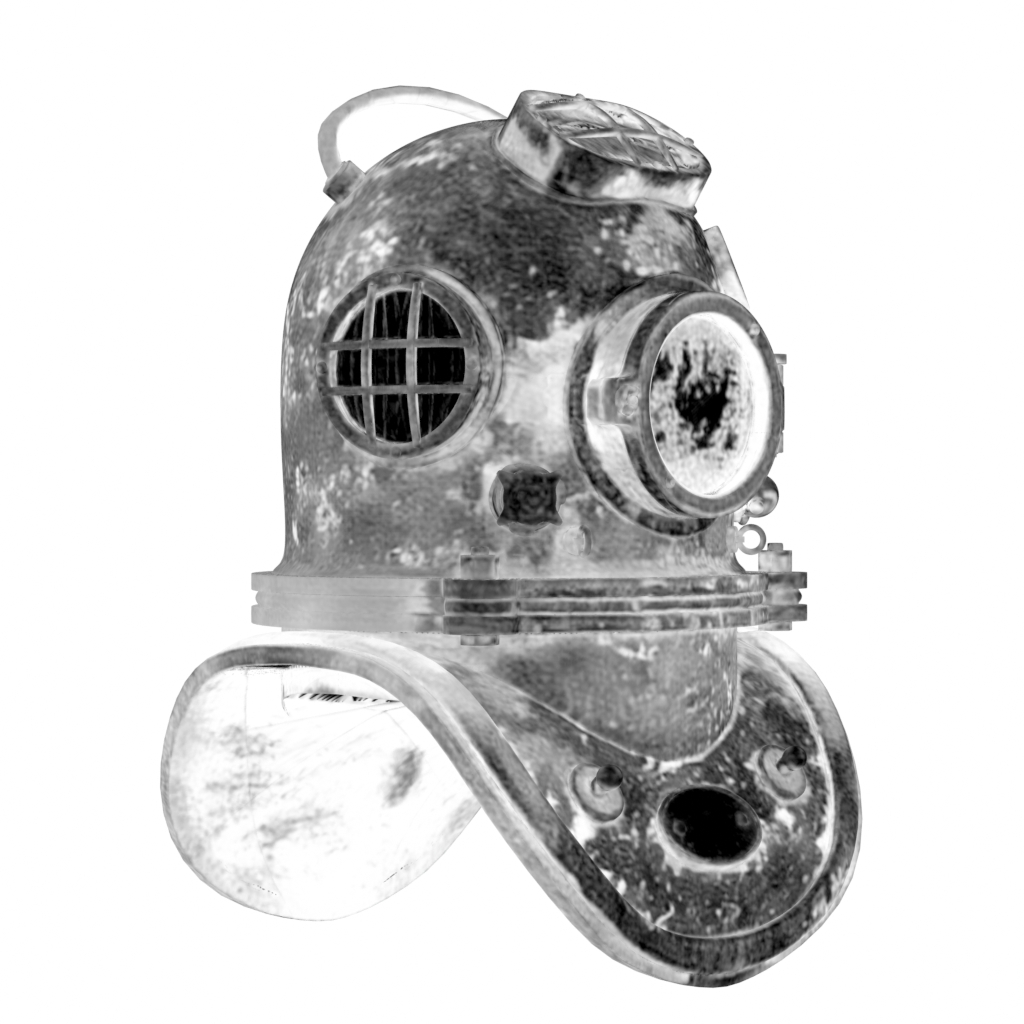}} &
        \raisebox{-0.5\height}{\includegraphics[width=0.093\textwidth]{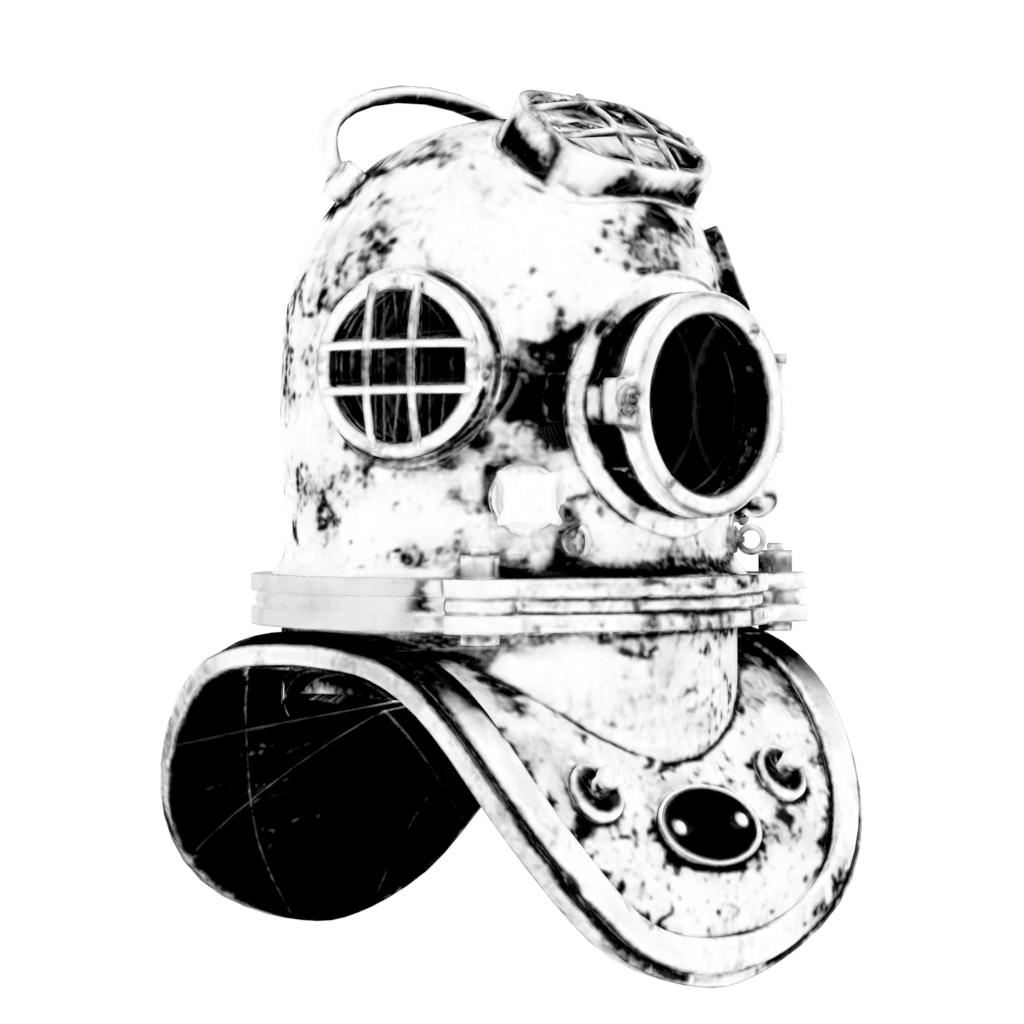}} \\

        \multirow{2}{*}{\rotatebox{90}{\makebox[0.16\textwidth]{\centering \textsc{Robot}}}} &
        \multirow{2}{*}{\includegraphics[width=0.16\textwidth]{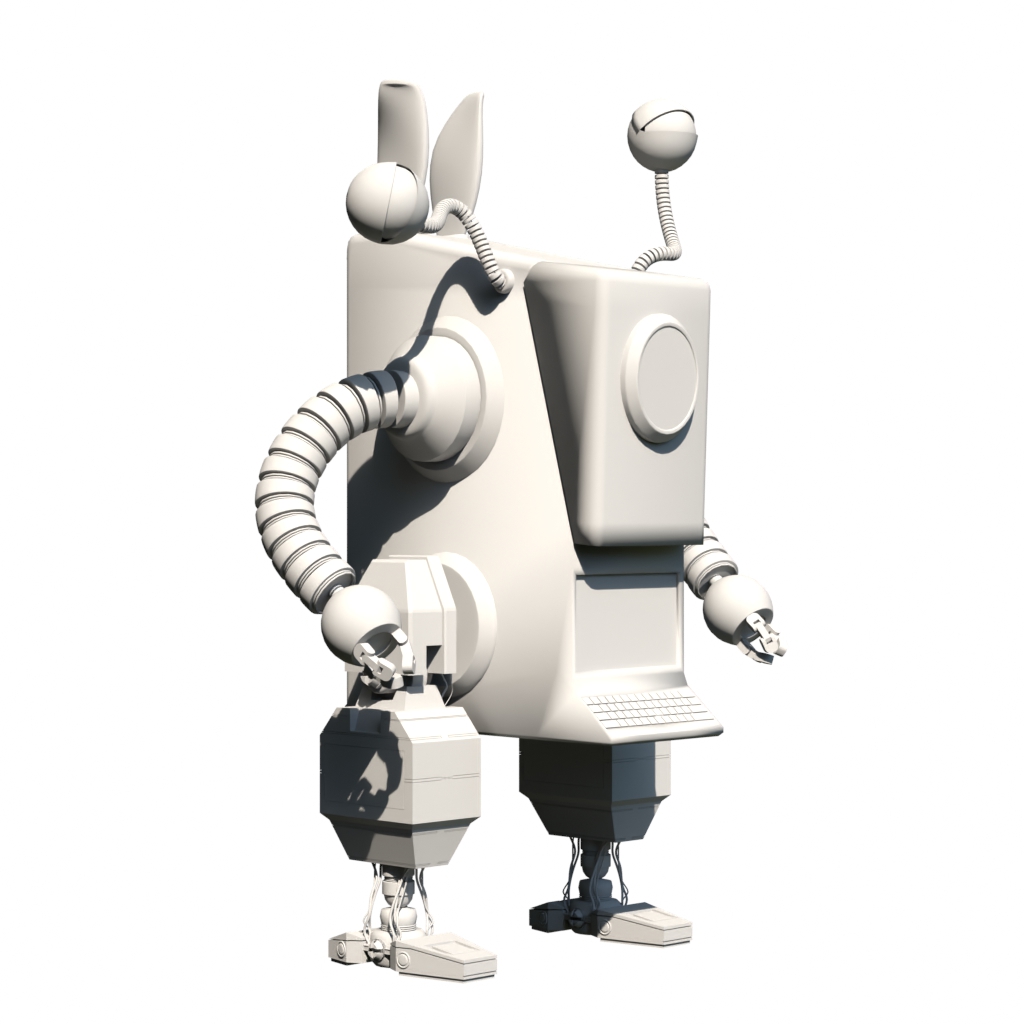}} & 
        \rotatebox[origin=c]{90}{Hunyuan(I)} &
        \raisebox{-0.5\height}{\includegraphics[width=0.093\textwidth]{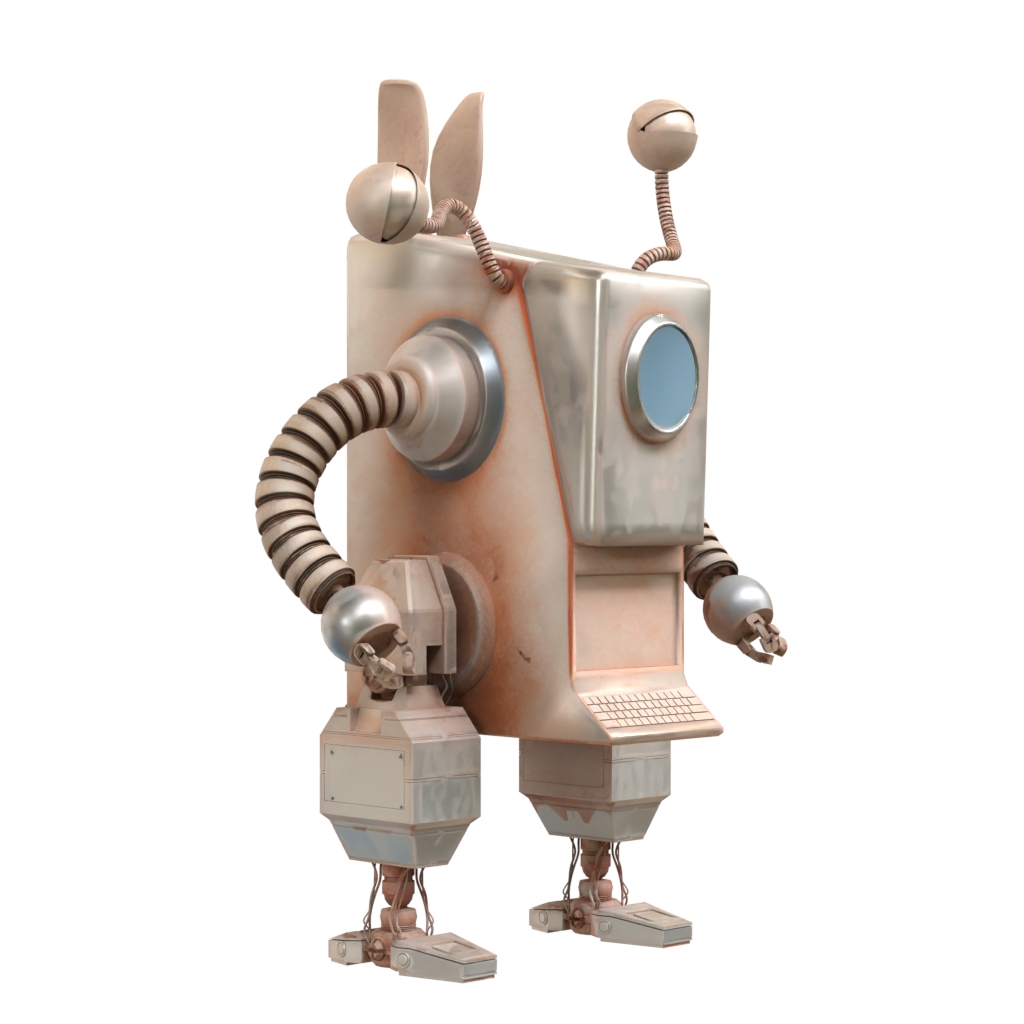}} &
        \raisebox{-0.5\height}{\includegraphics[width=0.093\textwidth]{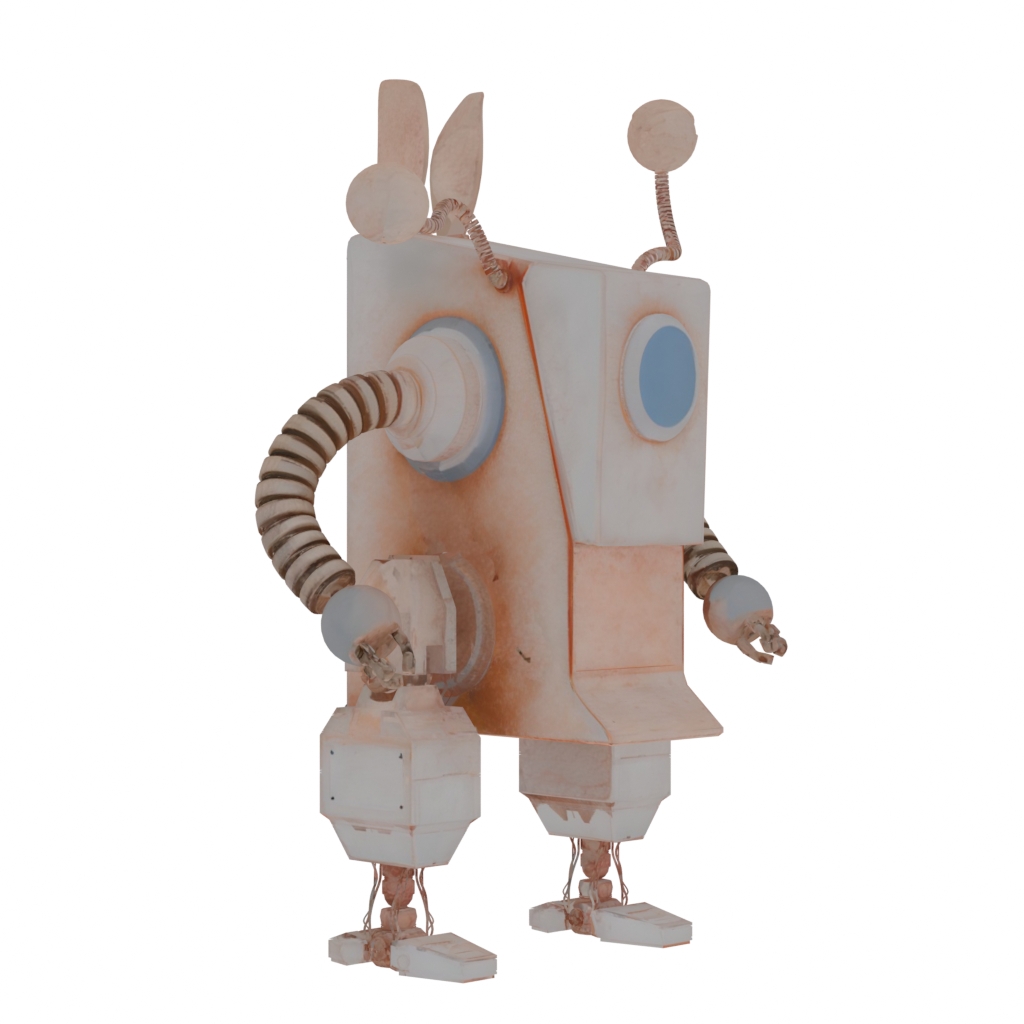}} &
        \raisebox{-0.5\height}{\includegraphics[width=0.093\textwidth]{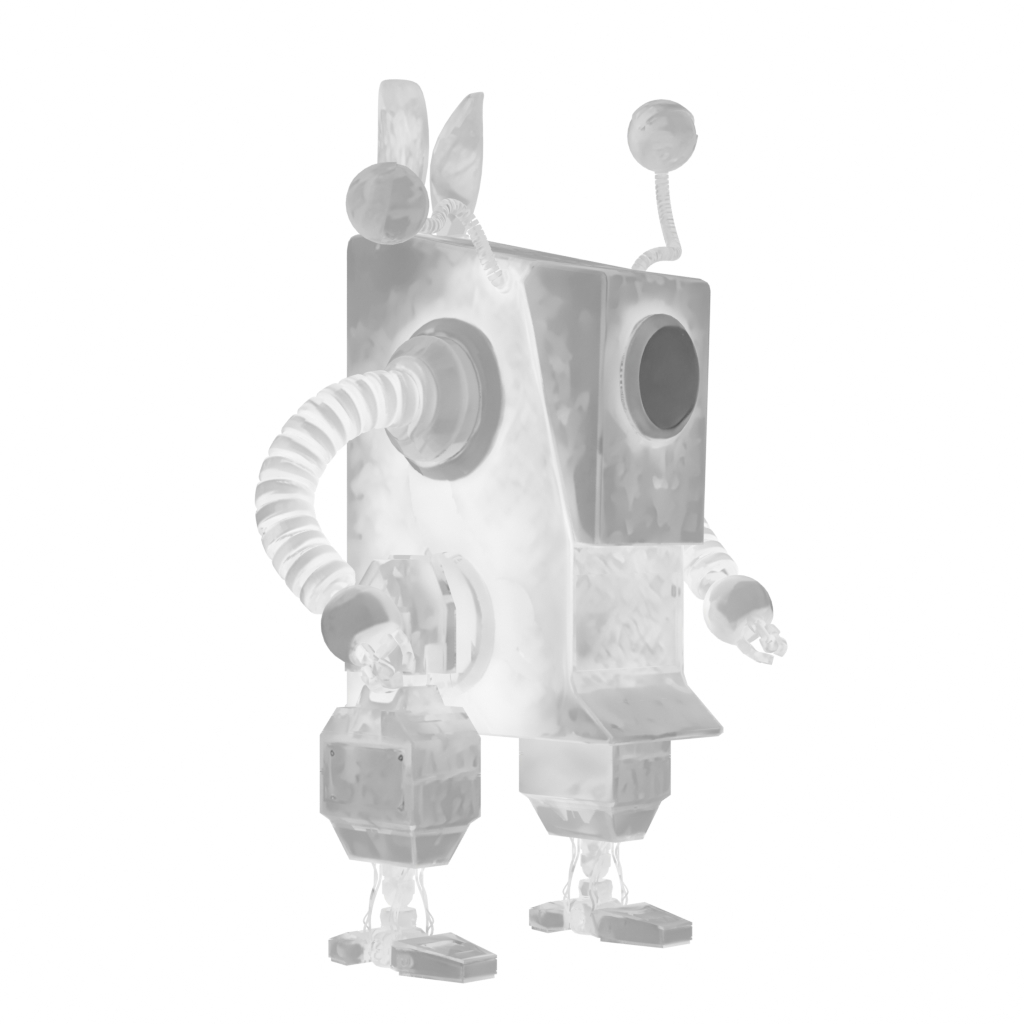}} &
        \raisebox{-0.5\height}{\includegraphics[width=0.093\textwidth]{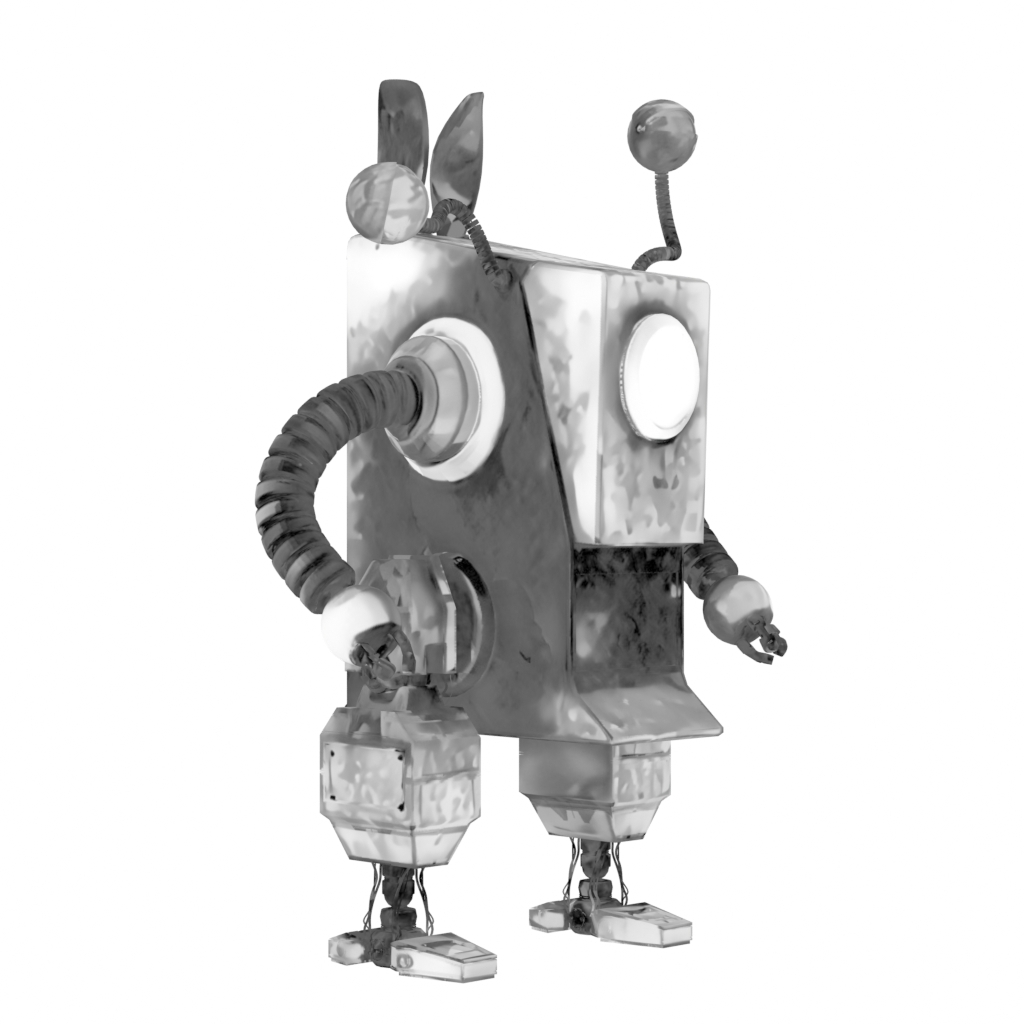}} &
        \rotatebox[origin=c]{90}{Hunyuan(T)} &
        \raisebox{-0.5\height}{\includegraphics[width=0.093\textwidth]{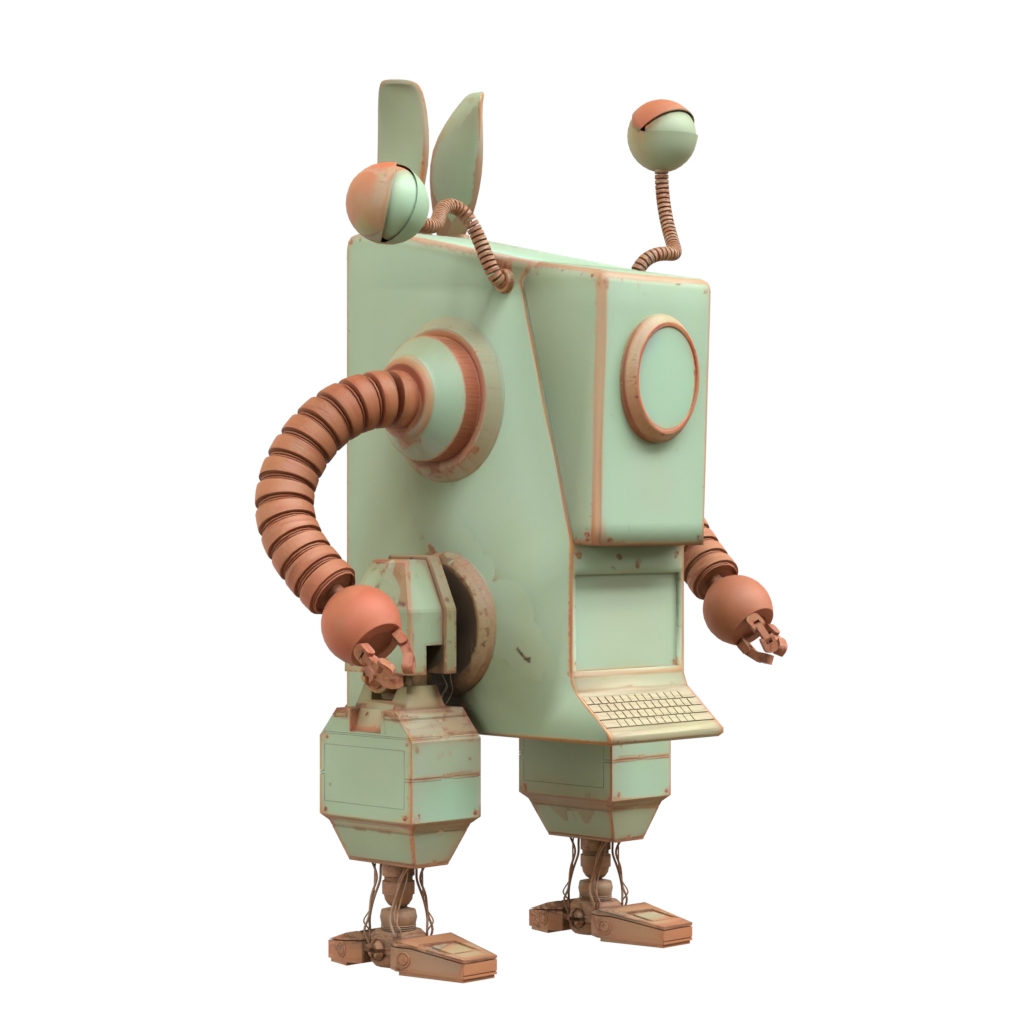}} &
        \raisebox{-0.5\height}{\includegraphics[width=0.093\textwidth]{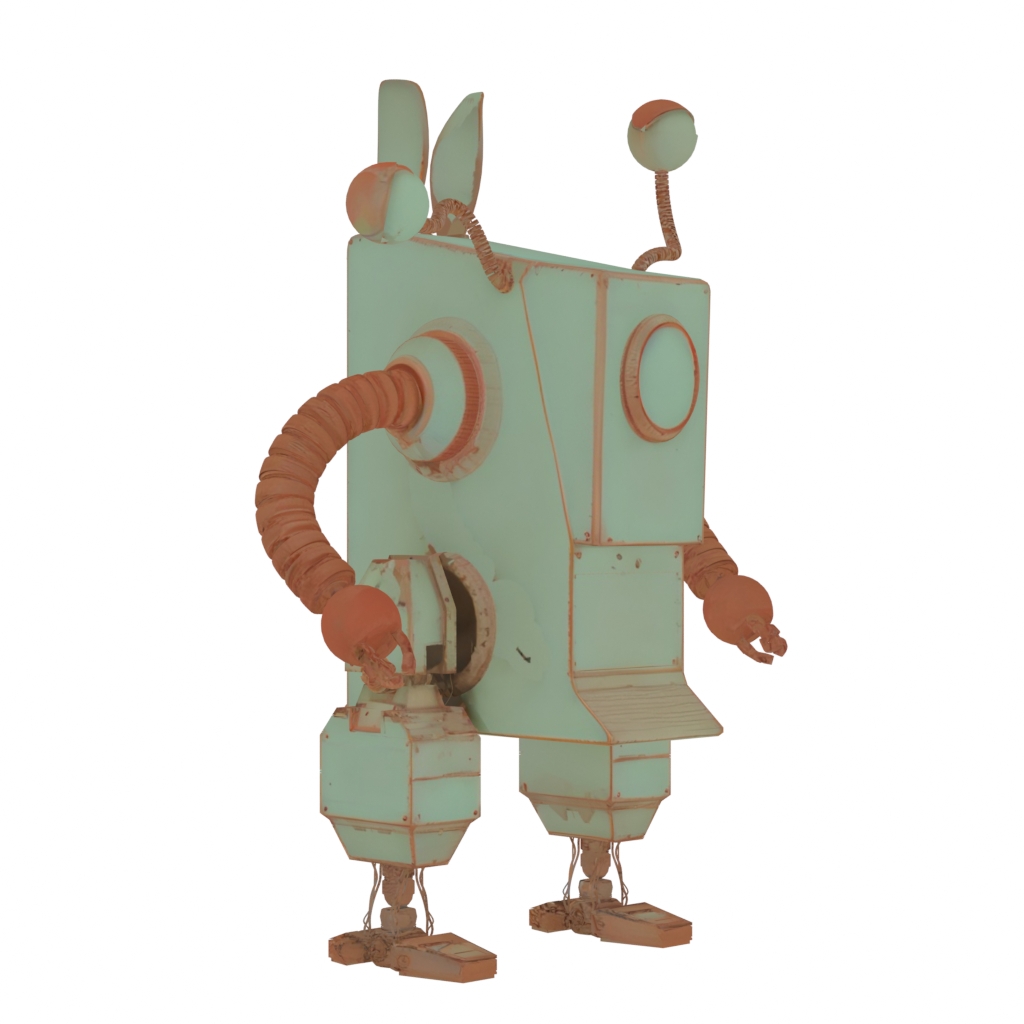}} &
        \raisebox{-0.5\height}{\includegraphics[width=0.093\textwidth]{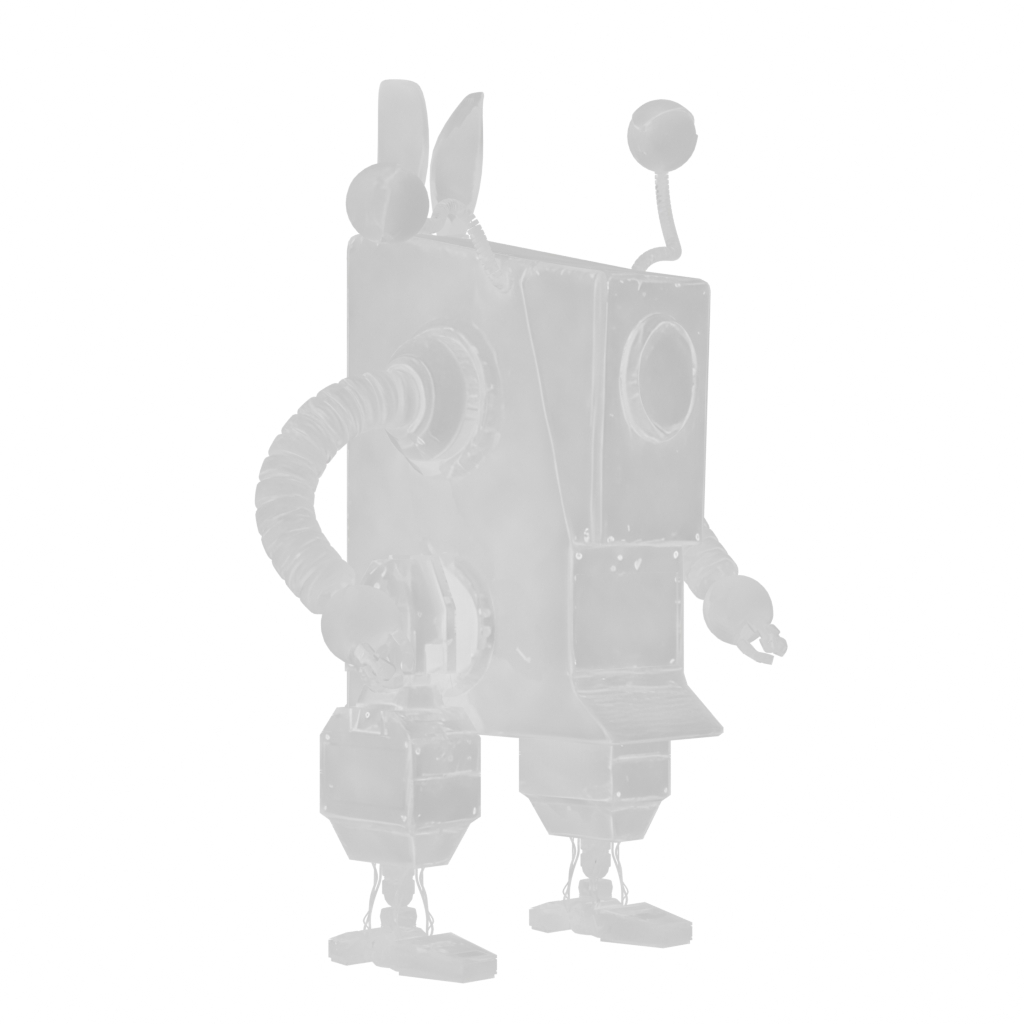}} &
        \raisebox{-0.5\height}{\includegraphics[width=0.093\textwidth]{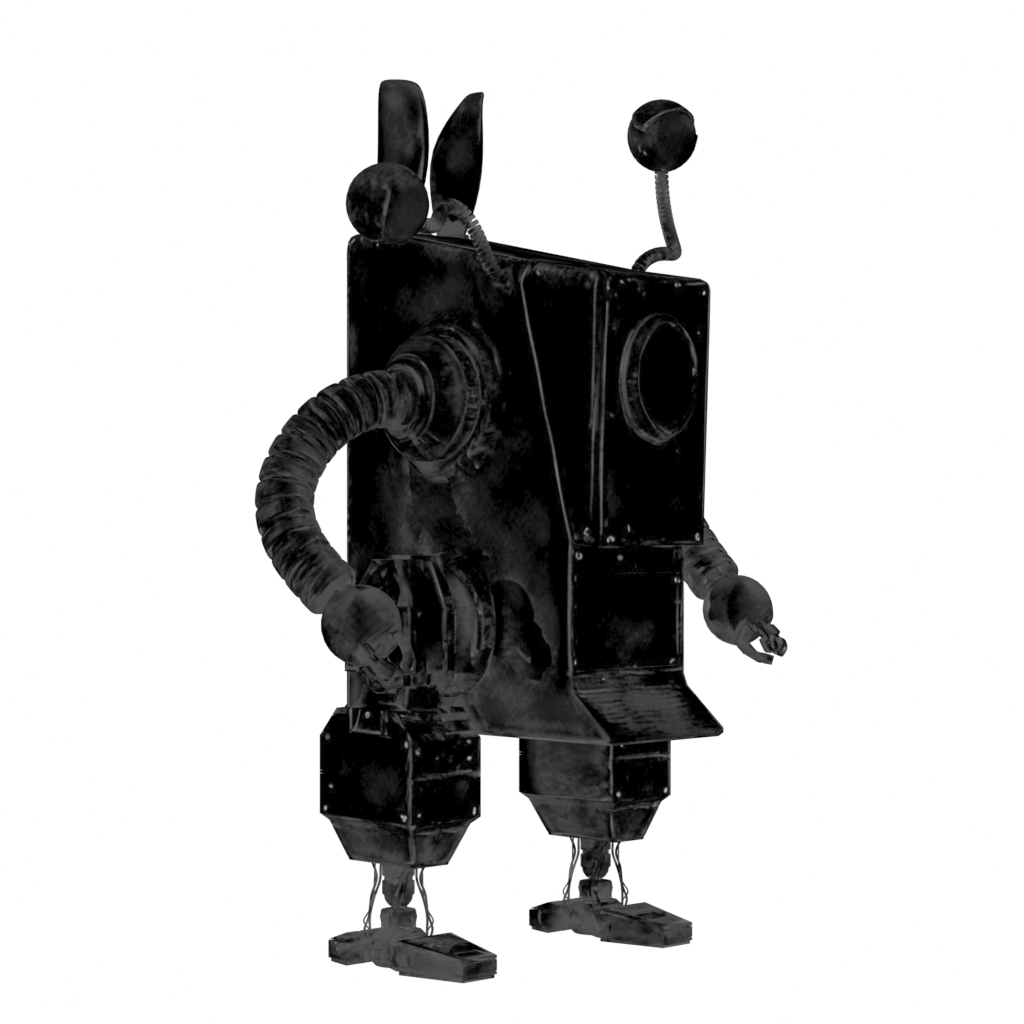}} \\
        & & 
        \rotatebox[origin=c]{90}{VideoMat (T)} &
        \raisebox{-0.5\height}{\includegraphics[width=0.093\textwidth]{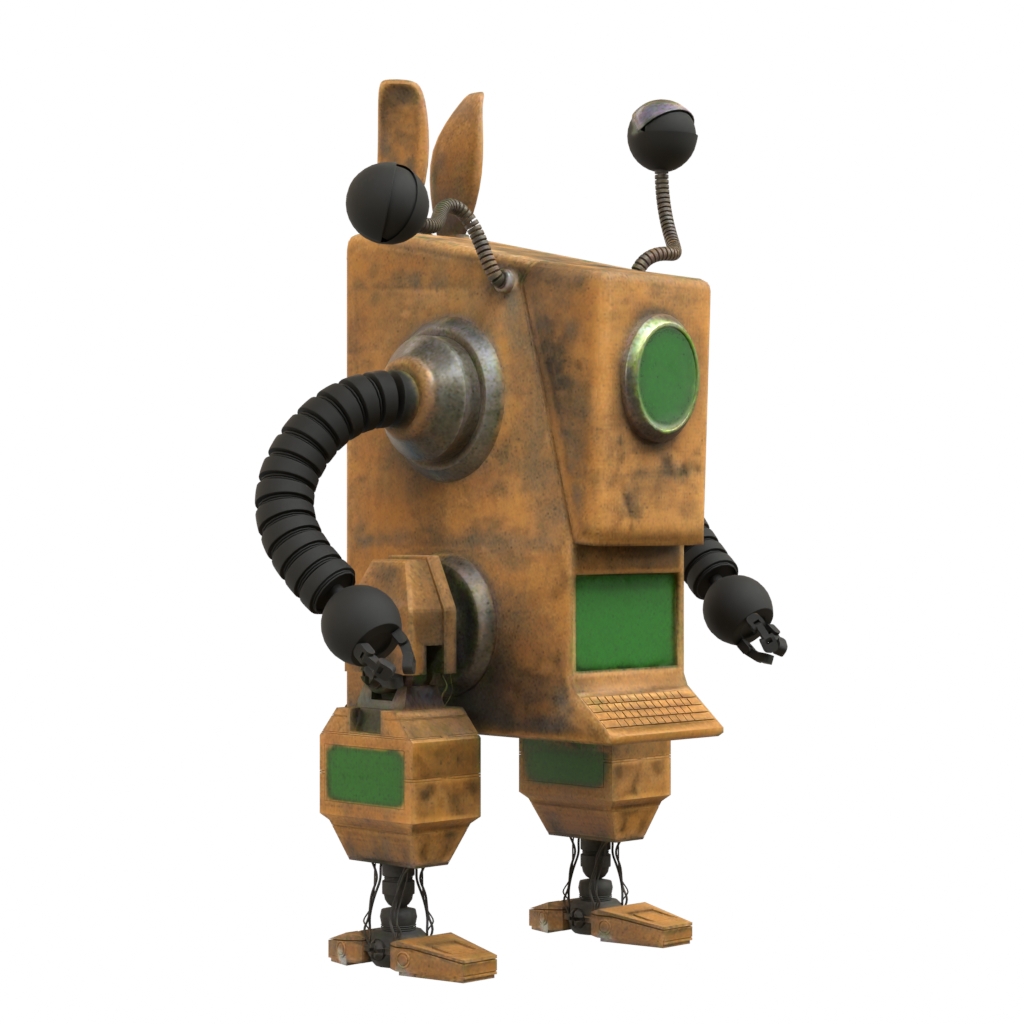}} &
        \raisebox{-0.5\height}{\includegraphics[width=0.093\textwidth]{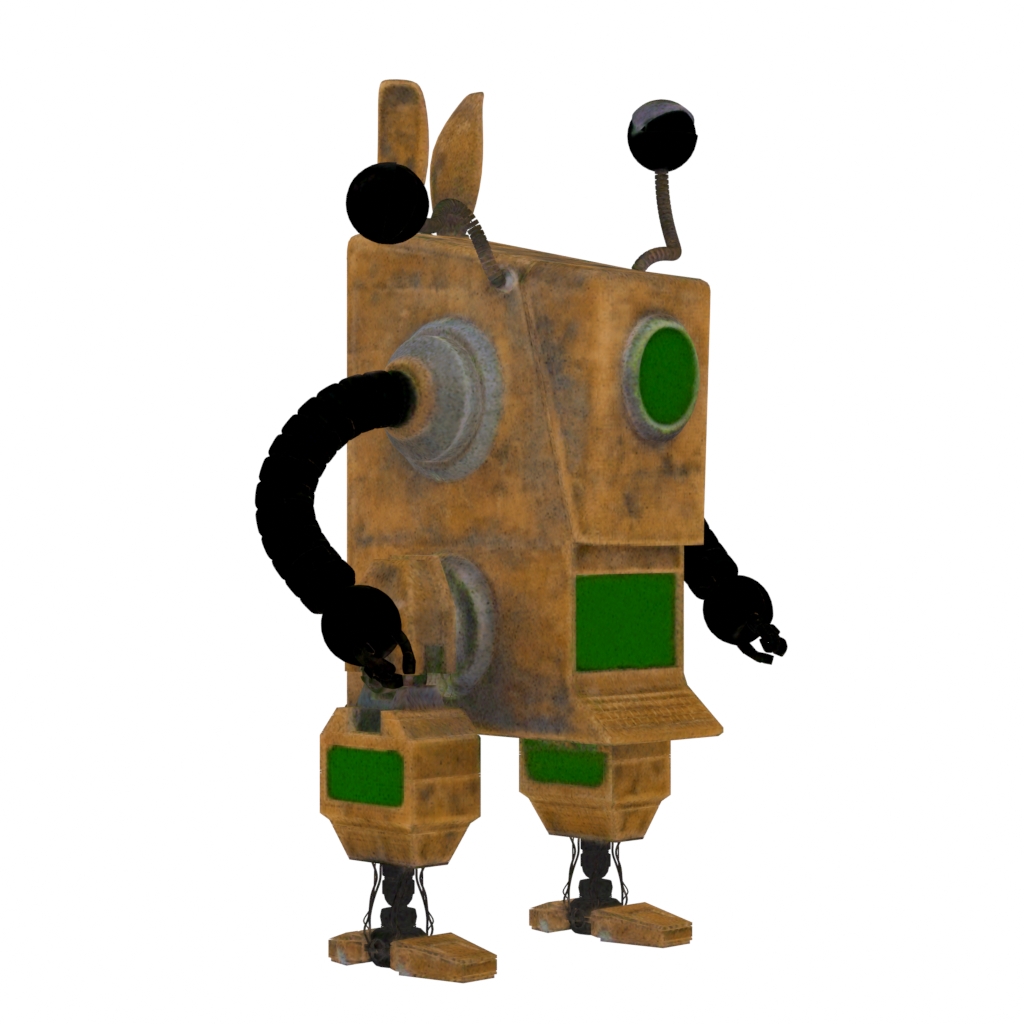}} &
        \raisebox{-0.5\height}{\includegraphics[width=0.093\textwidth]{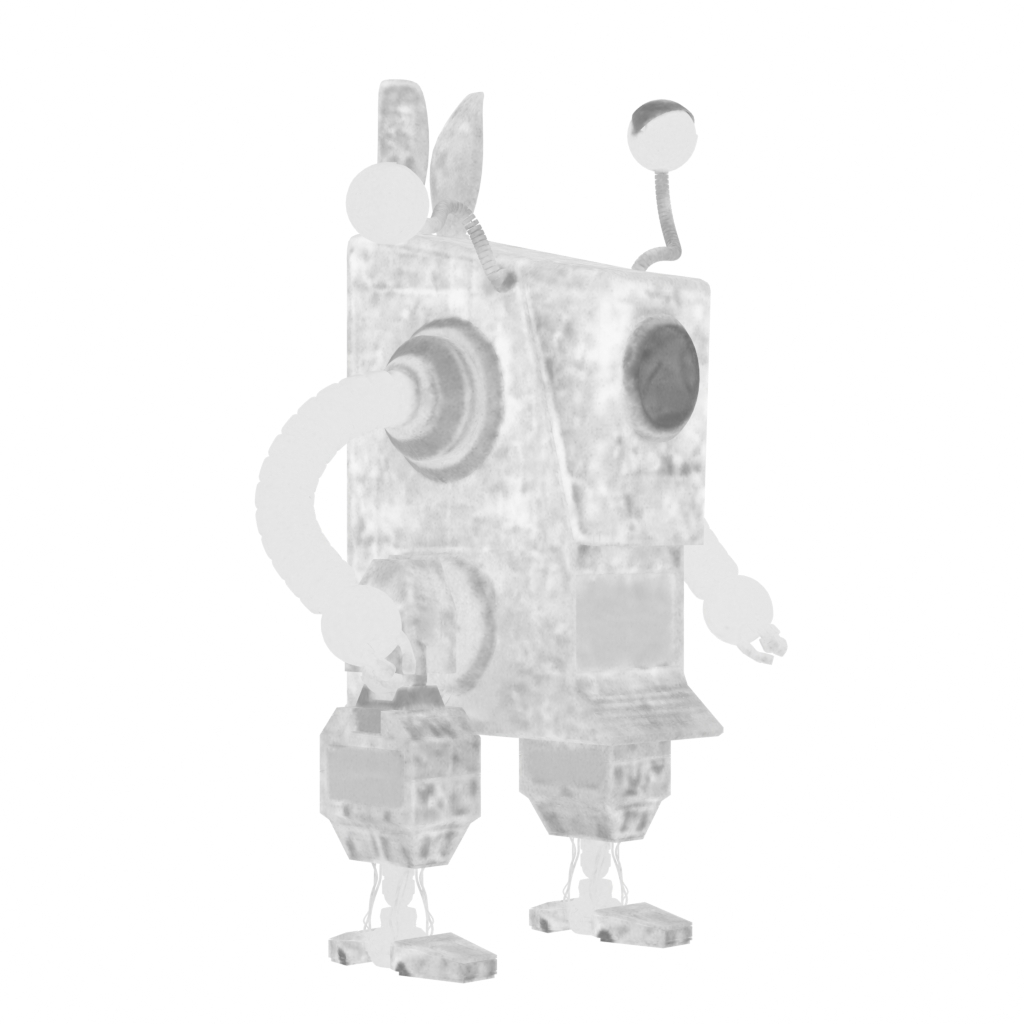}} &
        \raisebox{-0.5\height}{\includegraphics[width=0.093\textwidth]{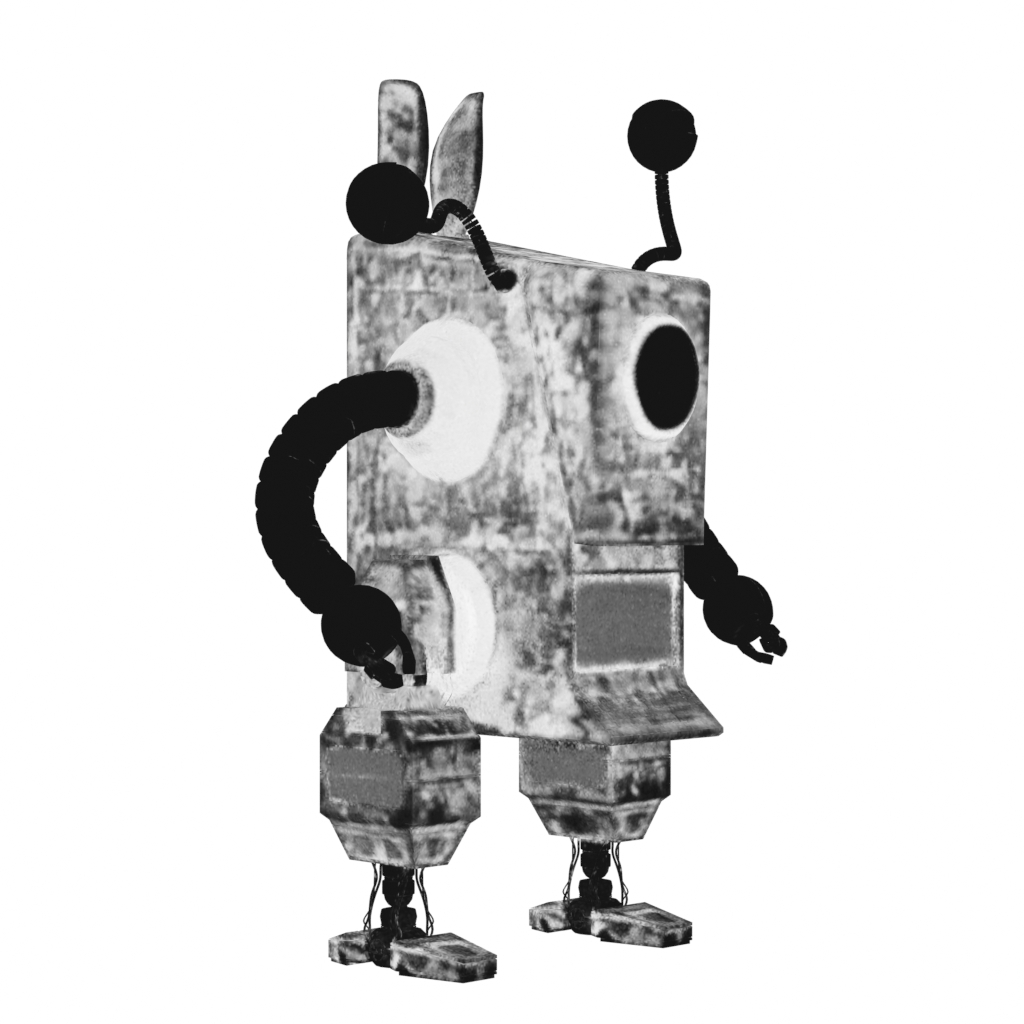}} &
        \rotatebox[origin=c]{90}{\bf Ours (T)} &
        \raisebox{-0.5\height}{\includegraphics[width=0.093\textwidth]{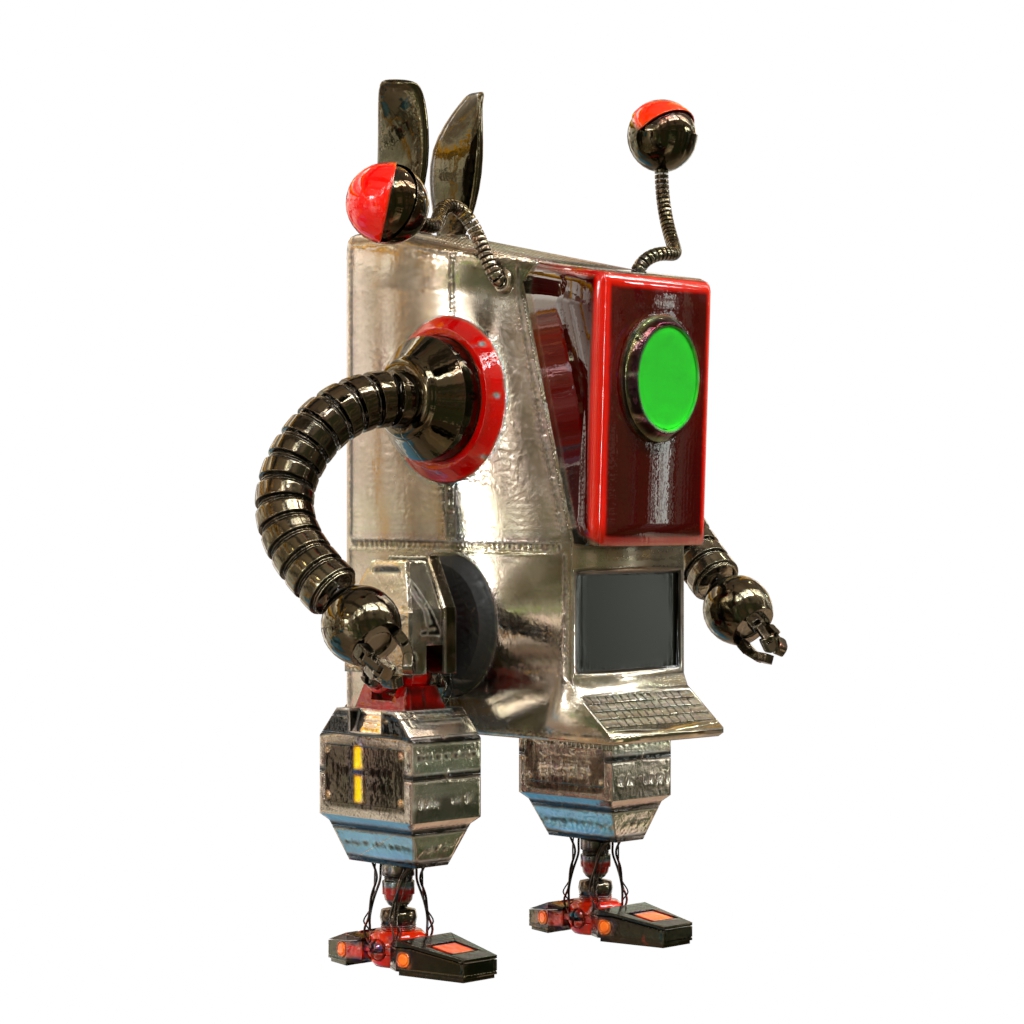}} &
        \raisebox{-0.5\height}{\includegraphics[width=0.093\textwidth]{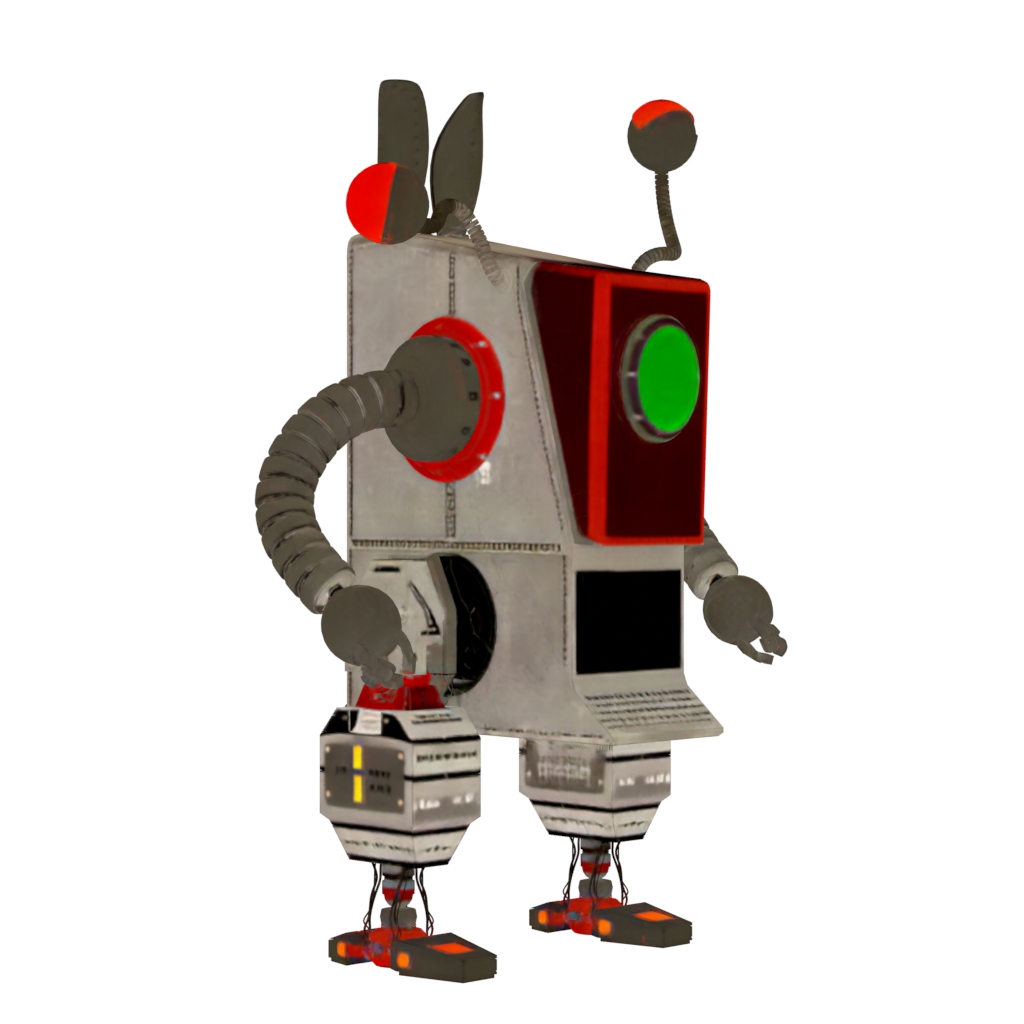}} &
        \raisebox{-0.5\height}{\includegraphics[width=0.093\textwidth]{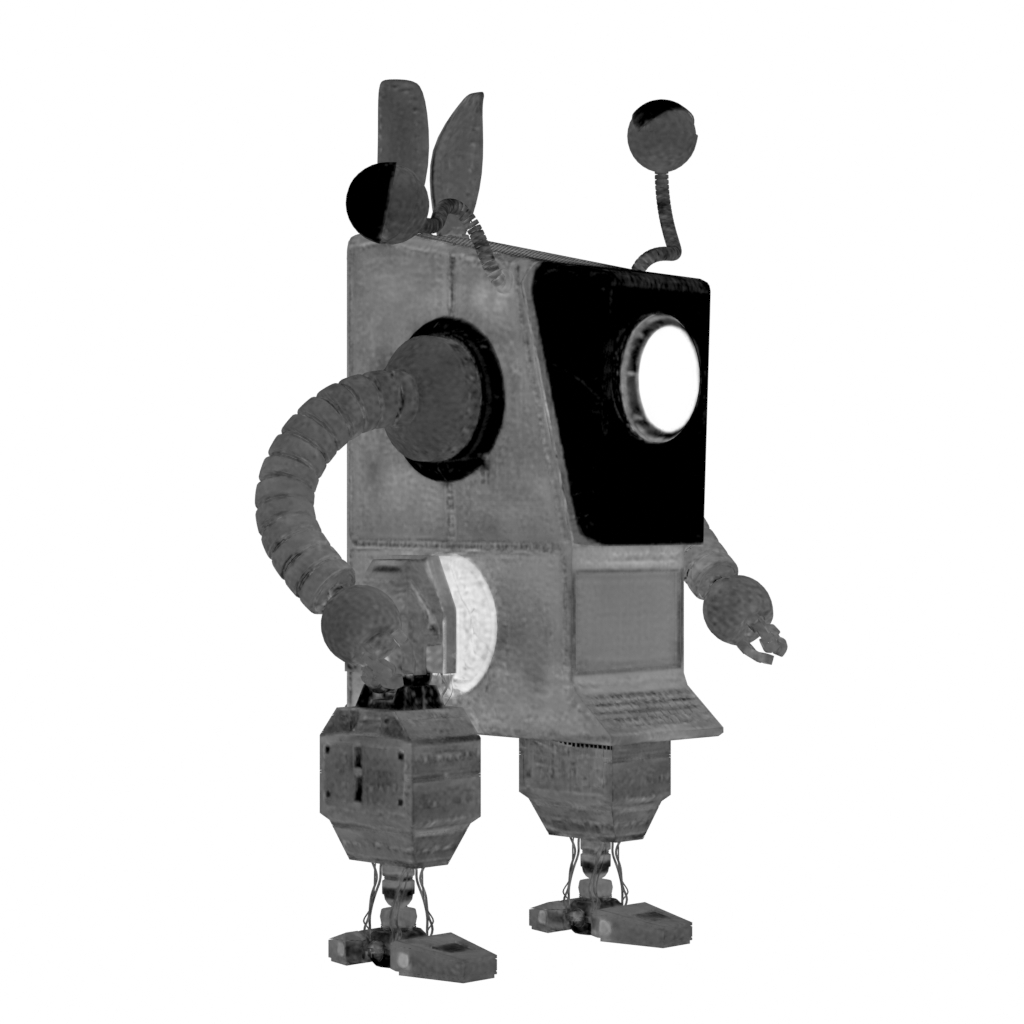}} &
        \raisebox{-0.5\height}{\includegraphics[width=0.093\textwidth]{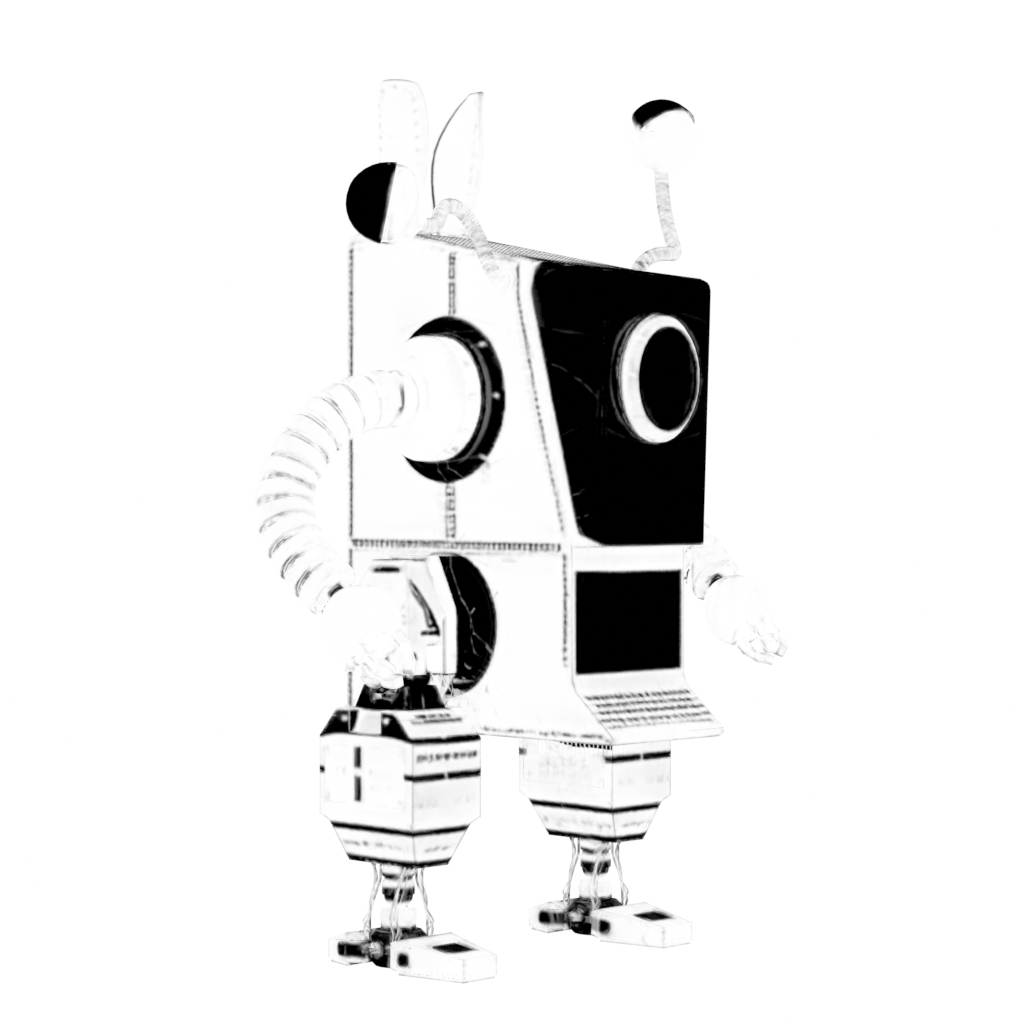}} \\

        \multirow{2}{*}{\rotatebox{90}{\makebox[0.16\textwidth]{\centering \textsc{Shed}}}} &
        \multirow{2}{*}{\includegraphics[width=0.16\textwidth]{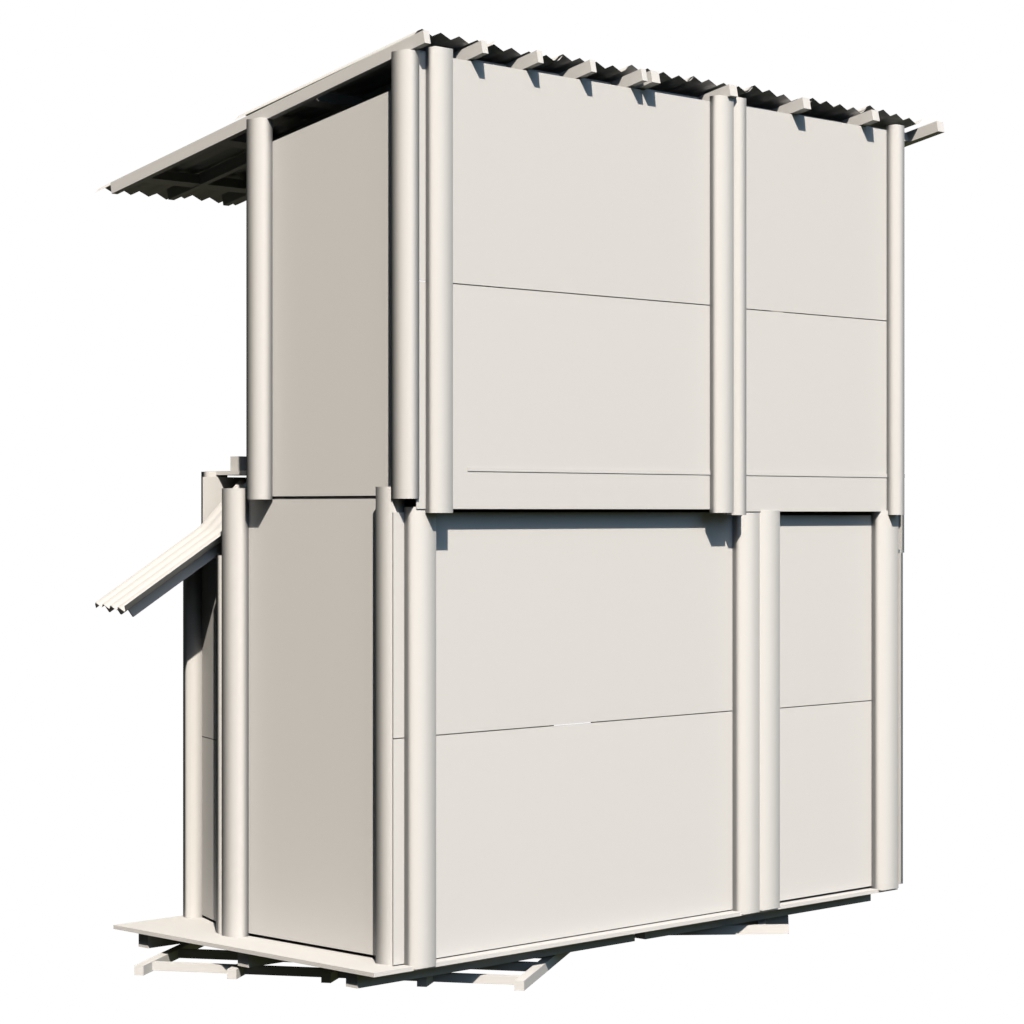}} & 
        \rotatebox[origin=c]{90}{Hunyuan(I)} &
        \raisebox{-0.5\height}{\includegraphics[width=0.093\textwidth]{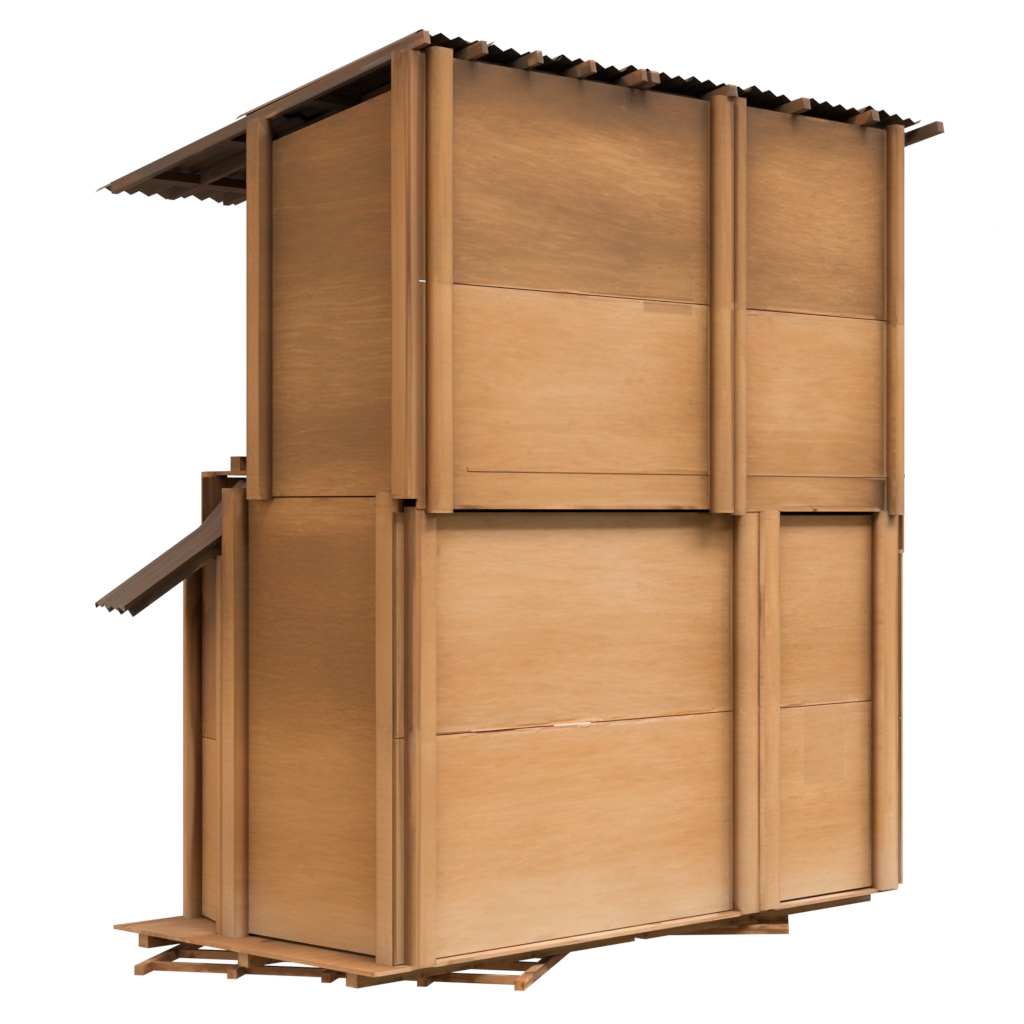}} &
        \raisebox{-0.5\height}{\includegraphics[width=0.093\textwidth]{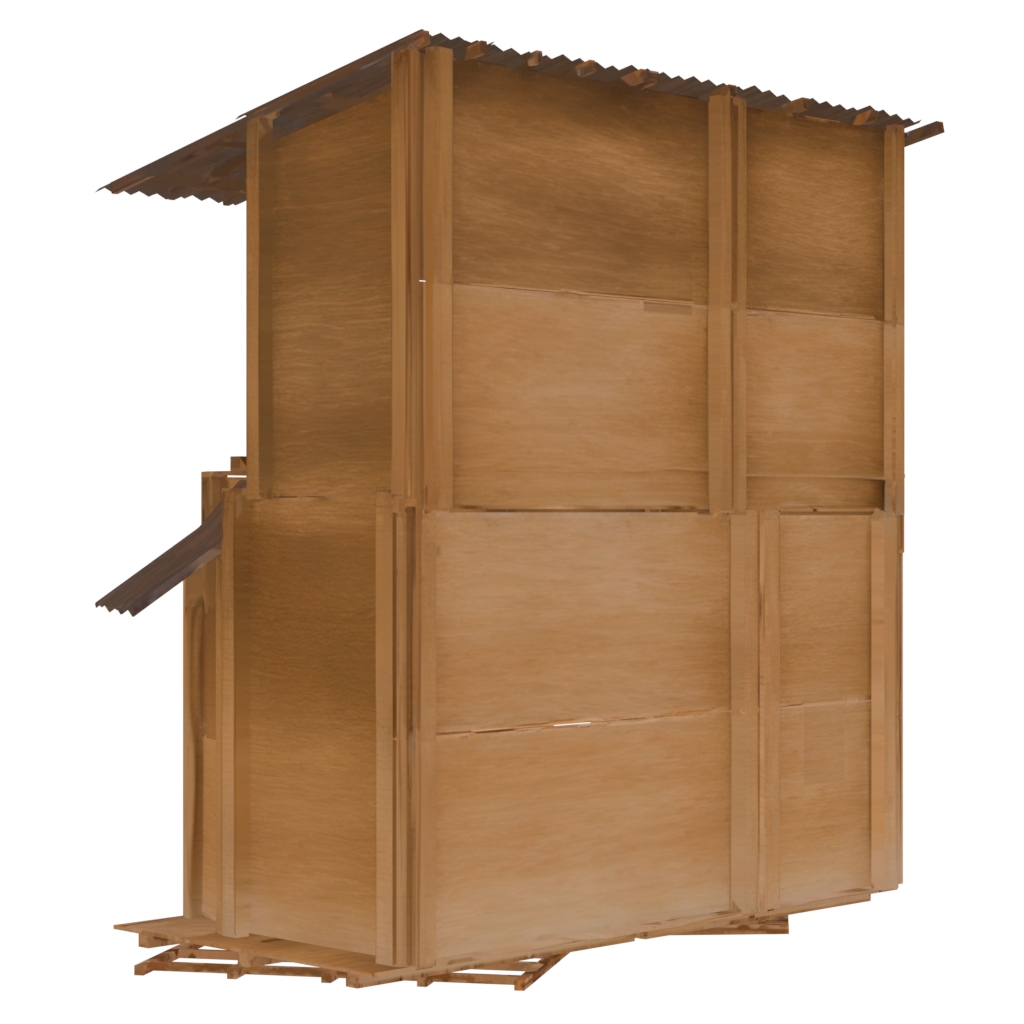}} &
        \raisebox{-0.5\height}{\includegraphics[width=0.093\textwidth]{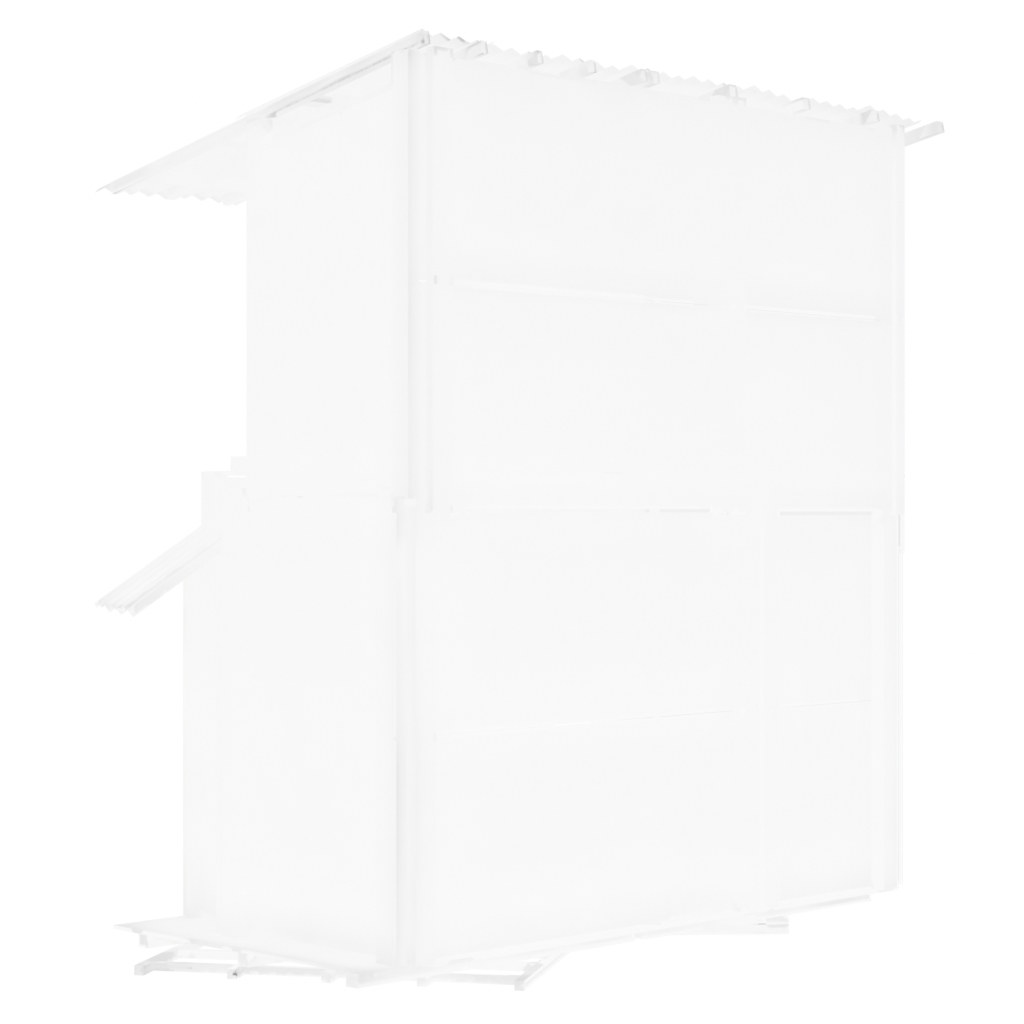}} &
        \raisebox{-0.5\height}{\includegraphics[width=0.093\textwidth]{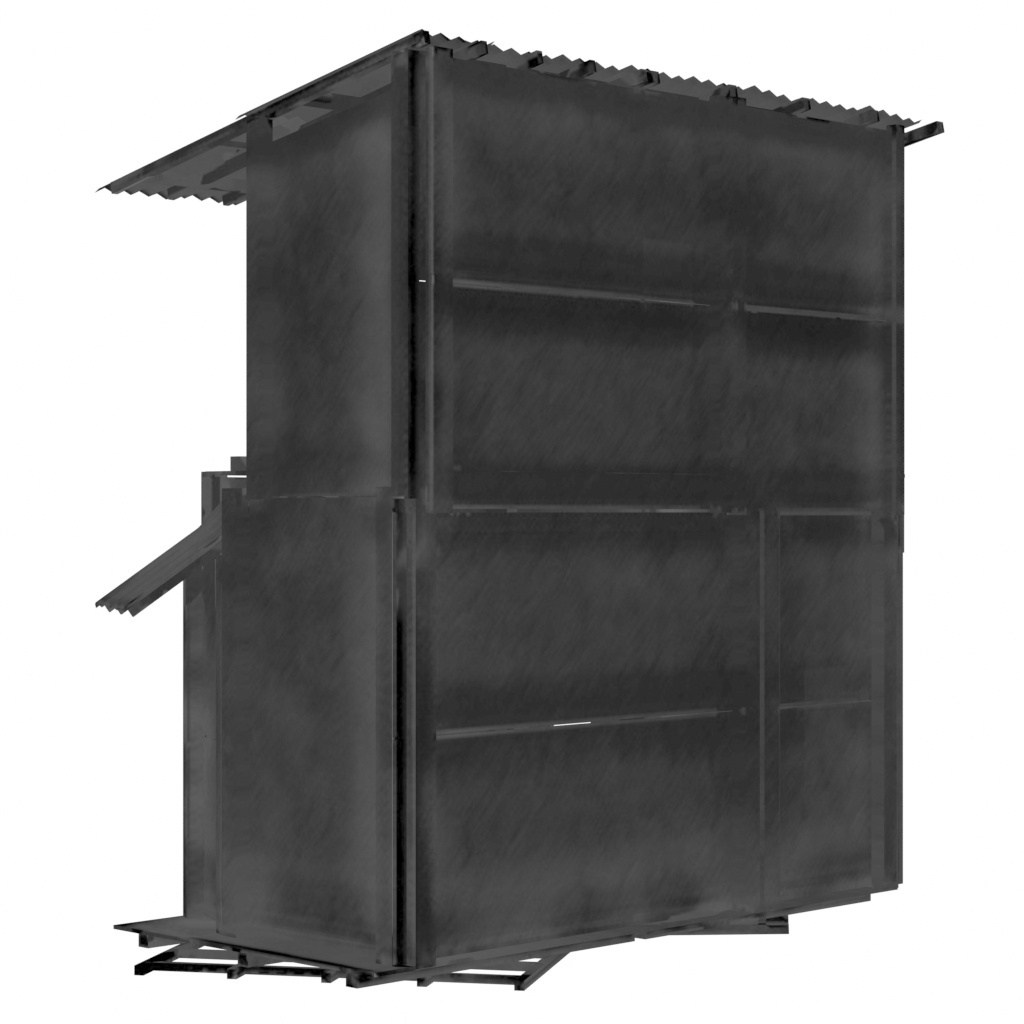}} &
        \rotatebox[origin=c]{90}{Hunyuan(T)} &
        \raisebox{-0.5\height}{\includegraphics[width=0.093\textwidth]{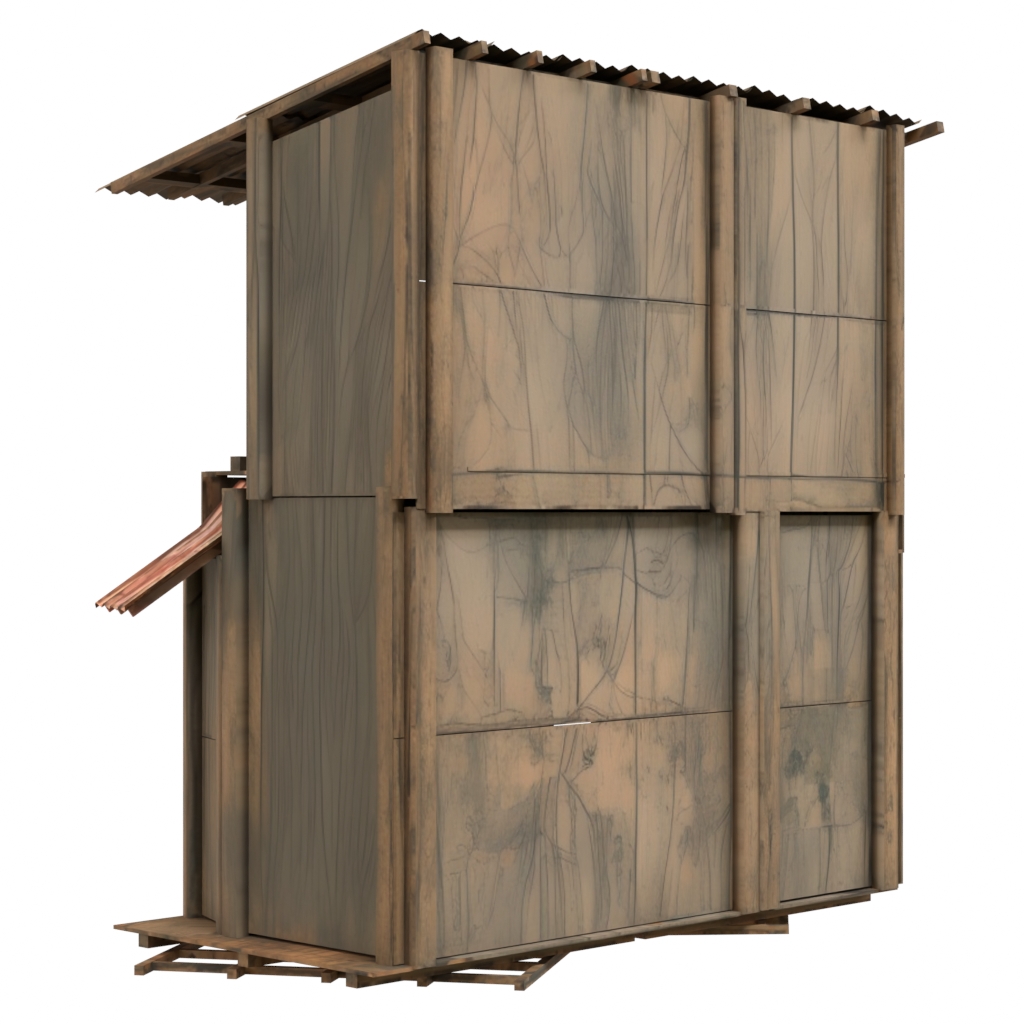}} &
        \raisebox{-0.5\height}{\includegraphics[width=0.093\textwidth]{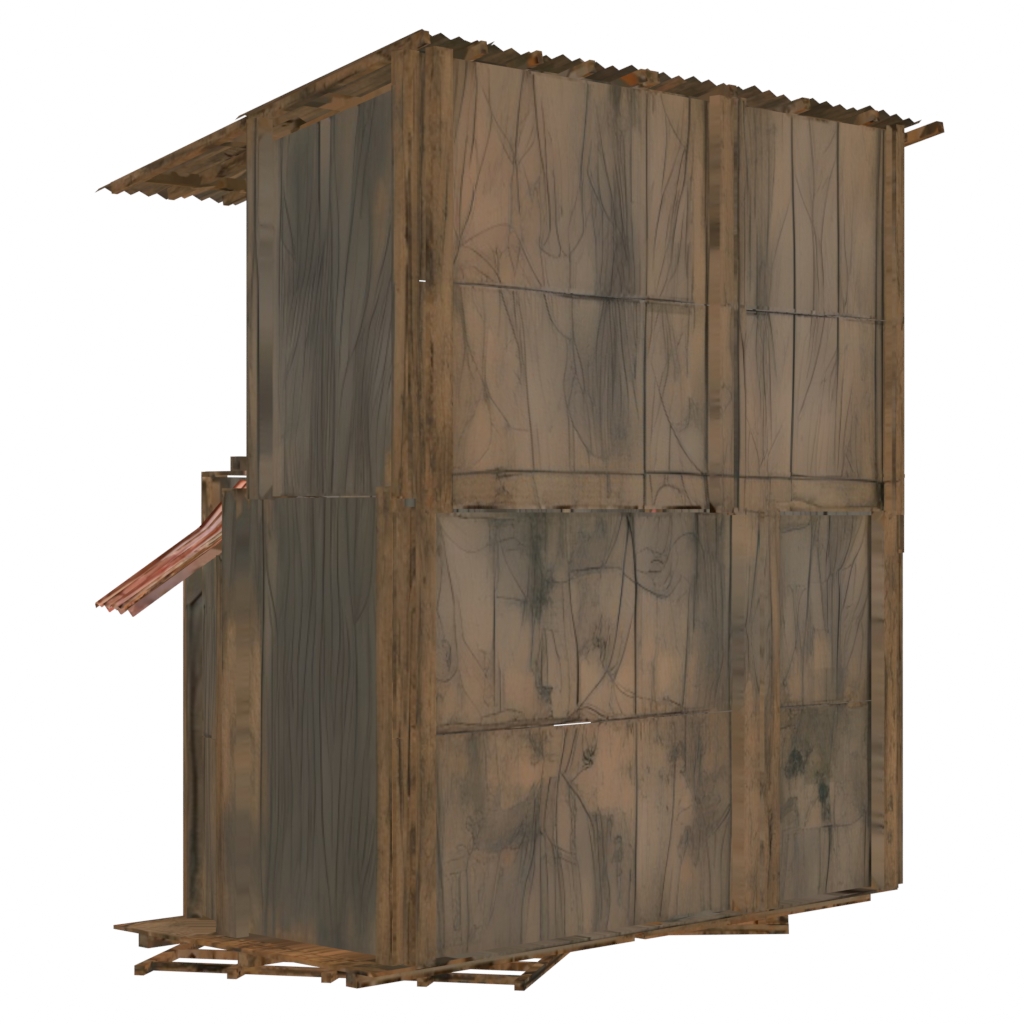}} &
        \raisebox{-0.5\height}{\includegraphics[width=0.093\textwidth]{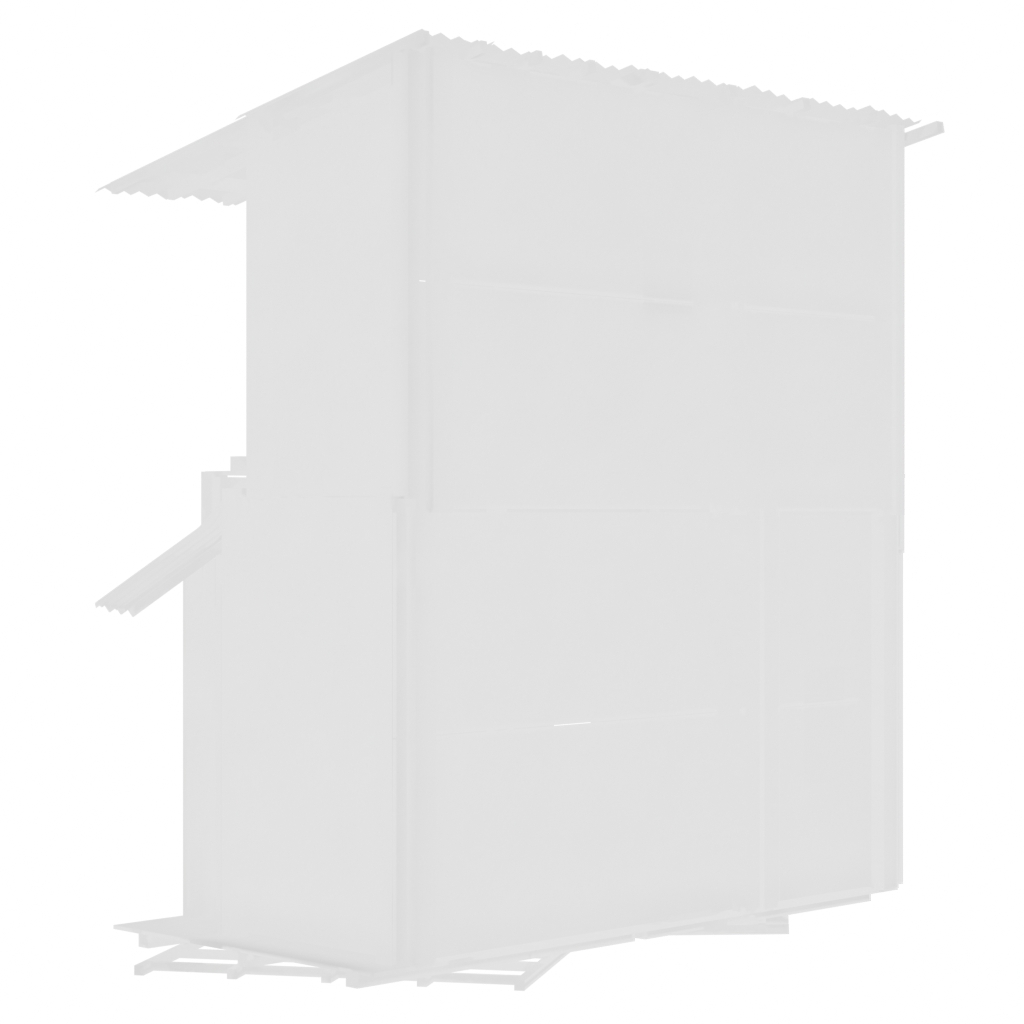}} &
        \raisebox{-0.5\height}{\includegraphics[width=0.093\textwidth]{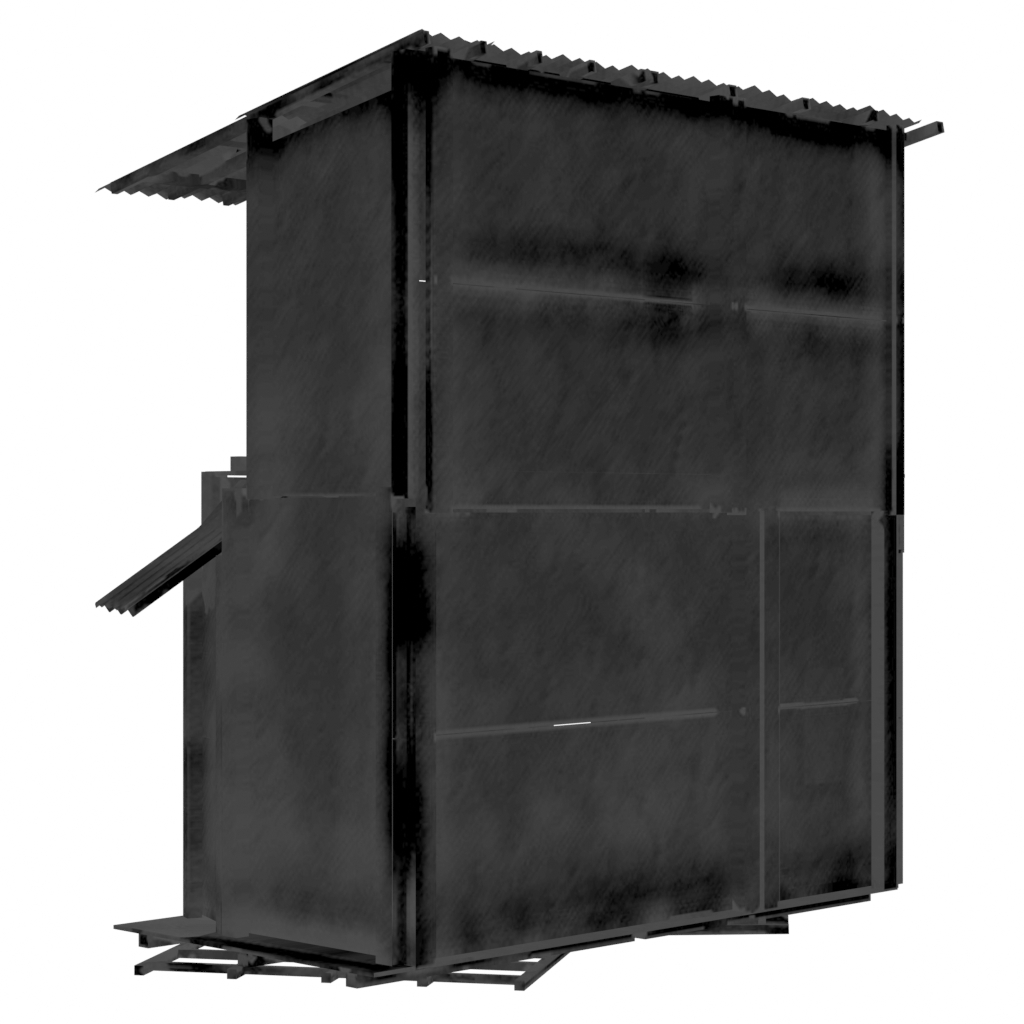}} \\
        & & 
        \rotatebox[origin=c]{90}{VideoMat (T)} &
        \raisebox{-0.5\height}{\includegraphics[width=0.093\textwidth]{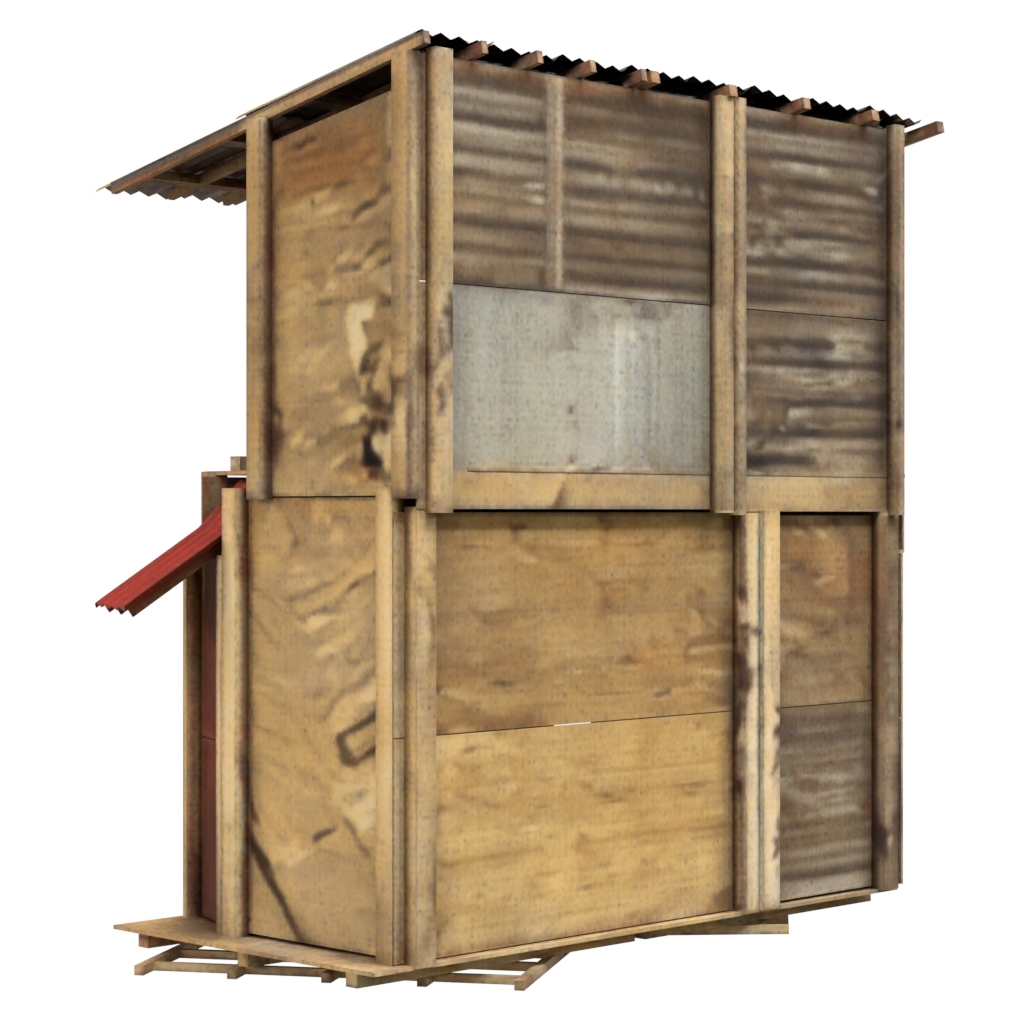}} &
        \raisebox{-0.5\height}{\includegraphics[width=0.093\textwidth]{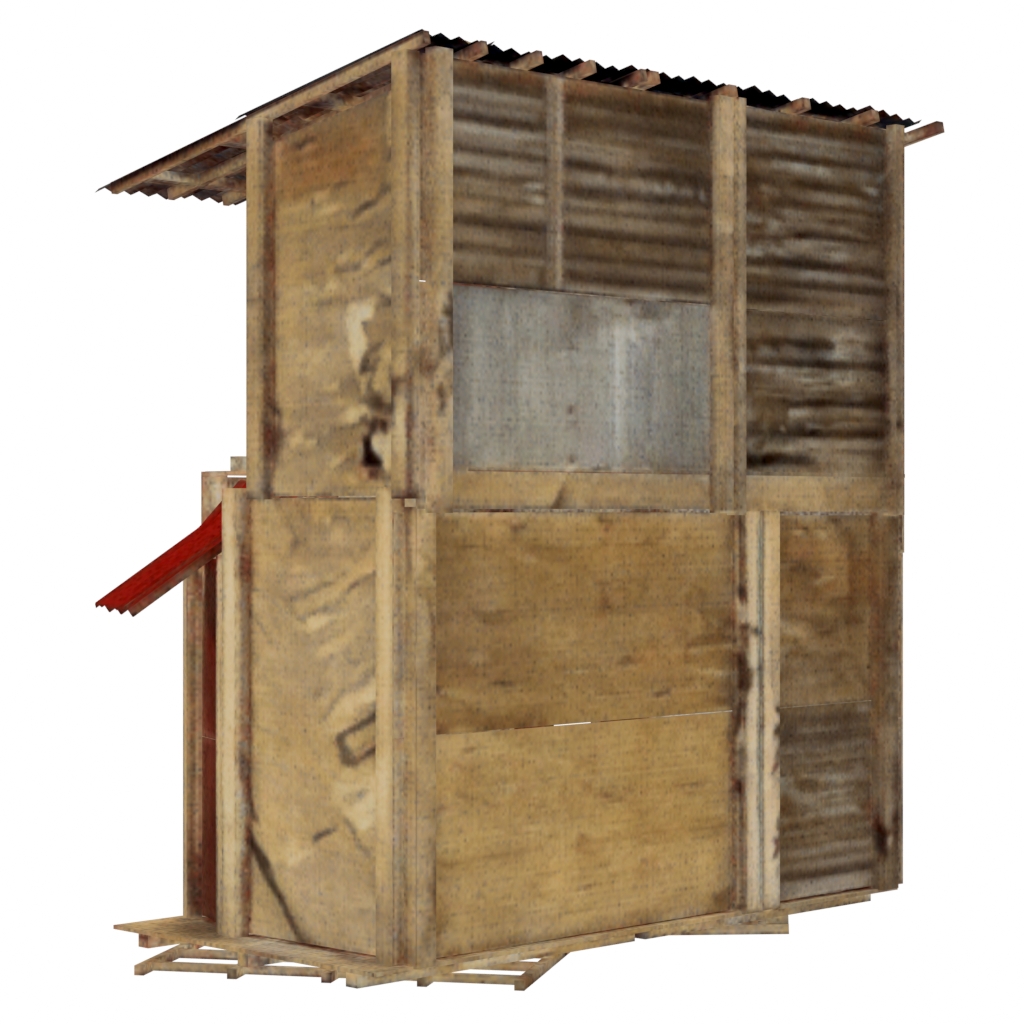}} &
        \raisebox{-0.5\height}{\includegraphics[width=0.093\textwidth]{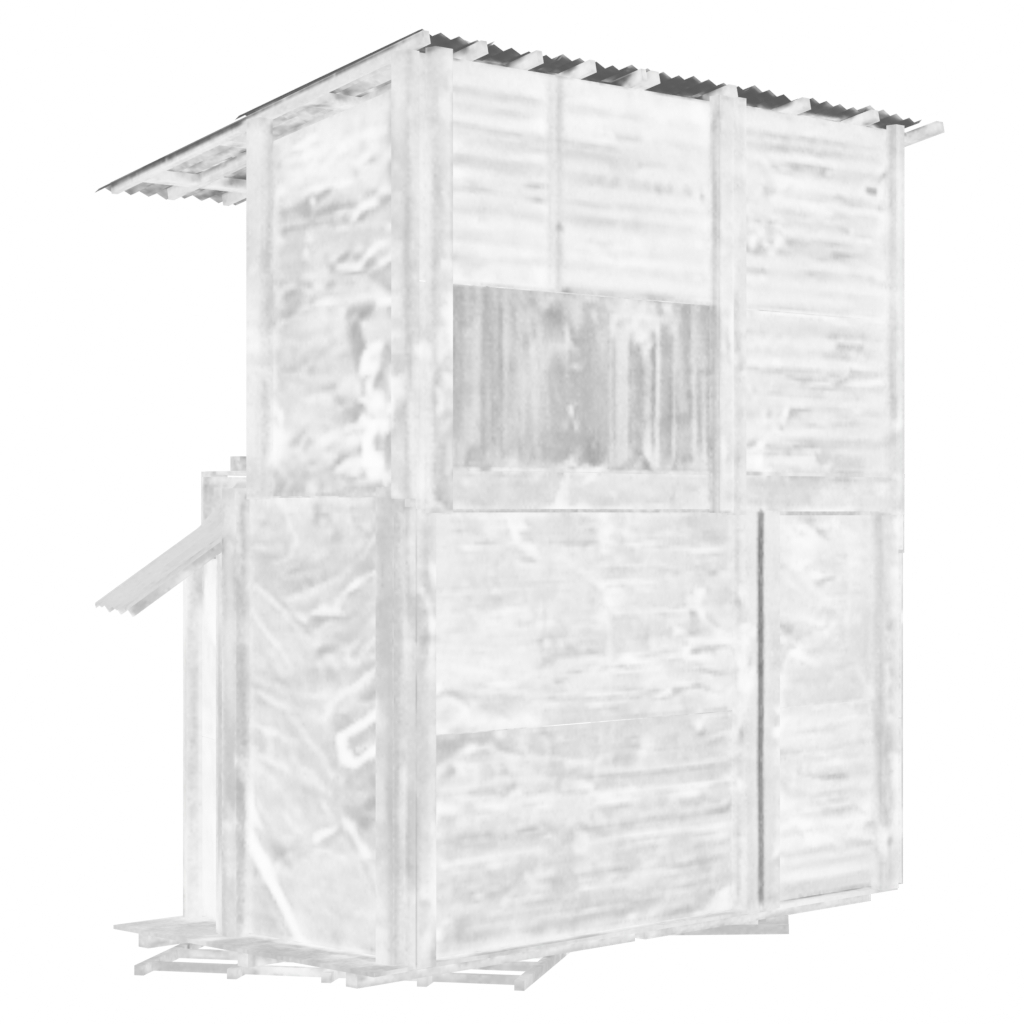}} &
        \raisebox{-0.5\height}{\includegraphics[width=0.093\textwidth]{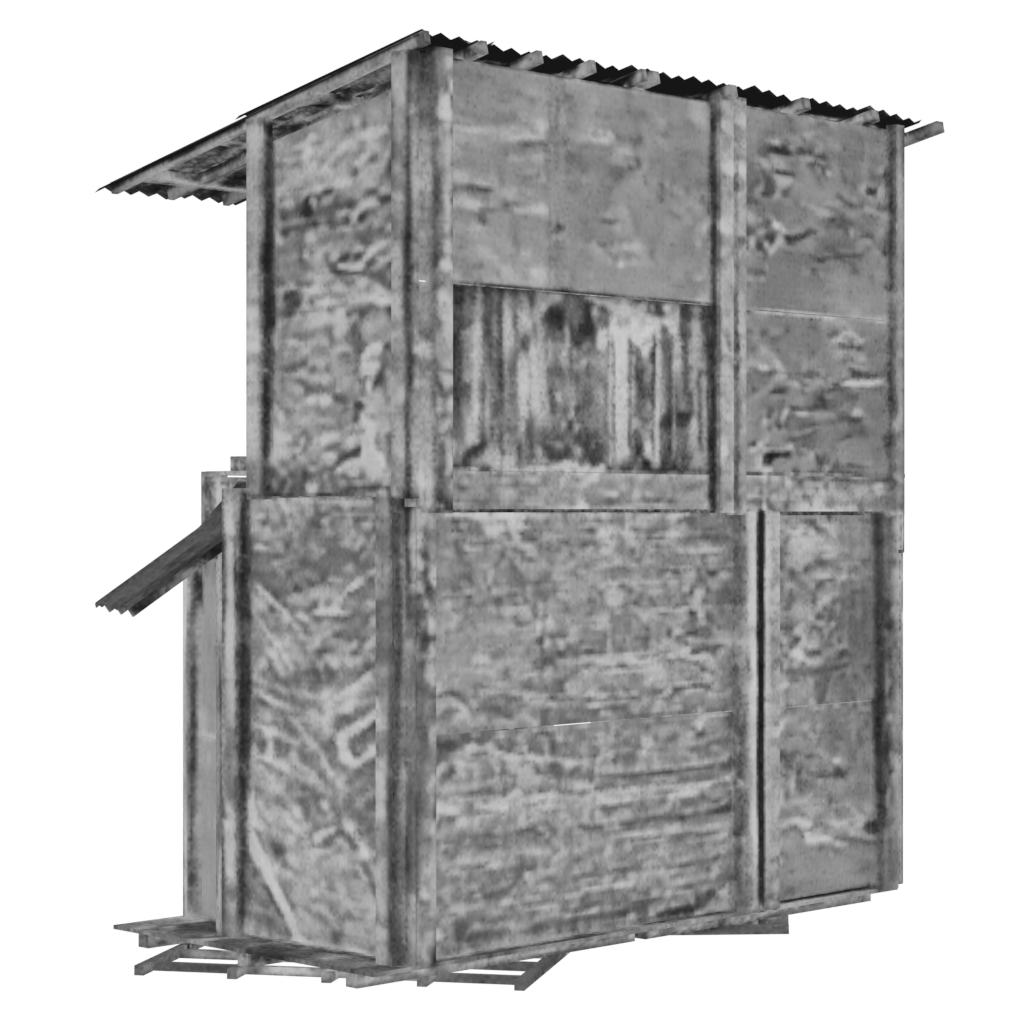}} &
        \rotatebox[origin=c]{90}{\bf Ours (T)} &
        \raisebox{-0.5\height}{\includegraphics[width=0.093\textwidth]{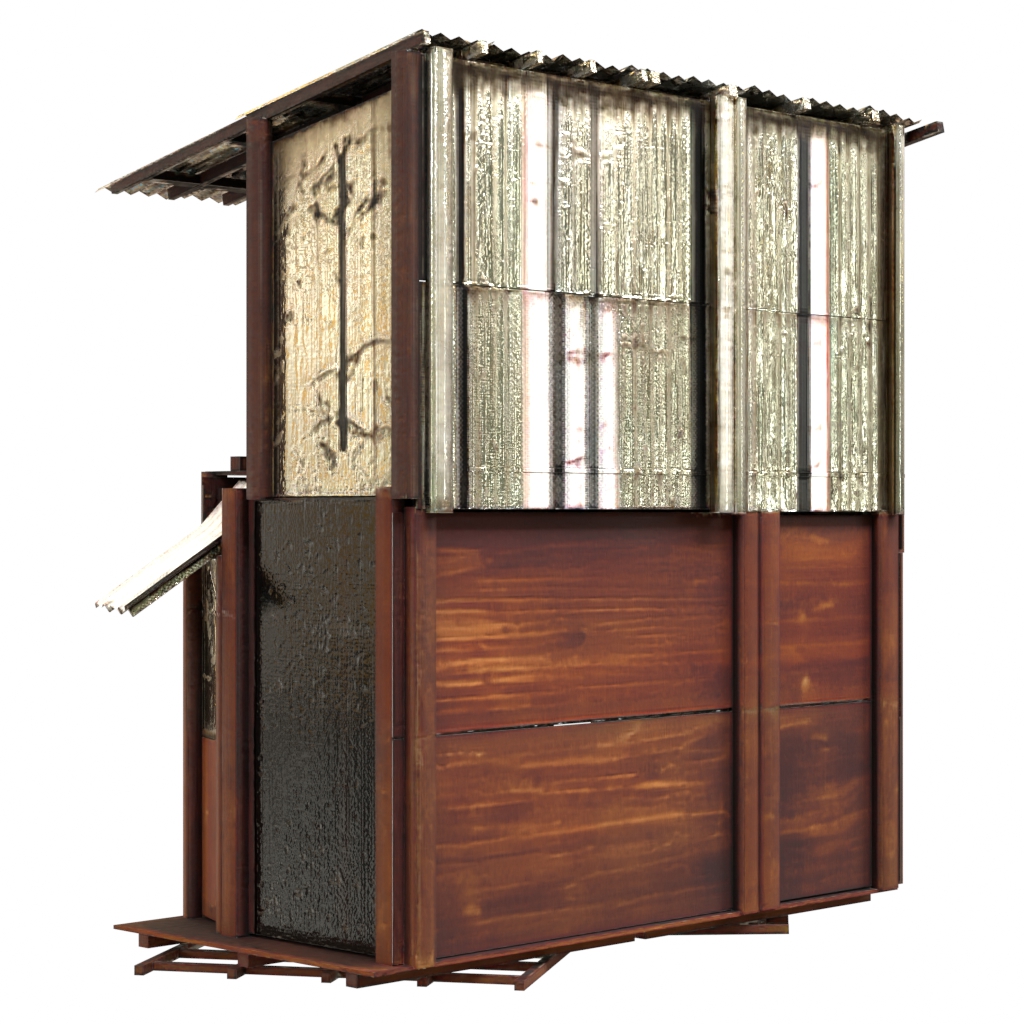}} &
        \raisebox{-0.5\height}{\includegraphics[width=0.093\textwidth]{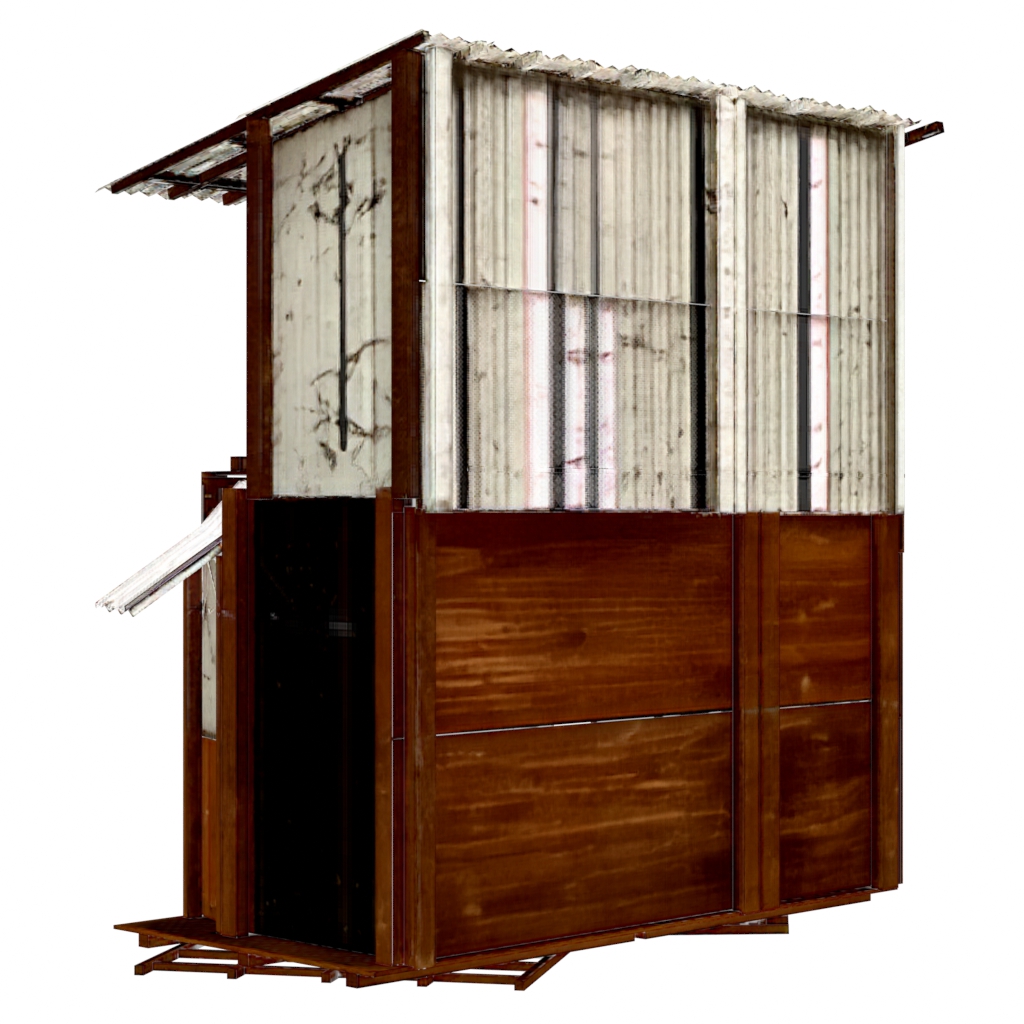}} &
        \raisebox{-0.5\height}{\includegraphics[width=0.093\textwidth]{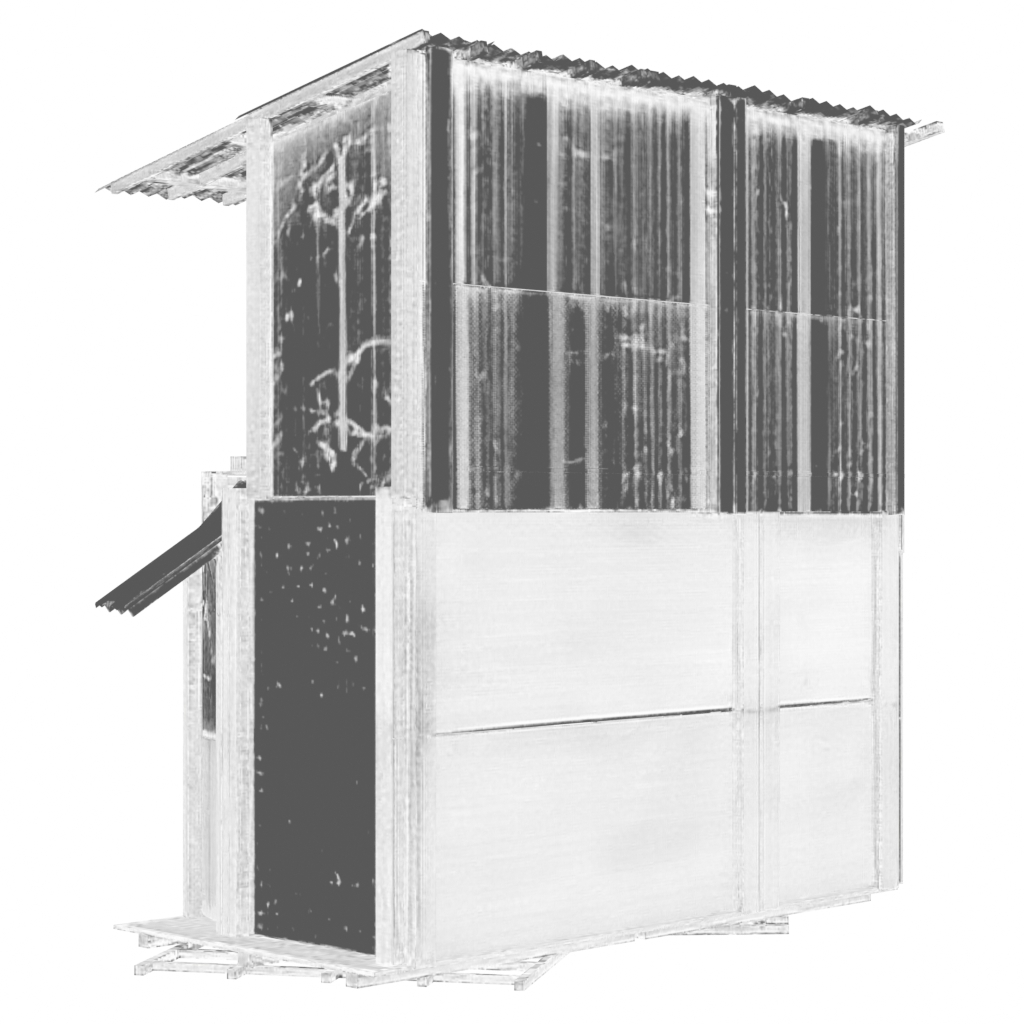}} &
        \raisebox{-0.5\height}{\includegraphics[width=0.093\textwidth]{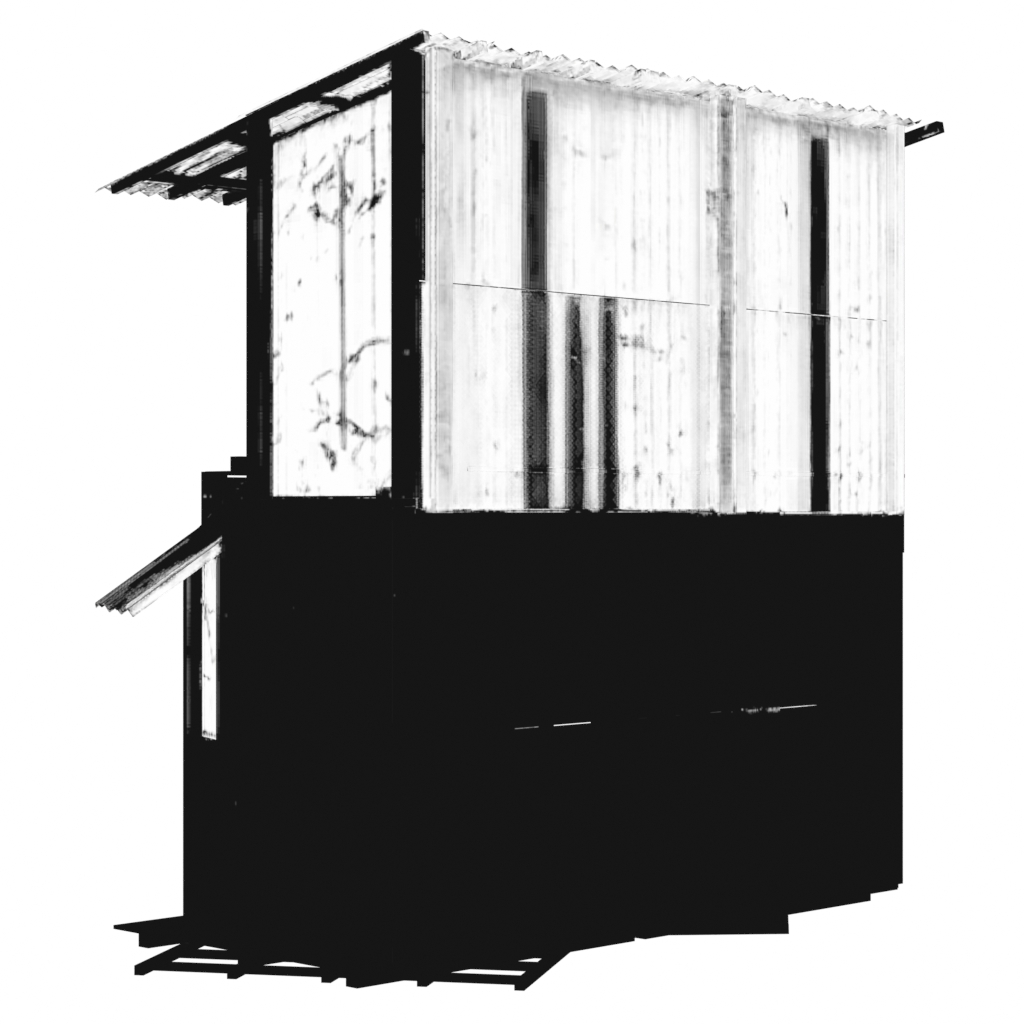}} \\

        & Geometry & & Relit & Base color & Roughness & Metallicity & & Relit & Base color & Roughness & Metallicity
    \end{tabular}
    \vspace*{-2mm}
    \caption{
        Material generation. We compare against Hunyuan3D Paint~2.1~\cite{he2025materialmvp} (image and text guided versions) and VideoMat~\cite{munkberg2025videomat} (text) on three example meshes from the BlenderVault~\cite{litman2025materialfusion} dataset. We encourage the reader to zoom in and compare the quality of the intrinsics (base color, roughness, metallicity), as well as to see the supplementary materials.
    }
    \label{fig:main_quality_results}
\end{figure*}
}


\newcommand{\figBumpmap}{
\begin{figure}
    \centering
    \small
    \includegraphics[width=0.99\columnwidth]{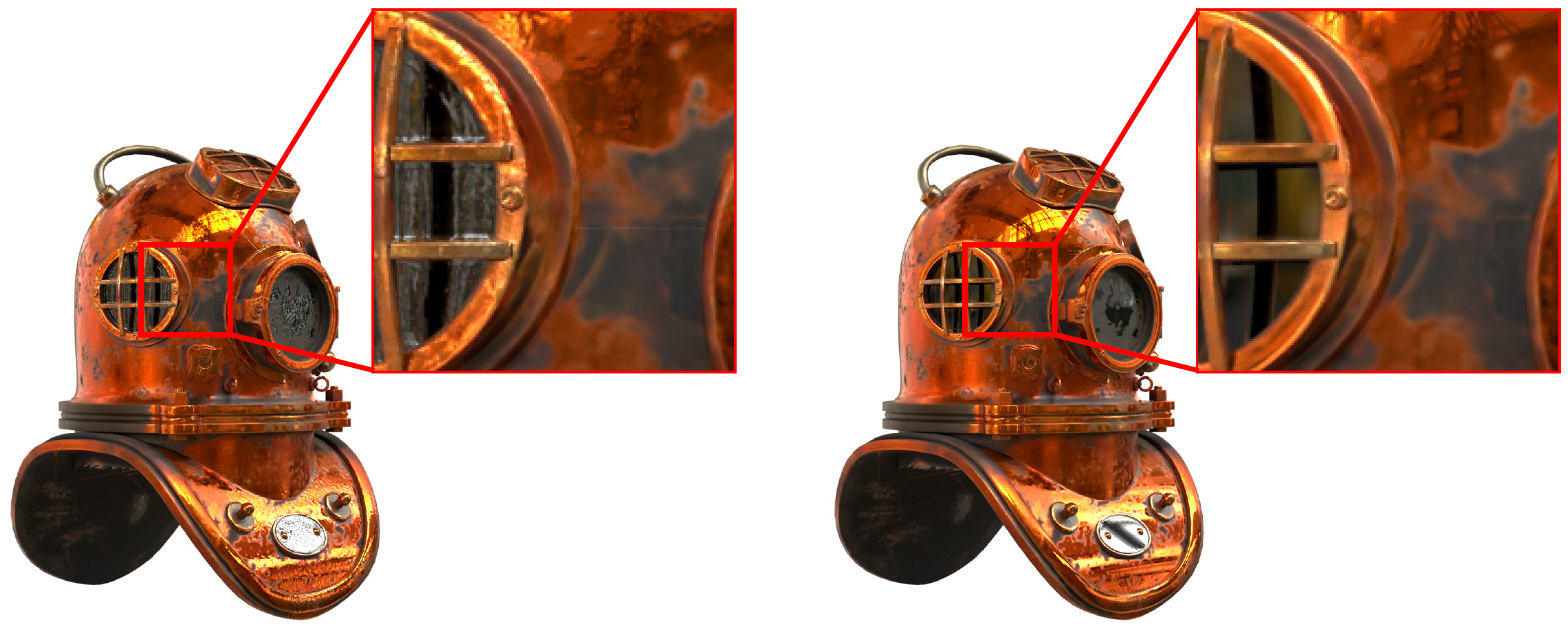}
    \vspace*{-2mm}
    \caption{
        \textbf{Left}: Our method predicts a height (bump) map, which improves the visual richness of the generated material. \textbf{Right}: corresponding rendering without bump map.
    }
    \label{fig:bump}
\end{figure}
}


\newcommand{\figFrameConcatSup}{
\begin{figure*}
\centering
\small
\setlength{\tabcolsep}{1pt}
\begin{tabular}{cccccc}
    & Relit & Base color & Roughness & Metallic & Normals \\

    \rotatebox[origin=c]{90}{Our} &
	\raisebox{-0.5\height}{\includegraphics[width=0.19\textwidth]{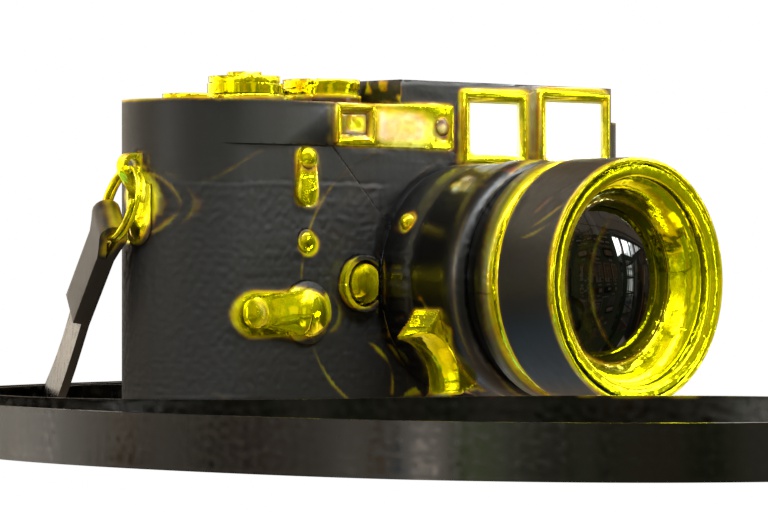}} &
	\raisebox{-0.5\height}{\includegraphics[width=0.19\textwidth]{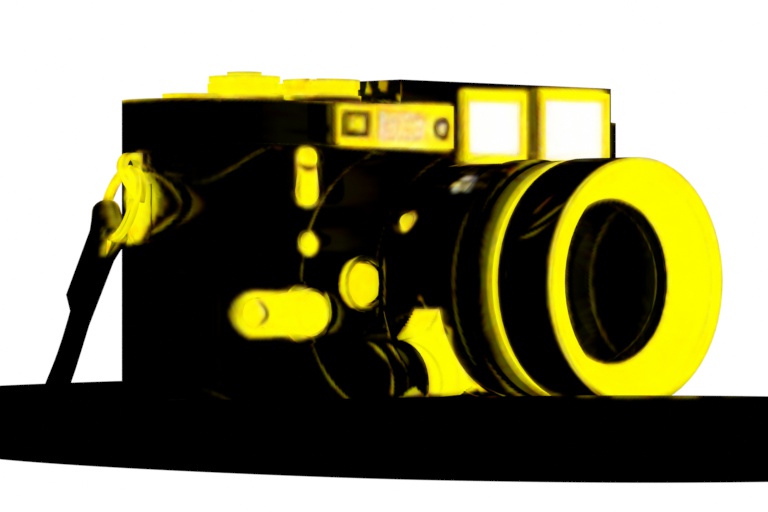}} &
	\raisebox{-0.5\height}{\includegraphics[width=0.19\textwidth]{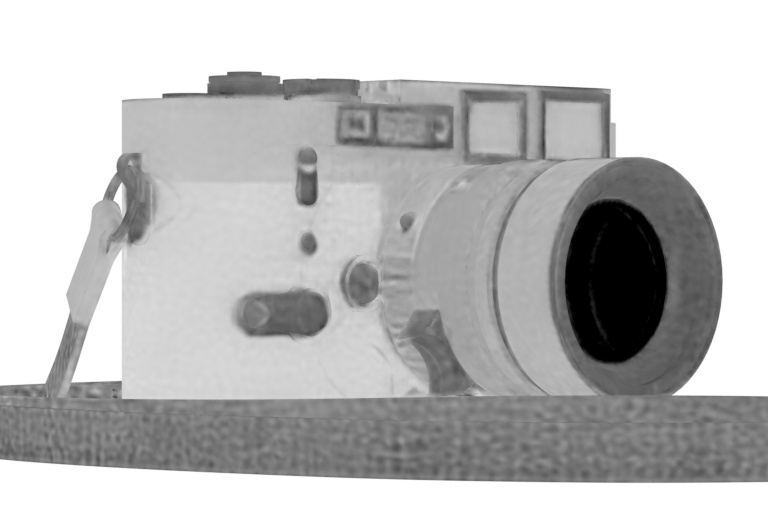}} &
	\raisebox{-0.5\height}{\includegraphics[width=0.19\textwidth]{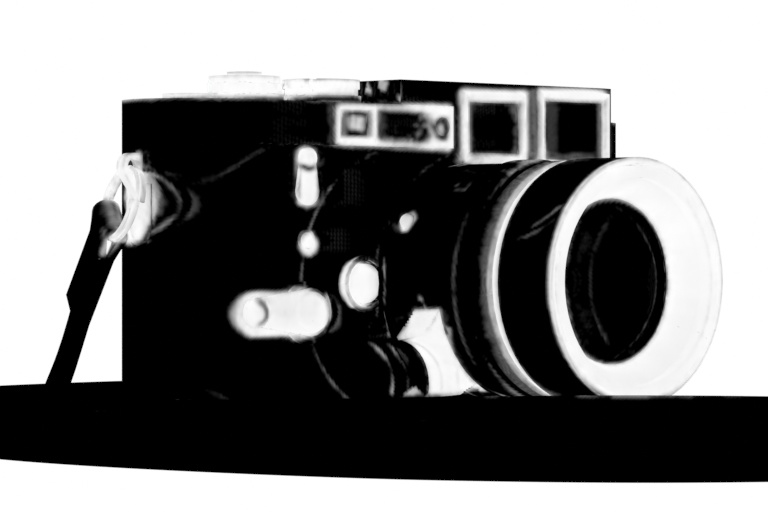}} &
    \raisebox{-0.5\height}{\includegraphics[width=0.19\textwidth]{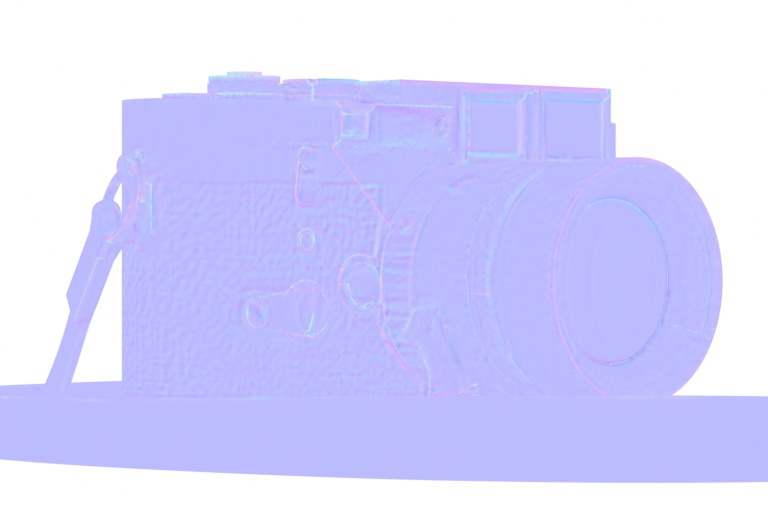}} \\

    \rotatebox[origin=c]{90}{FCat} &
	\raisebox{-0.5\height}{\includegraphics[width=0.19\textwidth]{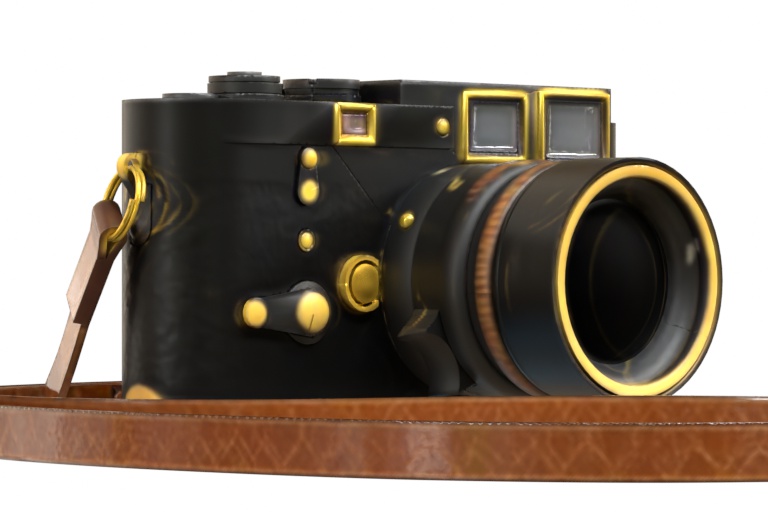}} &
	\raisebox{-0.5\height}{\includegraphics[width=0.19\textwidth]{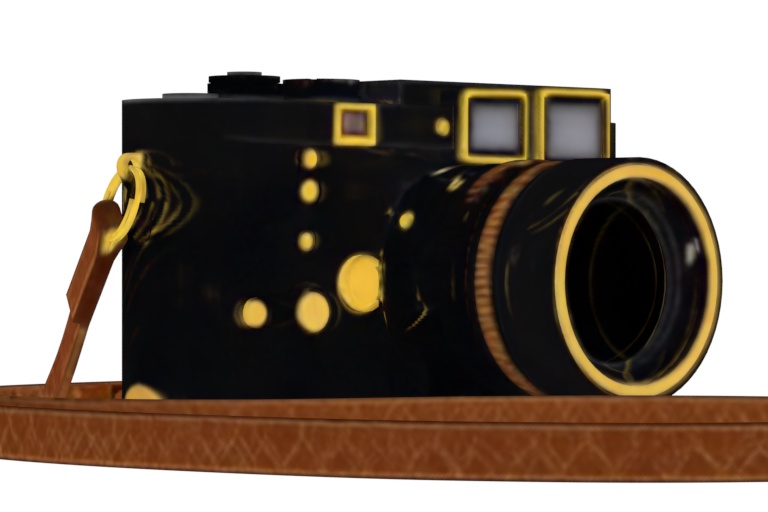}} &
	\raisebox{-0.5\height}{\includegraphics[width=0.19\textwidth]{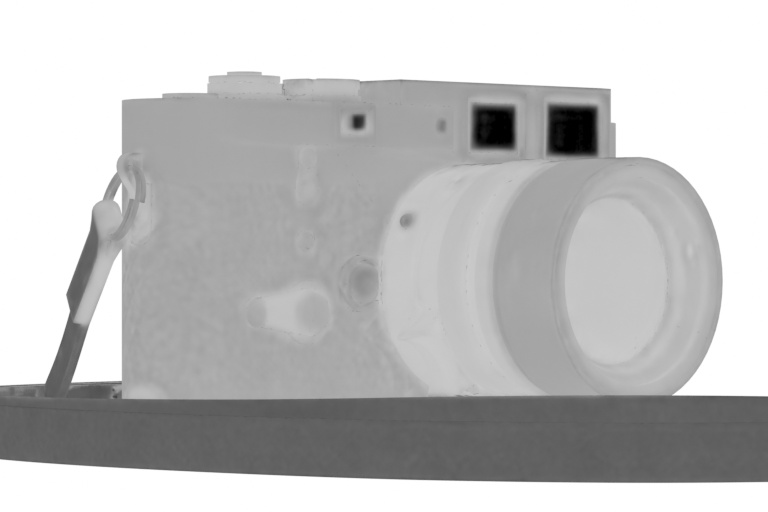}} &
	\raisebox{-0.5\height}{\includegraphics[width=0.19\textwidth]{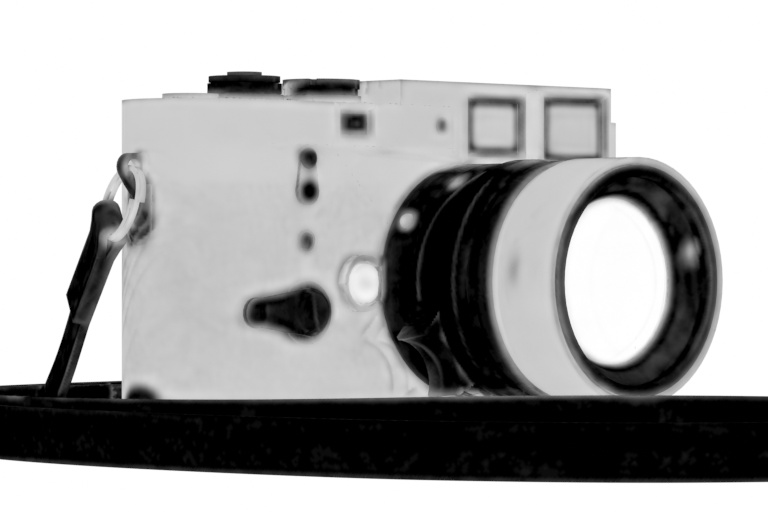}} &
    \raisebox{-0.5\height}{\includegraphics[width=0.19\textwidth]{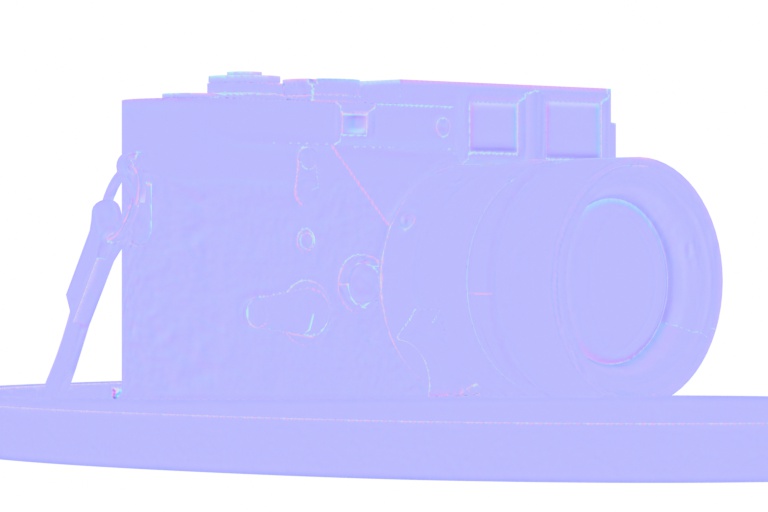}} \\
    
    \rotatebox[origin=c]{90}{Our} &
	\raisebox{-0.5\height}{\includegraphics[width=0.19\textwidth]{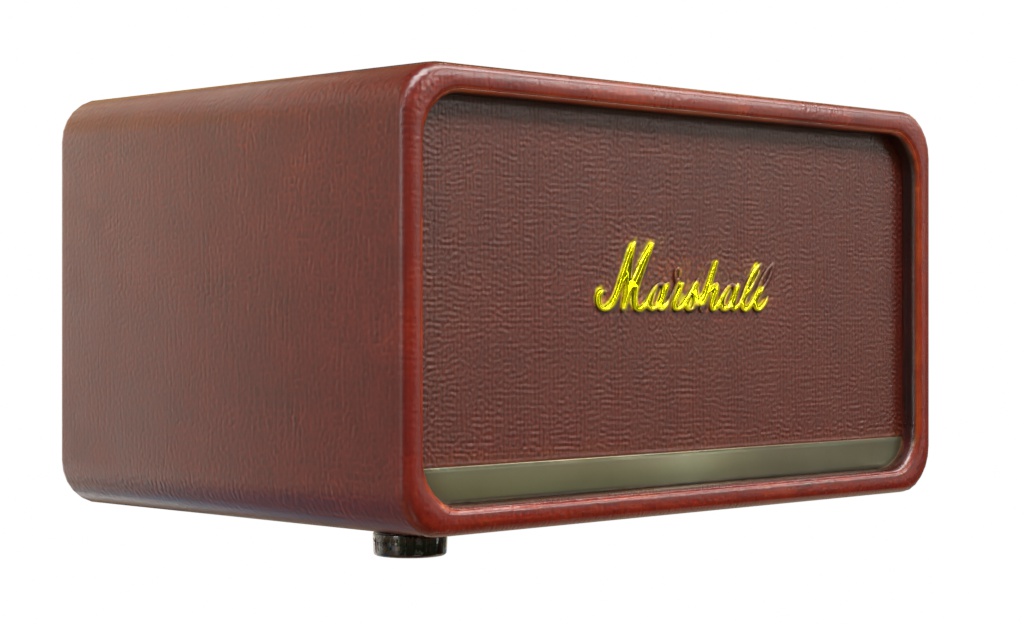}} &
	\raisebox{-0.5\height}{\includegraphics[width=0.19\textwidth]{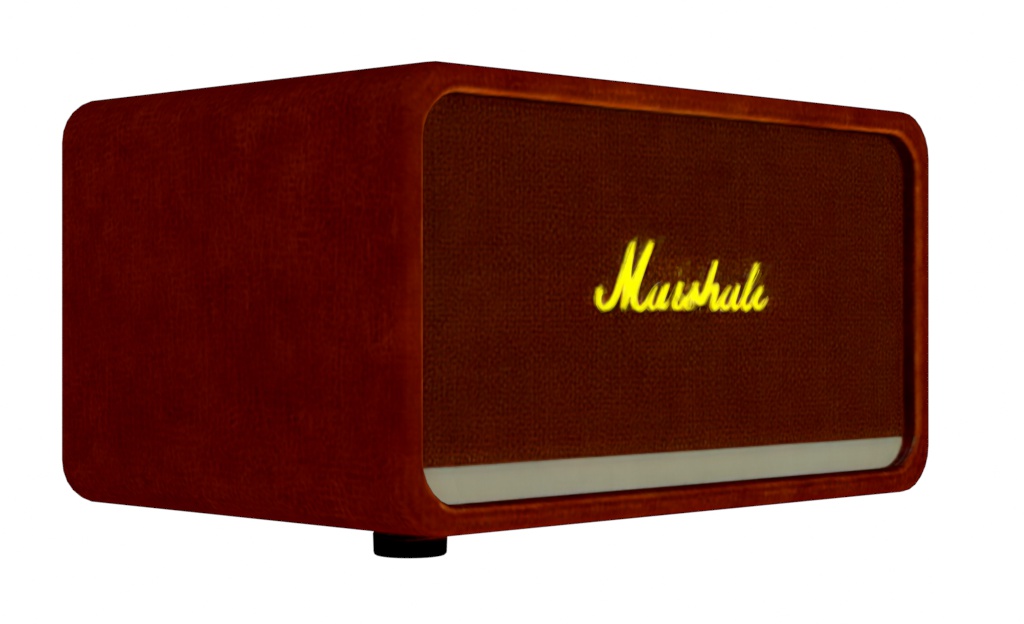}} &
	\raisebox{-0.5\height}{\includegraphics[width=0.19\textwidth]{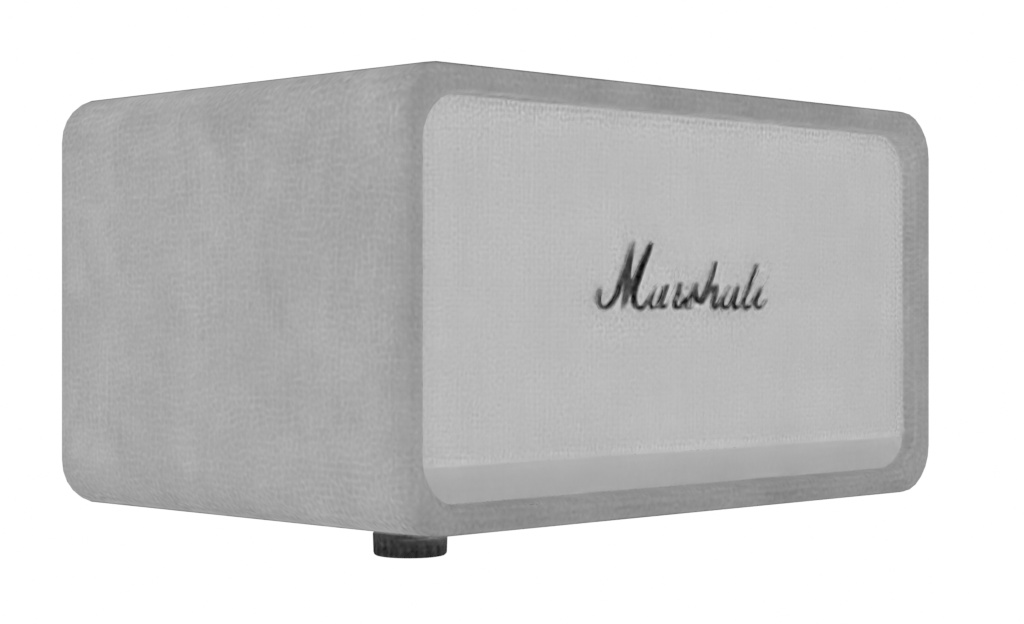}} &
	\raisebox{-0.5\height}{\includegraphics[width=0.19\textwidth]{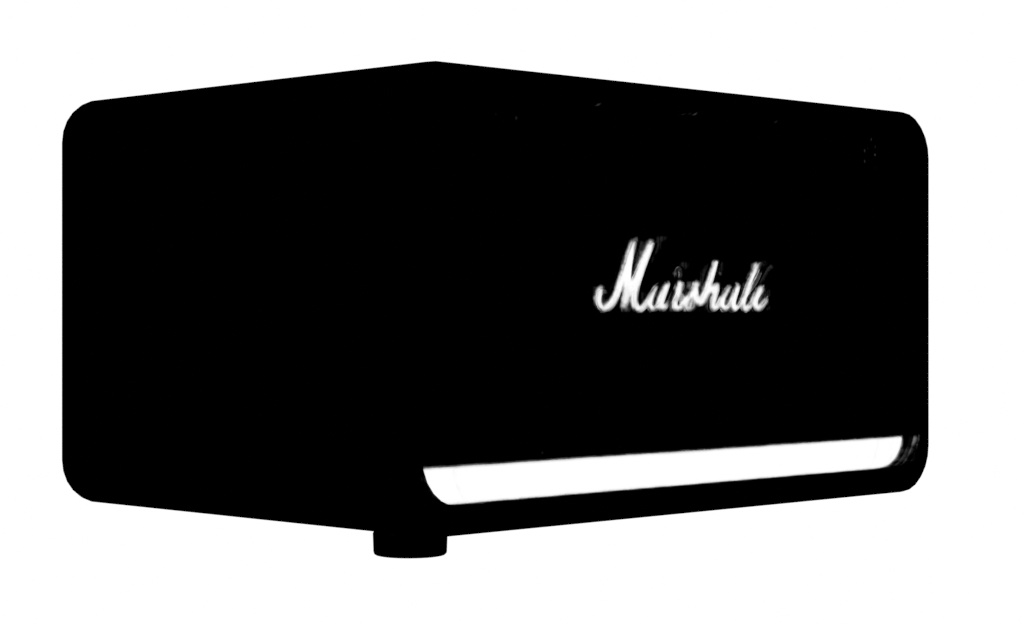}} &
    \raisebox{-0.5\height}{\includegraphics[width=0.19\textwidth]{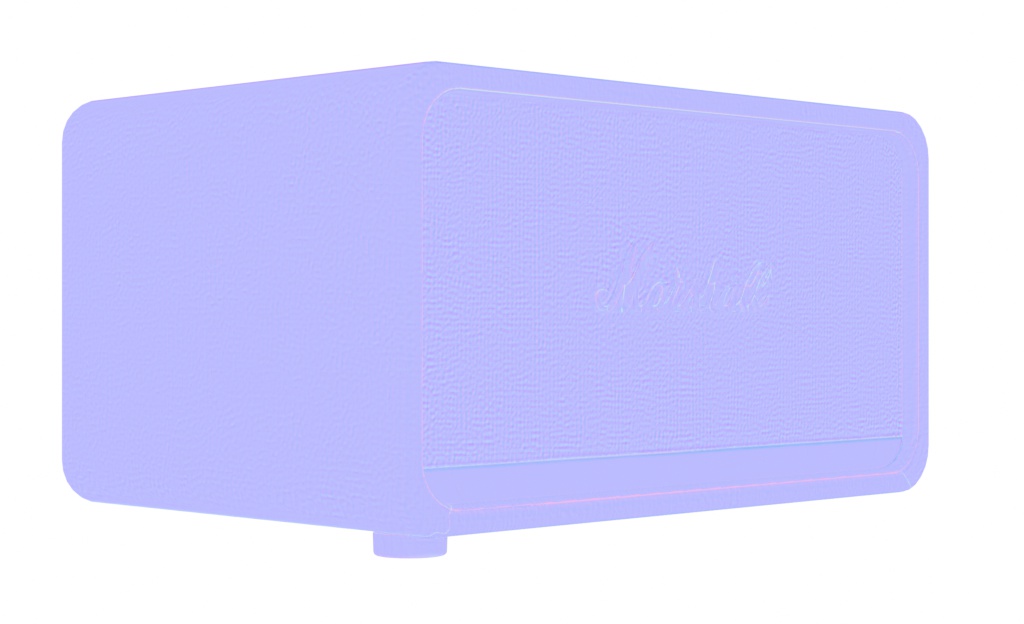}} \\

    \rotatebox[origin=c]{90}{Frame Concat} &
	\raisebox{-0.5\height}{\includegraphics[width=0.19\textwidth]{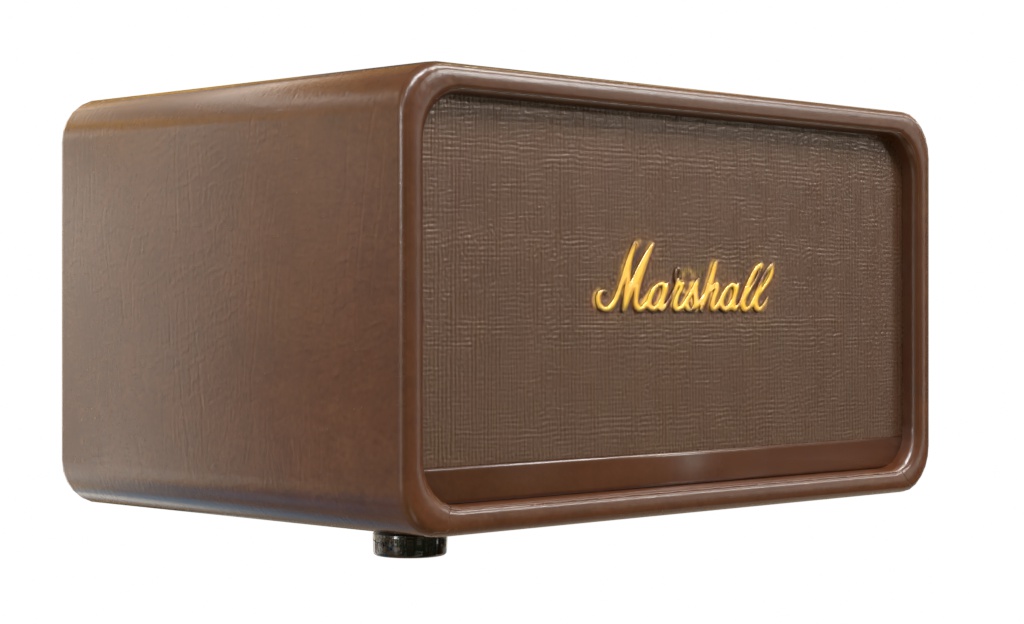}} &
	\raisebox{-0.5\height}{\includegraphics[width=0.19\textwidth]{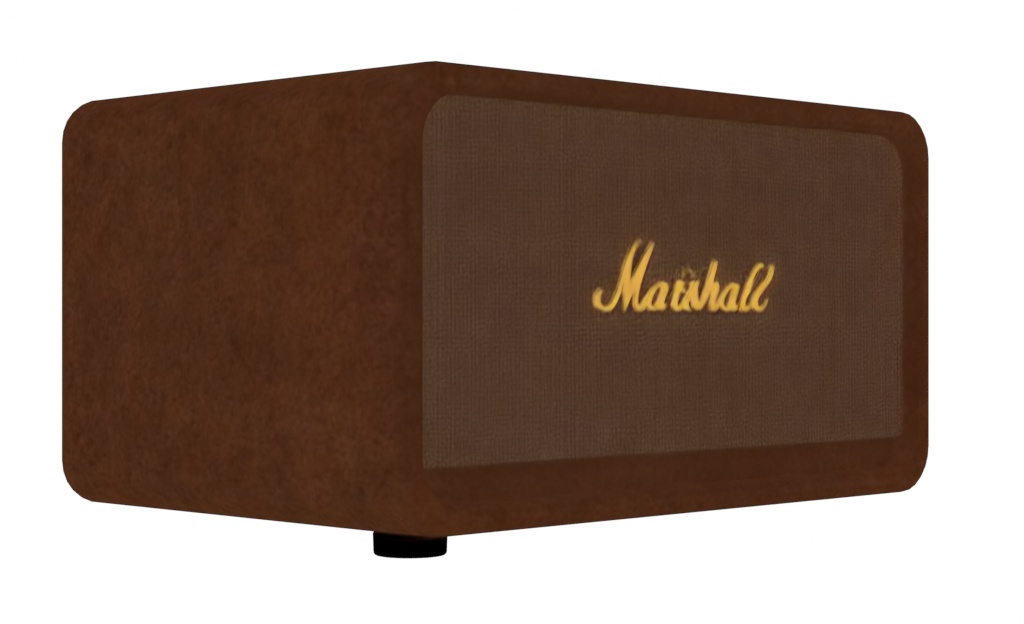}} &
	\raisebox{-0.5\height}{\includegraphics[width=0.19\textwidth]{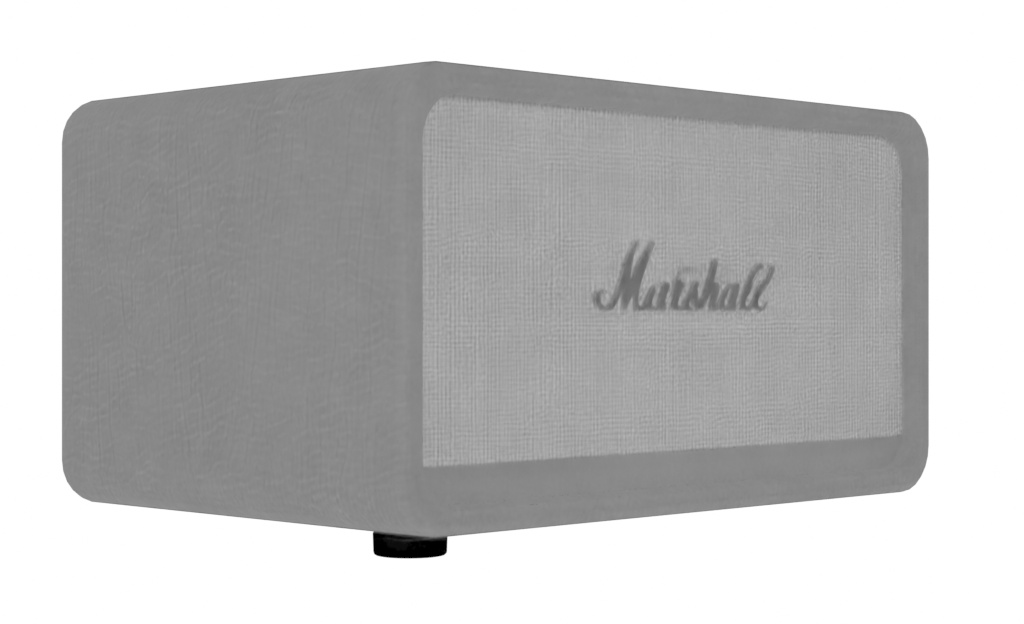}} &
	\raisebox{-0.5\height}{\includegraphics[width=0.19\textwidth]{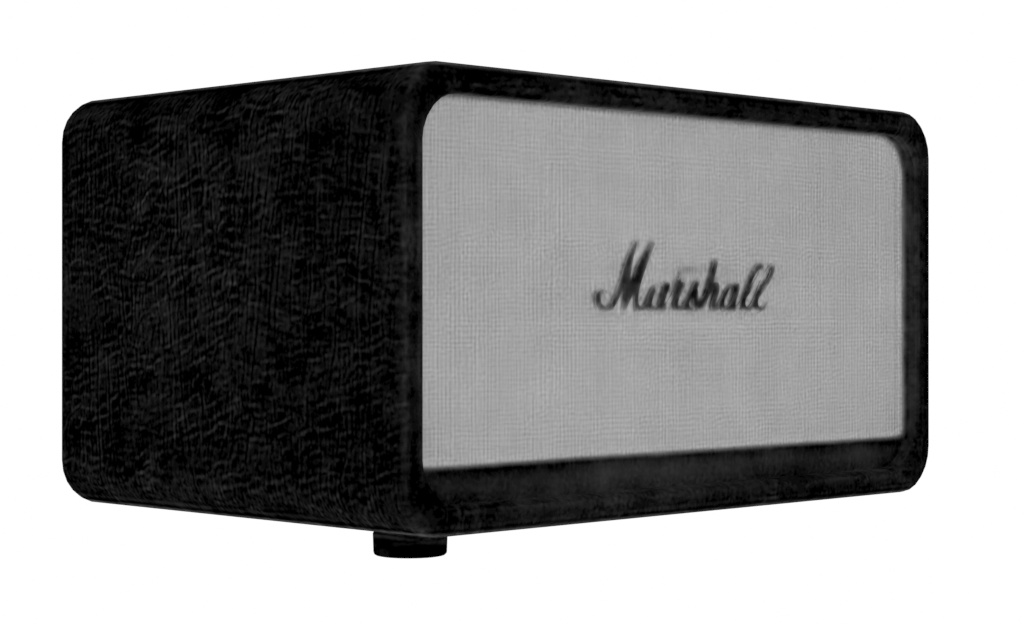}} &
    \raisebox{-0.5\height}{\includegraphics[width=0.19\textwidth]{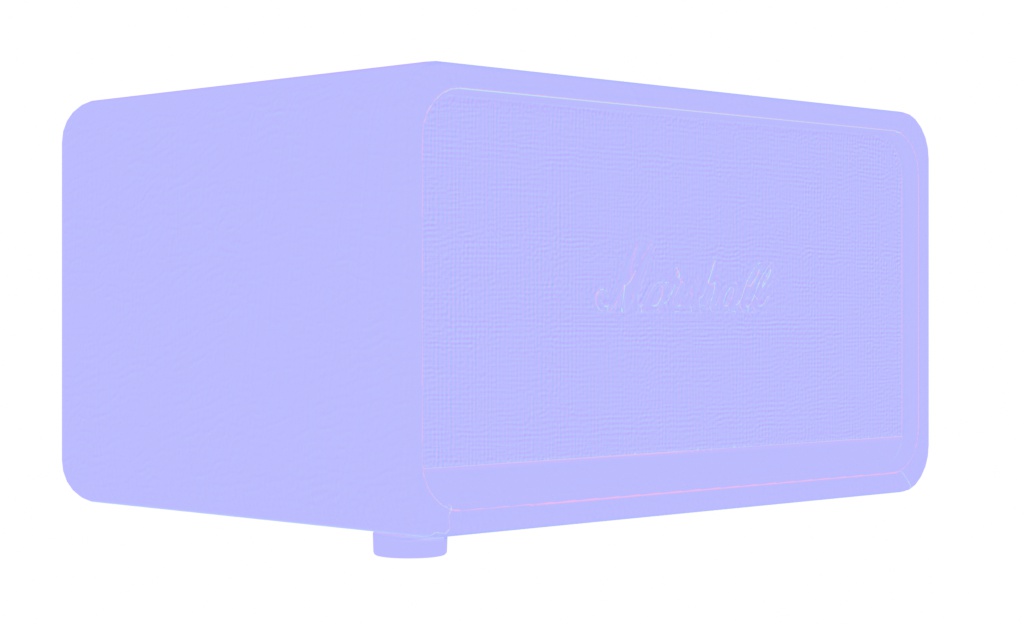}} \\

    \rotatebox[origin=c]{90}{Our} &
	\raisebox{-0.5\height}{\includegraphics[width=0.19\textwidth]{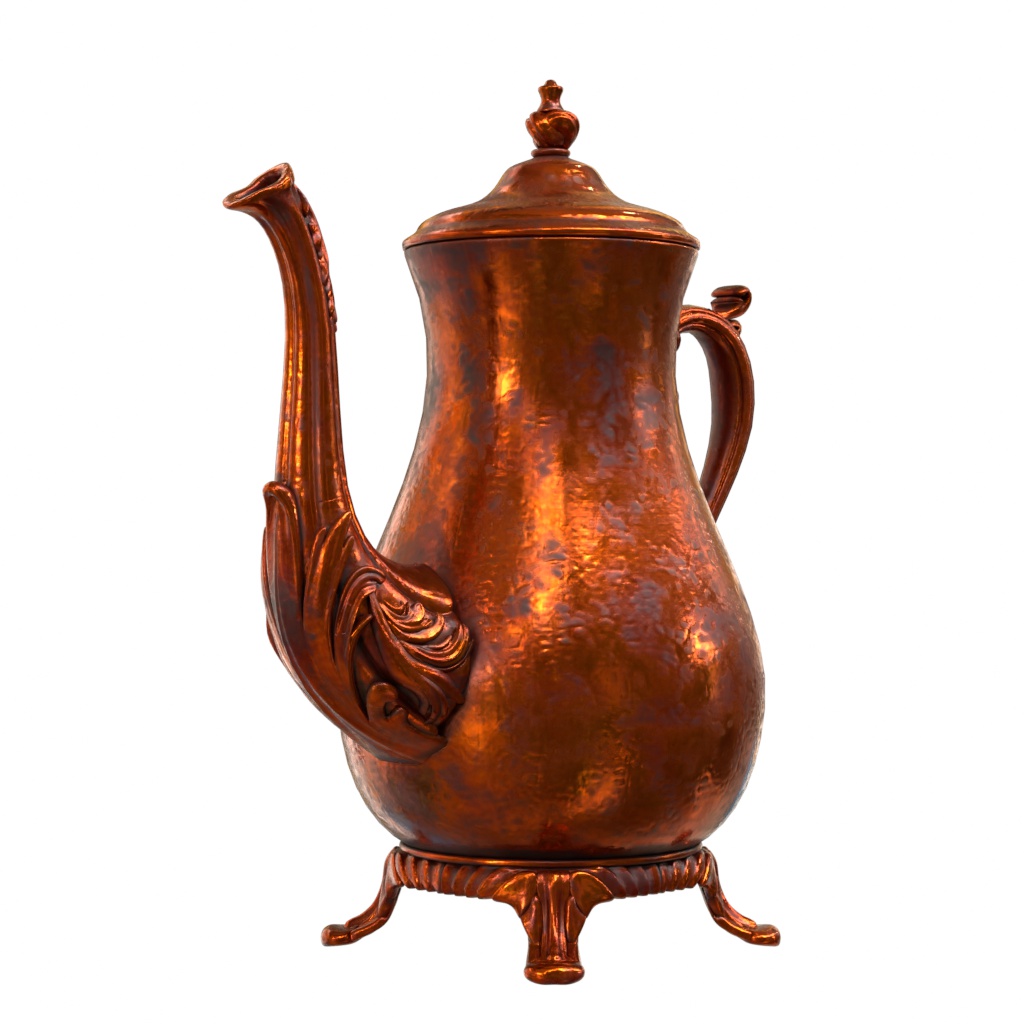}} &
	\raisebox{-0.5\height}{\includegraphics[width=0.19\textwidth]{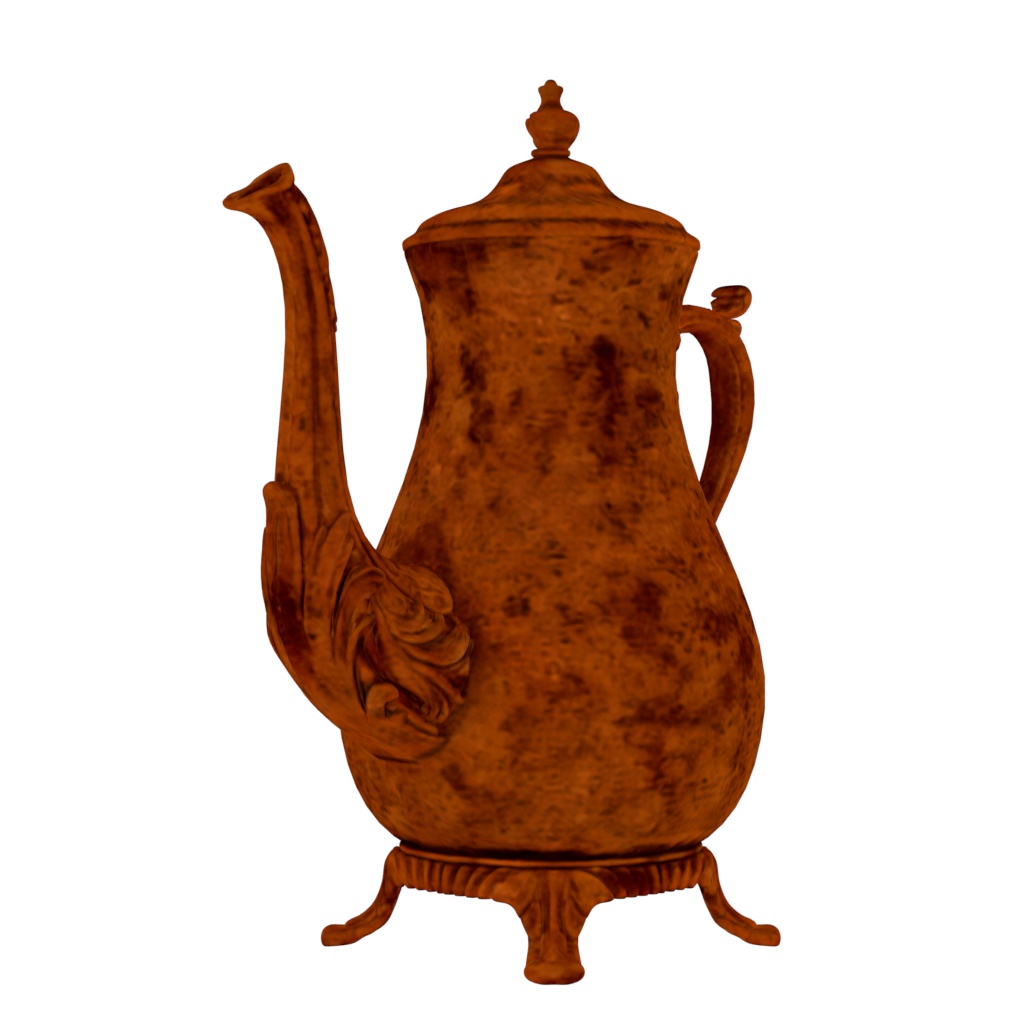}} &
	\raisebox{-0.5\height}{\includegraphics[width=0.19\textwidth]{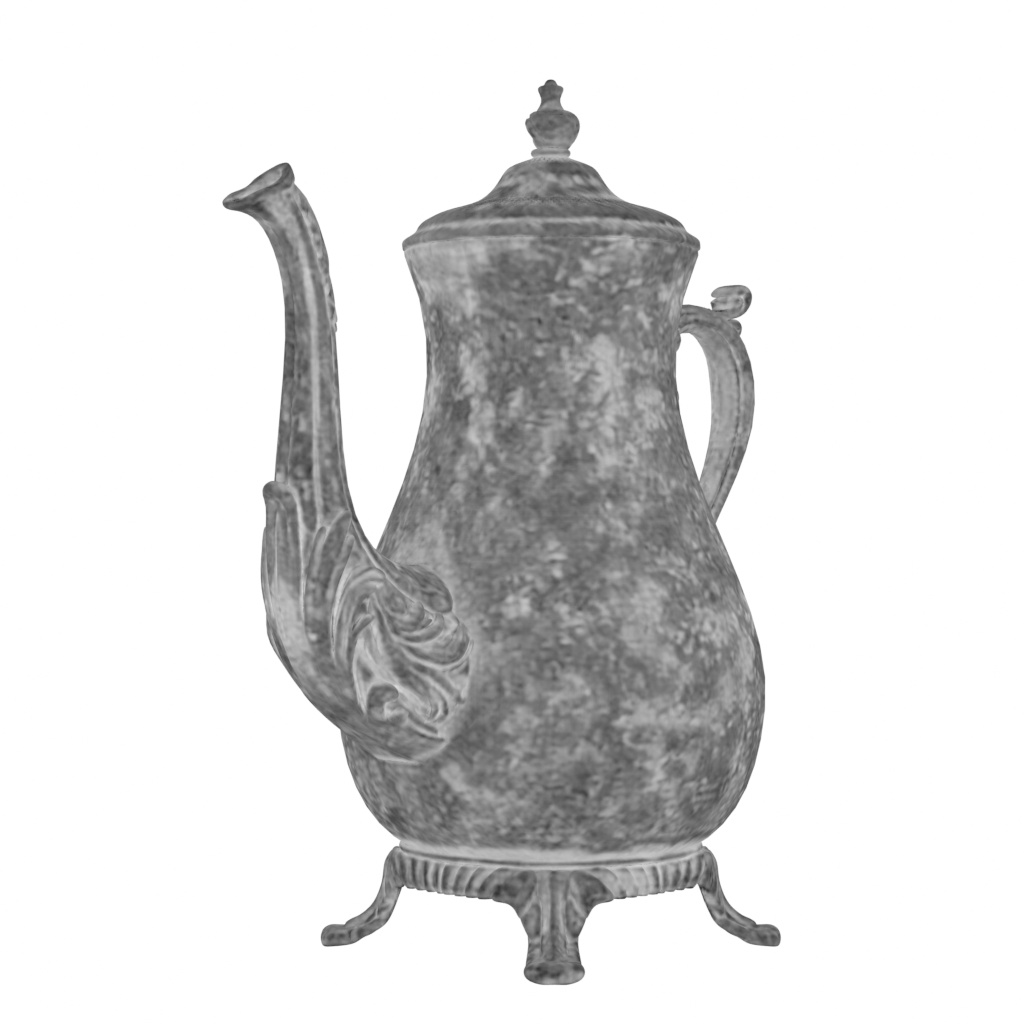}} &
	\raisebox{-0.5\height}{\includegraphics[width=0.19\textwidth]{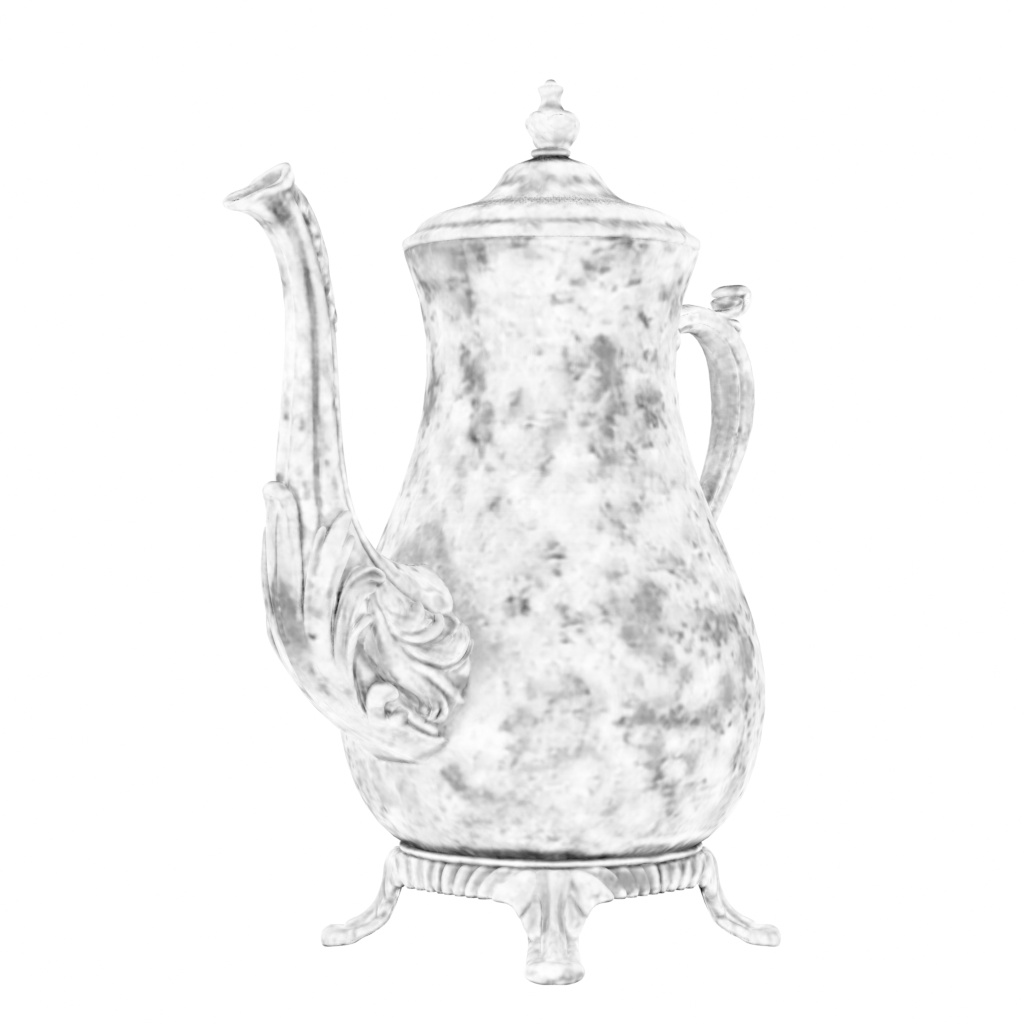}} &
    \raisebox{-0.5\height}{\includegraphics[width=0.19\textwidth]{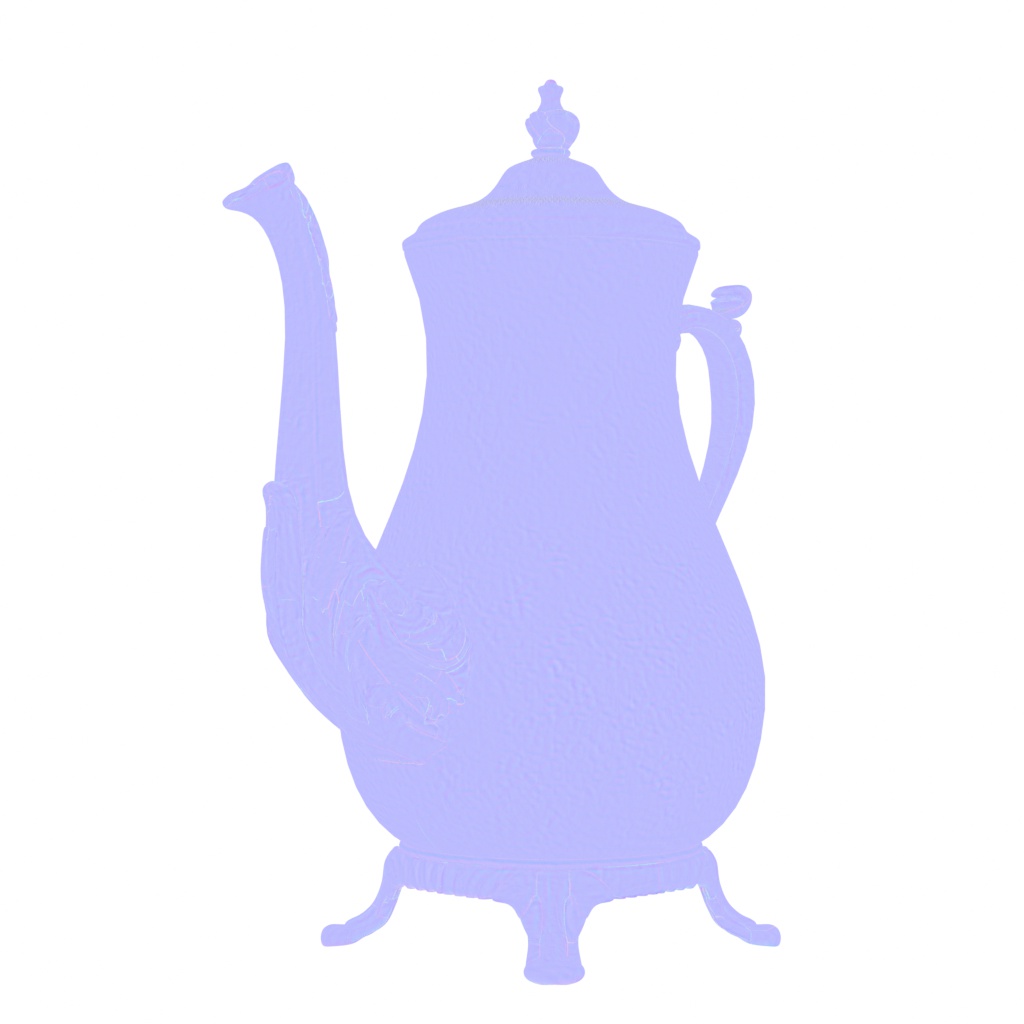}} \\

    \rotatebox[origin=c]{90}{Frame Concat} &
	\raisebox{-0.5\height}{\includegraphics[width=0.19\textwidth]{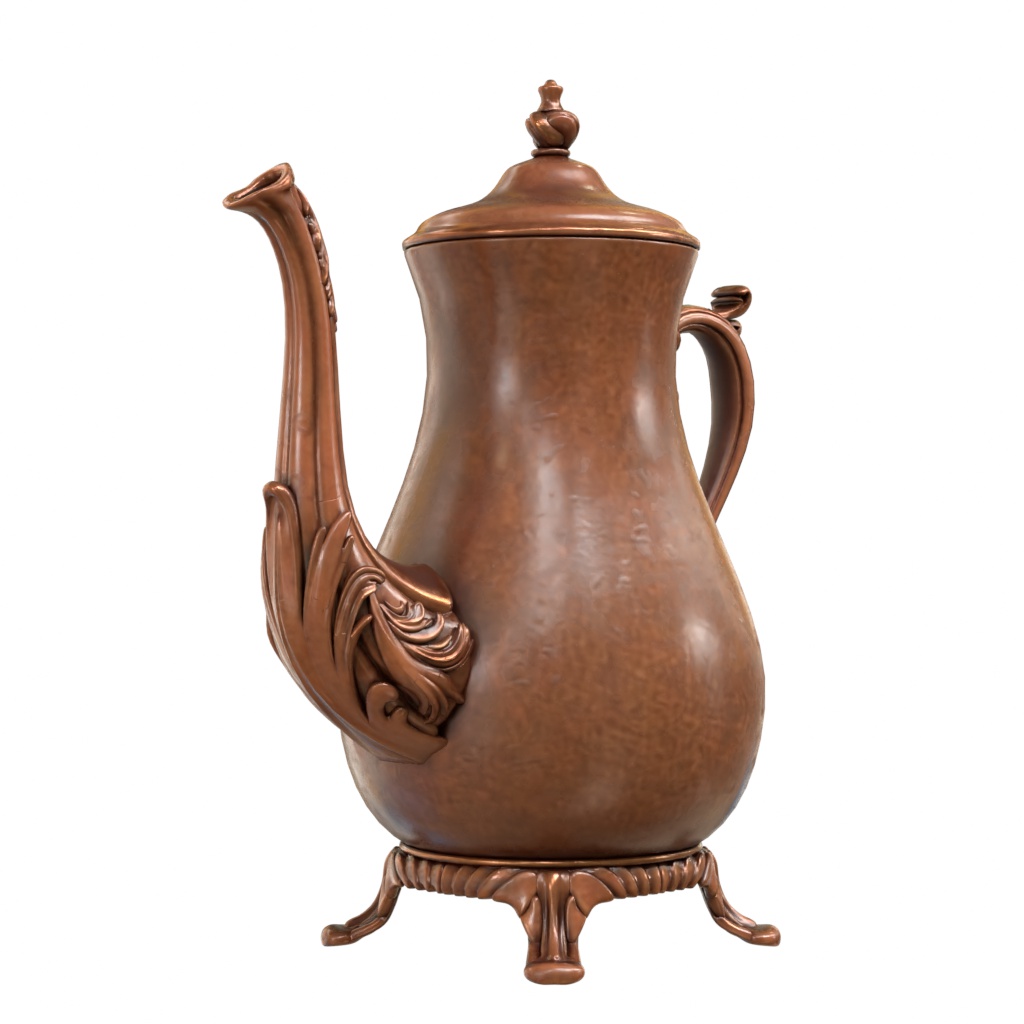}} &
	\raisebox{-0.5\height}{\includegraphics[width=0.19\textwidth]{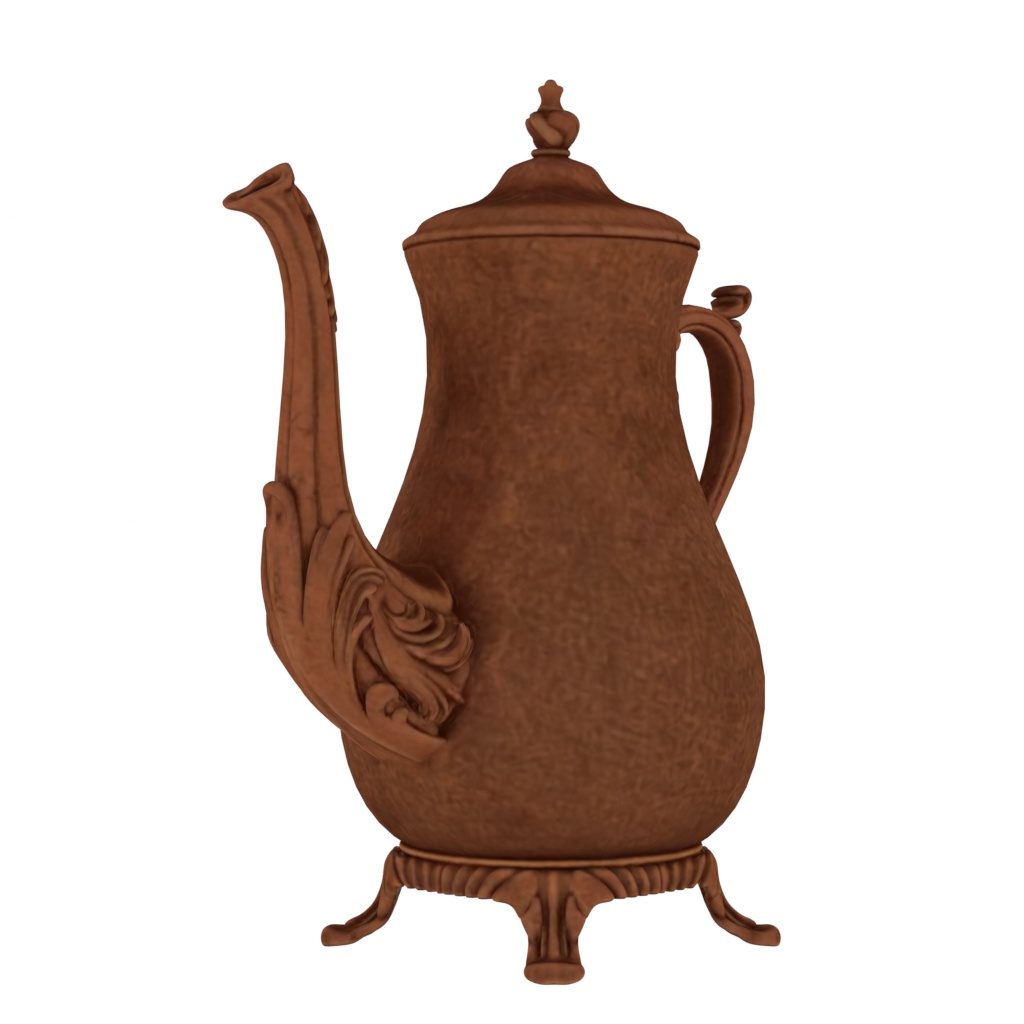}} &
	\raisebox{-0.5\height}{\includegraphics[width=0.19\textwidth]{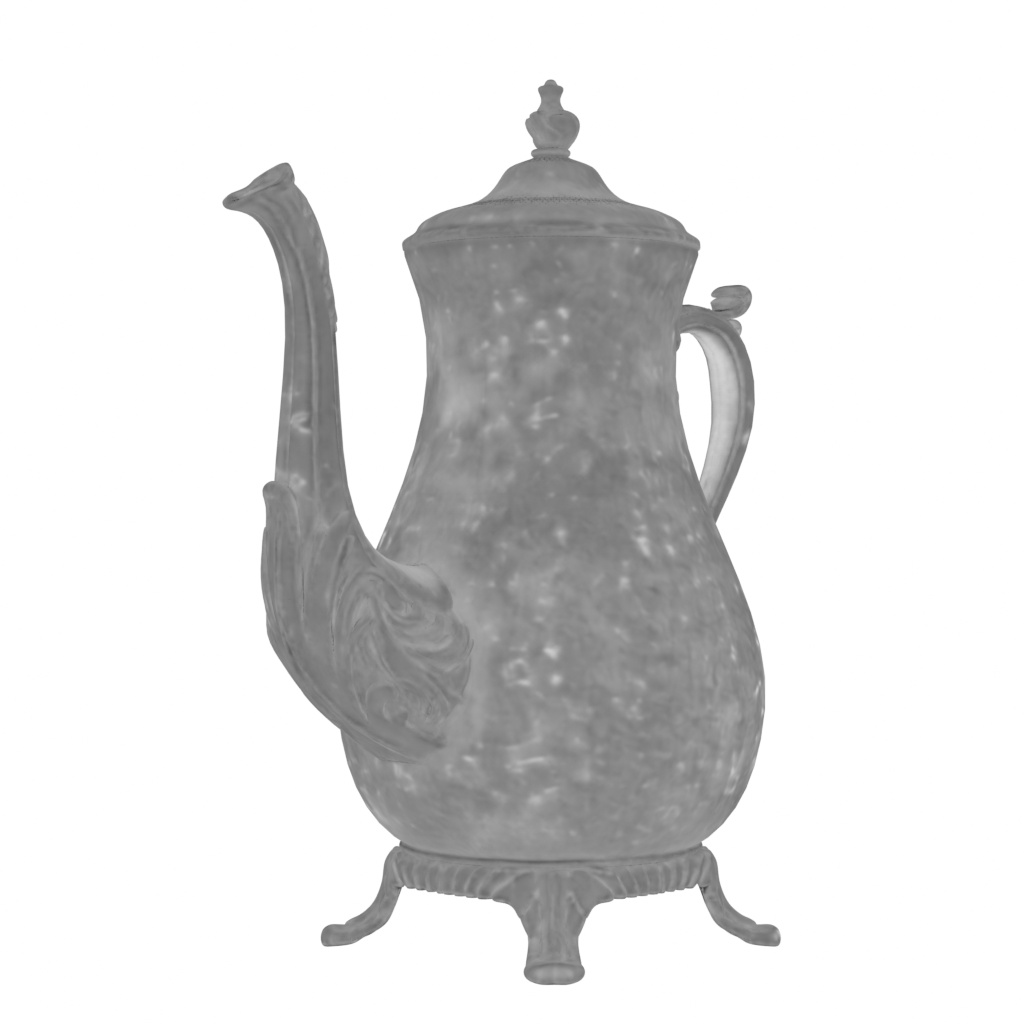}} &
	\raisebox{-0.5\height}{\includegraphics[width=0.19\textwidth]{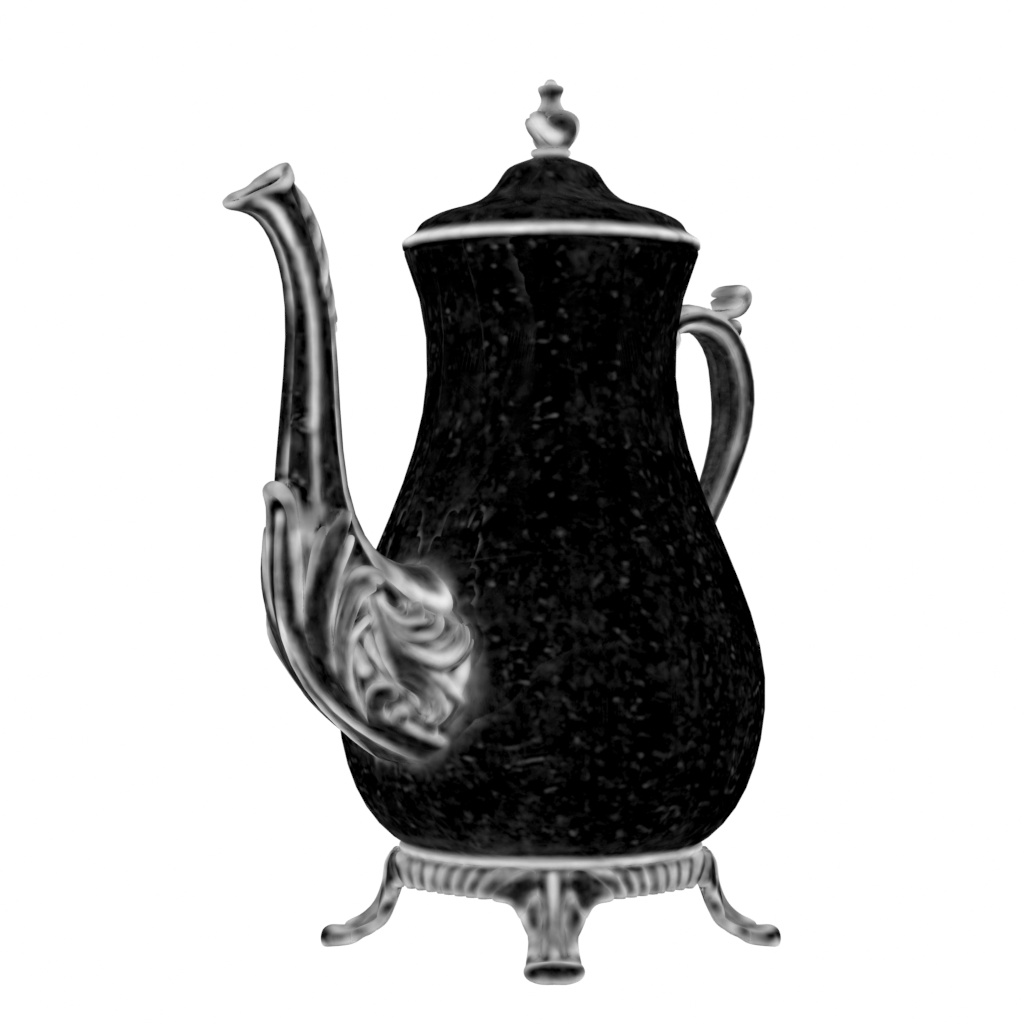}} &
    \raisebox{-0.5\height}{\includegraphics[width=0.19\textwidth]{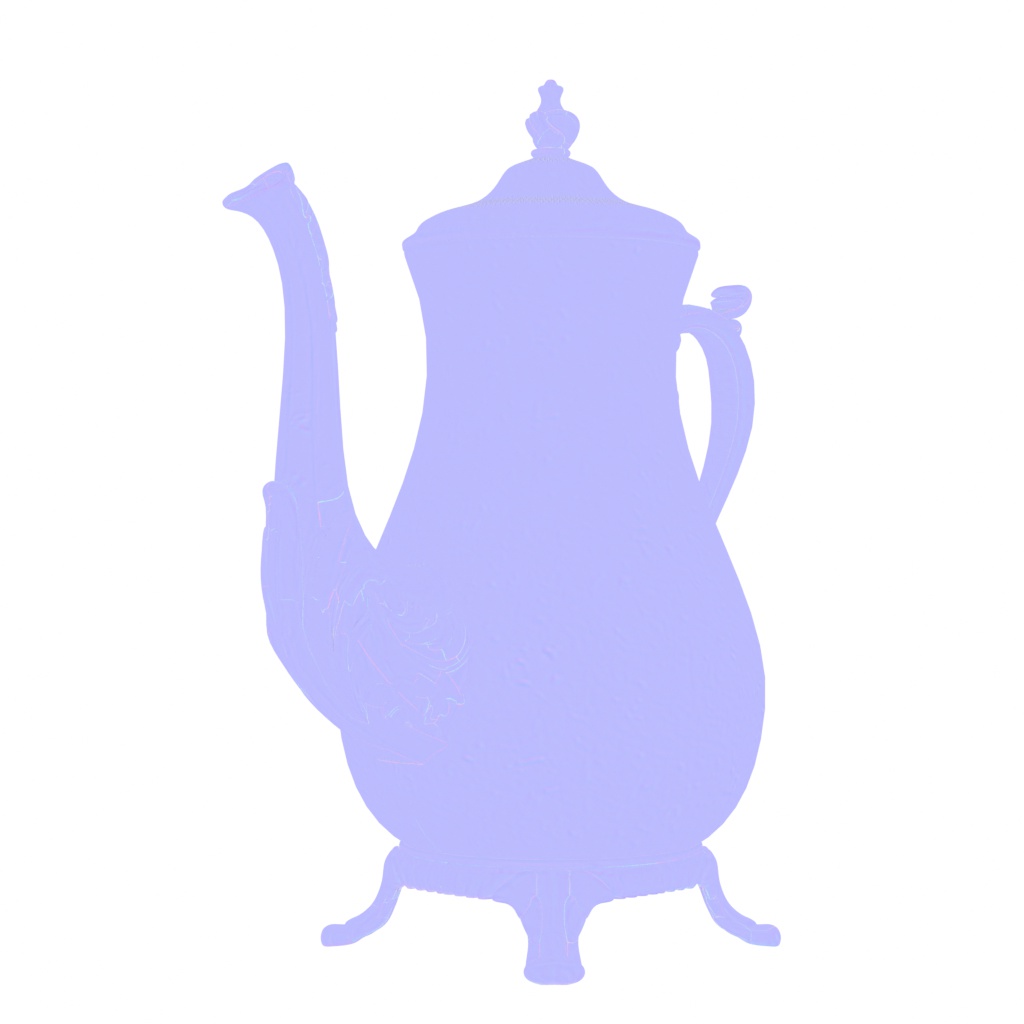}} \\
	& Relit & Base color & Roughness & Metallic & Normals
\end{tabular}
\vspace*{-2mm}
\caption{
	We compare two versions of joint prediction.  Our version uses the latent space of
    $\mathrm{VAE}_\mathrm{pbr}$ and predicts a video with 16 frames. Frame concatenation doubles the number of tokens by treating the material prediction as diffusing a video with twice the number of frames [16$\times$ base color, 16$\times$ HRM]. The resulting quality is similar, arguably with more coherent 
    material predictions in our approach. Please zoom to see the details.
}
\label{fig:frame_concat_sup}
\end{figure*}
}


\newcommand{\figImgCond}{
\begin{figure*}[t]
    \footnotesize
    \centering
    \setlength{\tabcolsep}{1pt}
    \begin{tabular}{ccccccc}
       \rotatebox[origin=c]{90}{Our (Text)} &
       \raisebox{-0.5\height}{\includegraphics[width=0.16\textwidth]{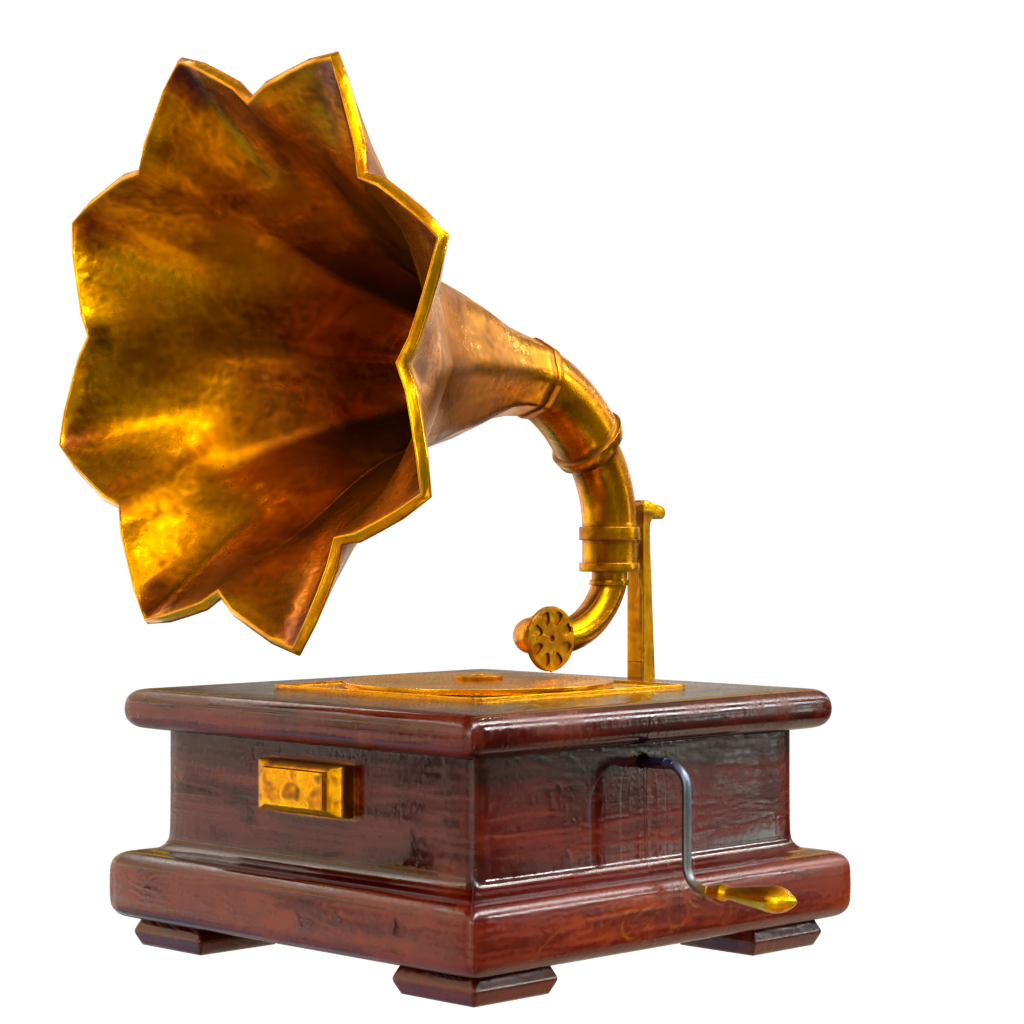}} &
       \raisebox{-0.5\height}{\includegraphics[width=0.16\textwidth]{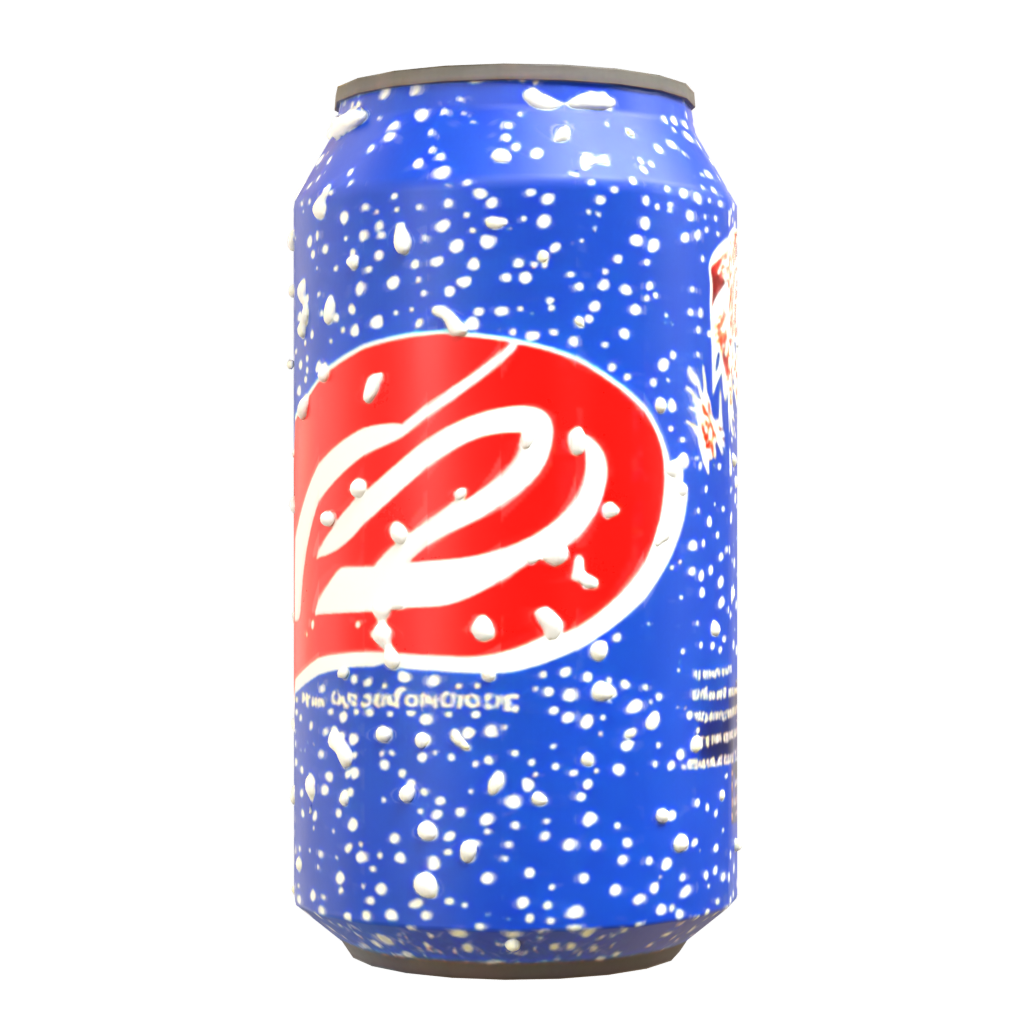}} &
       \raisebox{-0.5\height}{\includegraphics[width=0.16\textwidth]{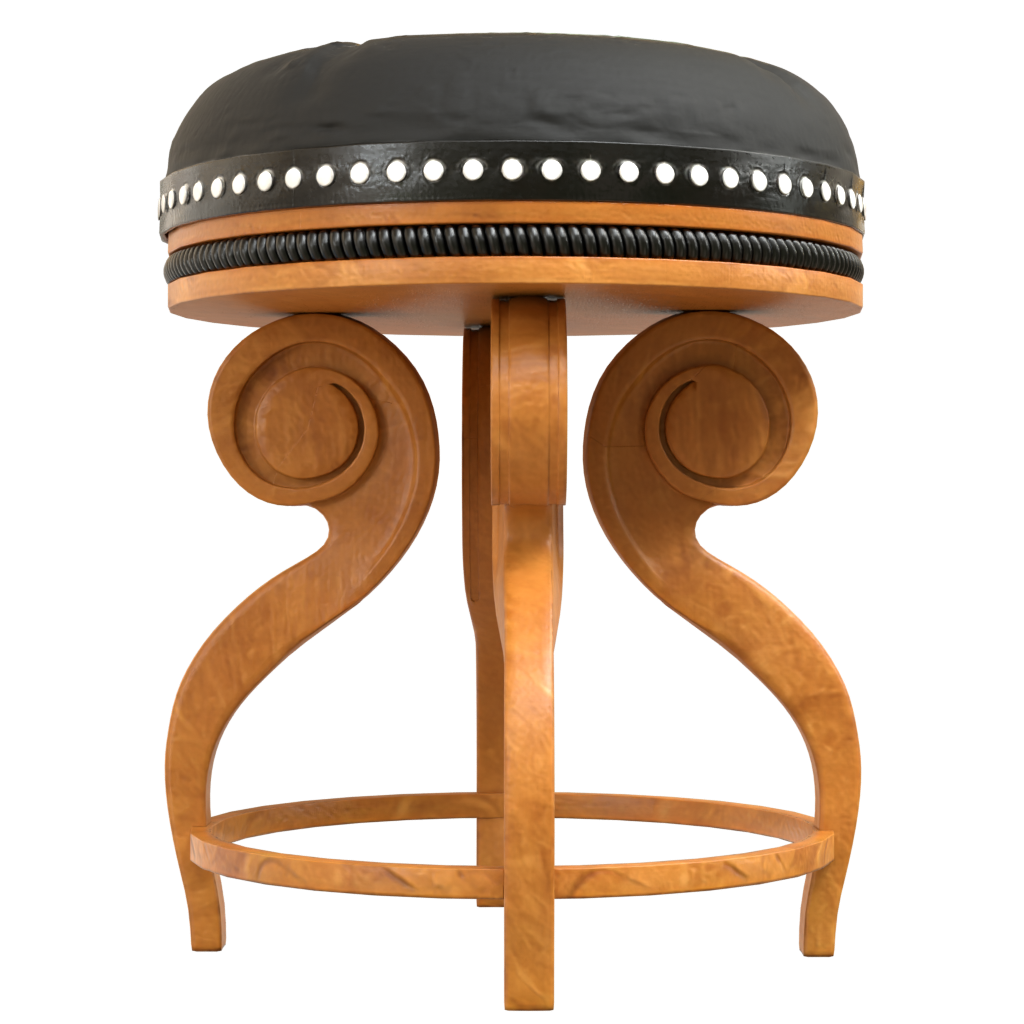}} &
       \raisebox{-0.5\height}{\includegraphics[width=0.16\textwidth]{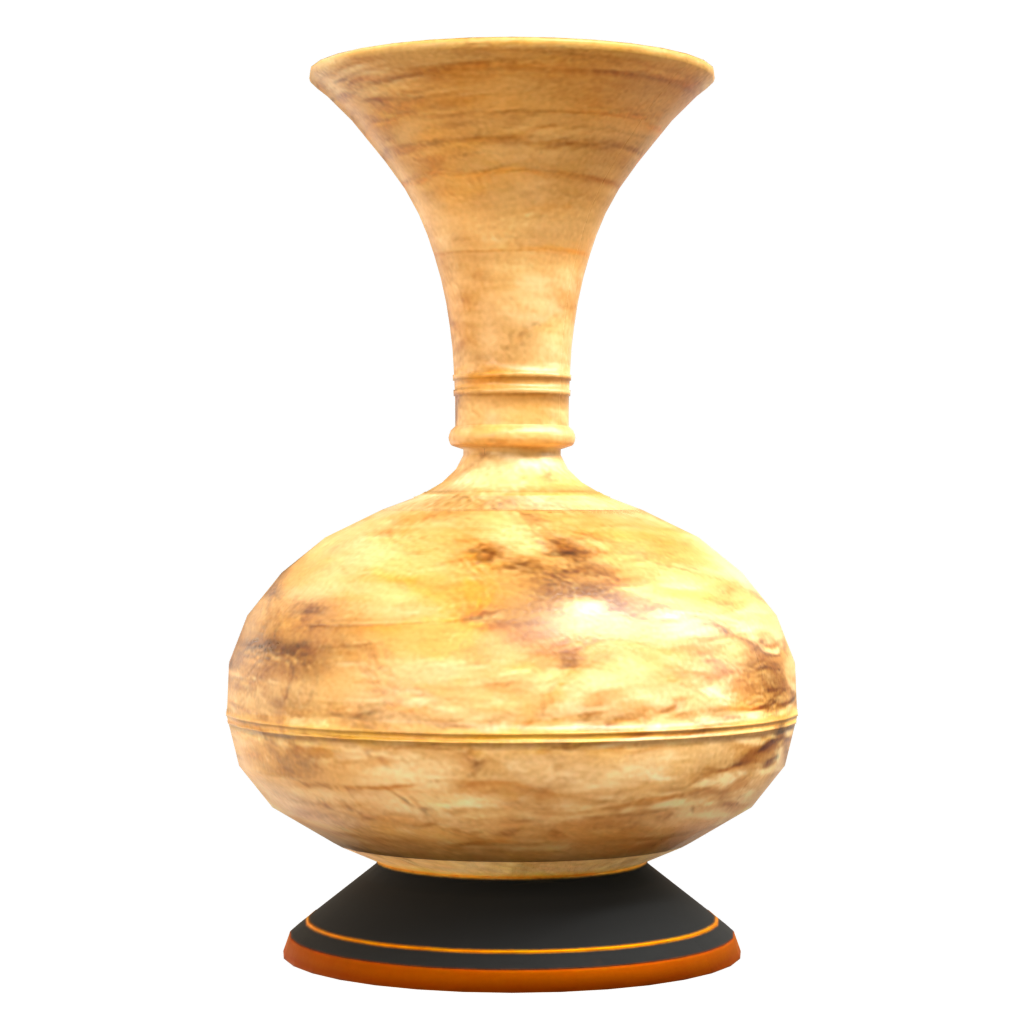}} &
       \raisebox{-0.5\height}{\includegraphics[width=0.16\textwidth]{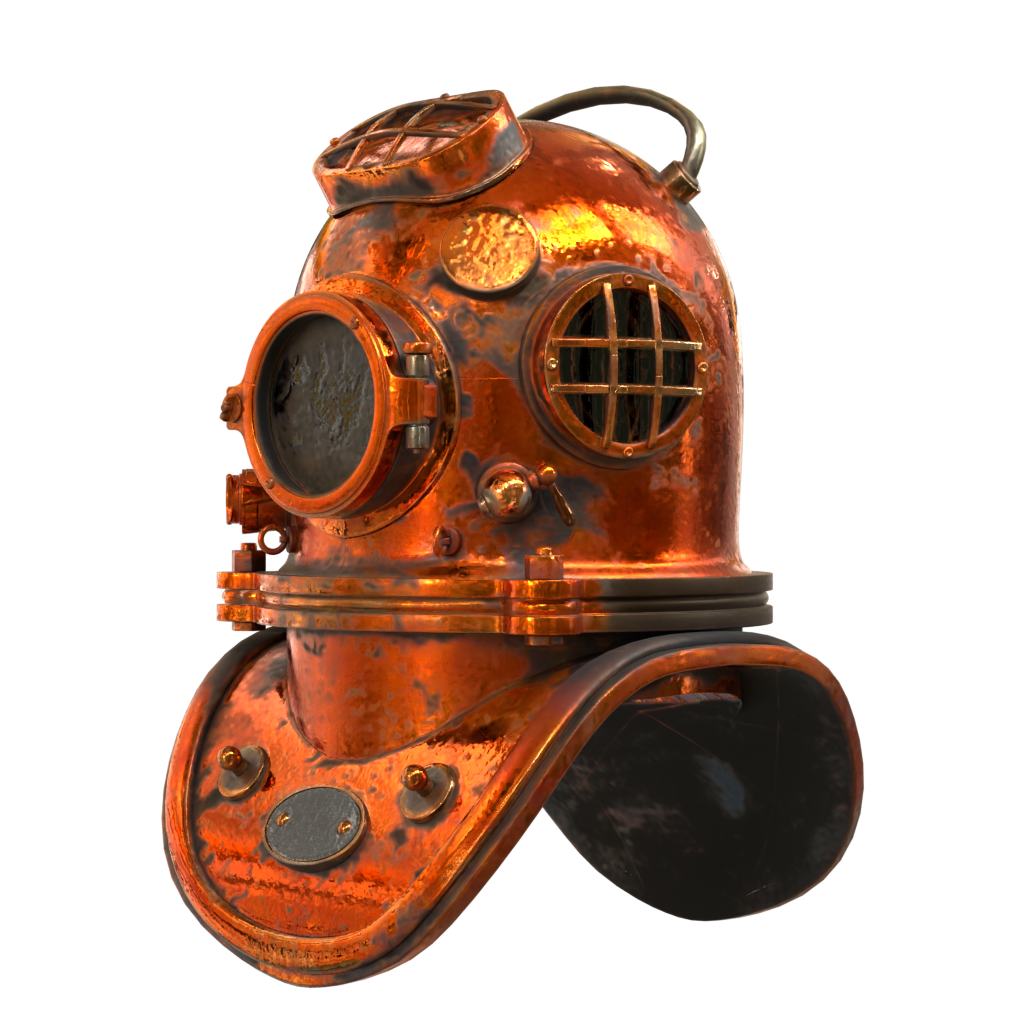}} &
       \raisebox{-0.5\height}{\includegraphics[width=0.16\textwidth]{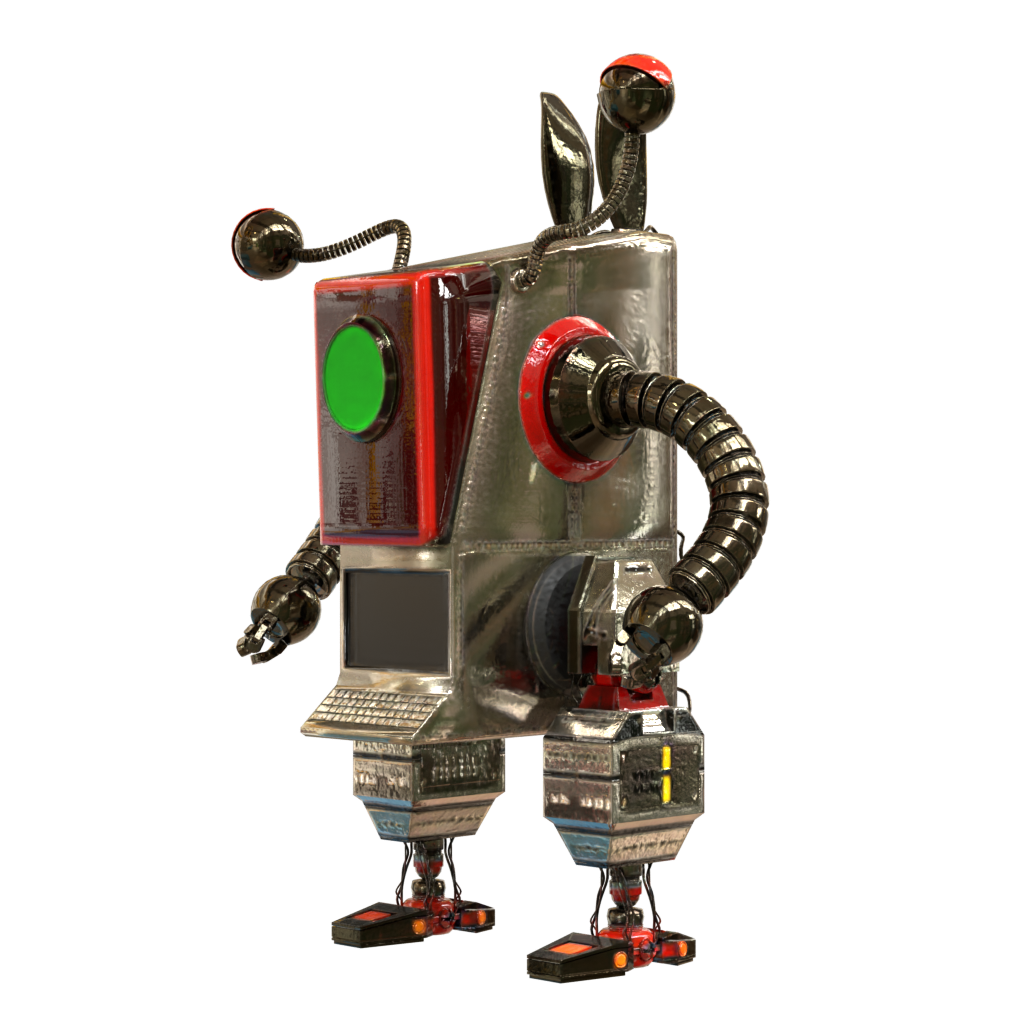}} \\
       \rotatebox[origin=c]{90}{Our (Image)} &
       \raisebox{-0.5\height}{\includegraphics[width=0.16\textwidth]{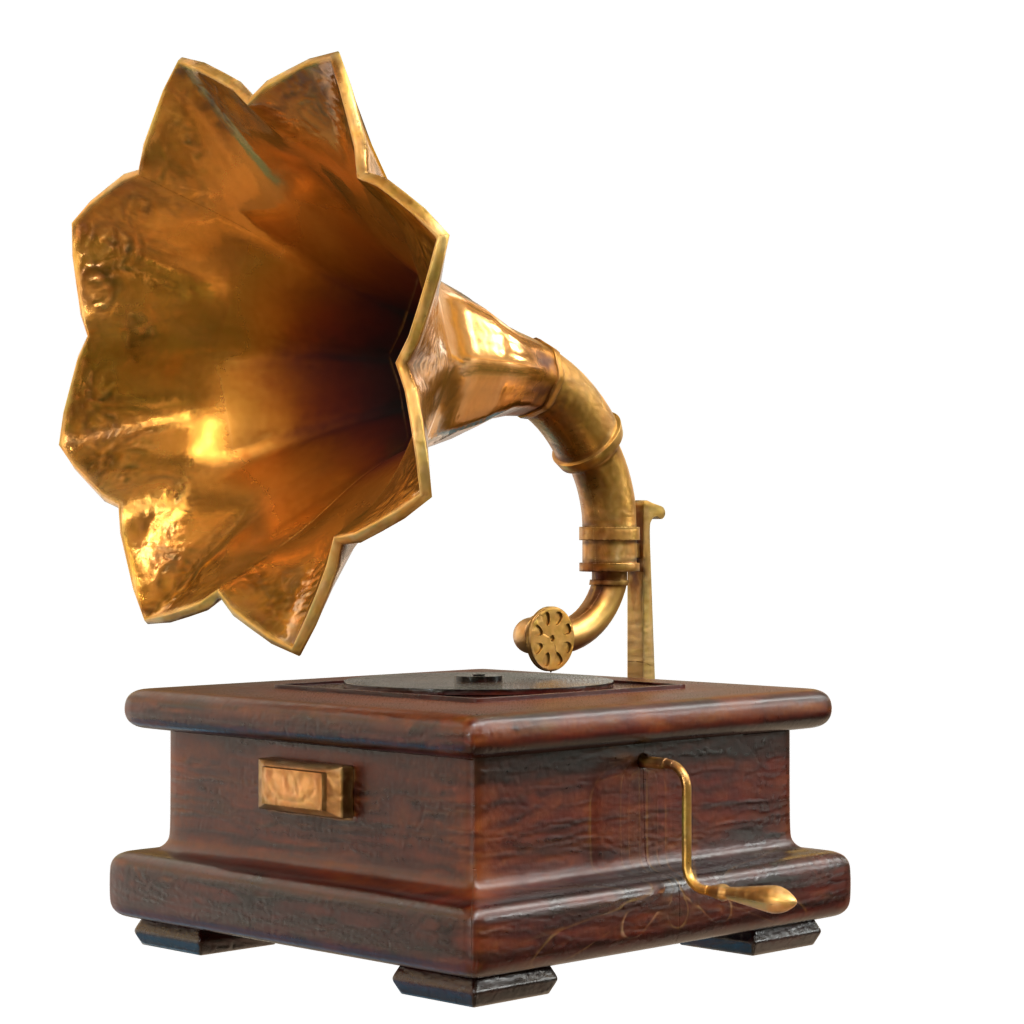}} &
       \raisebox{-0.5\height}{\includegraphics[width=0.16\textwidth]{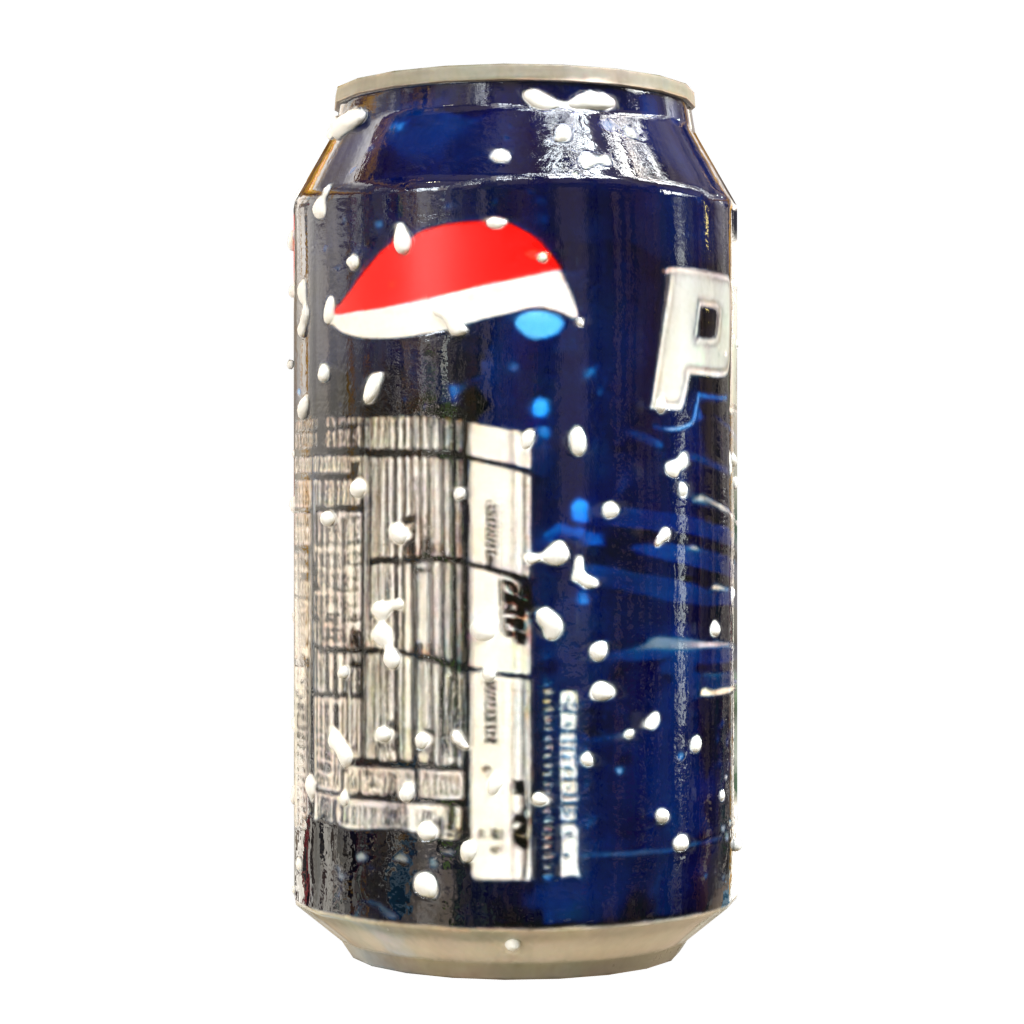}} &
       \raisebox{-0.5\height}{\includegraphics[width=0.16\textwidth]{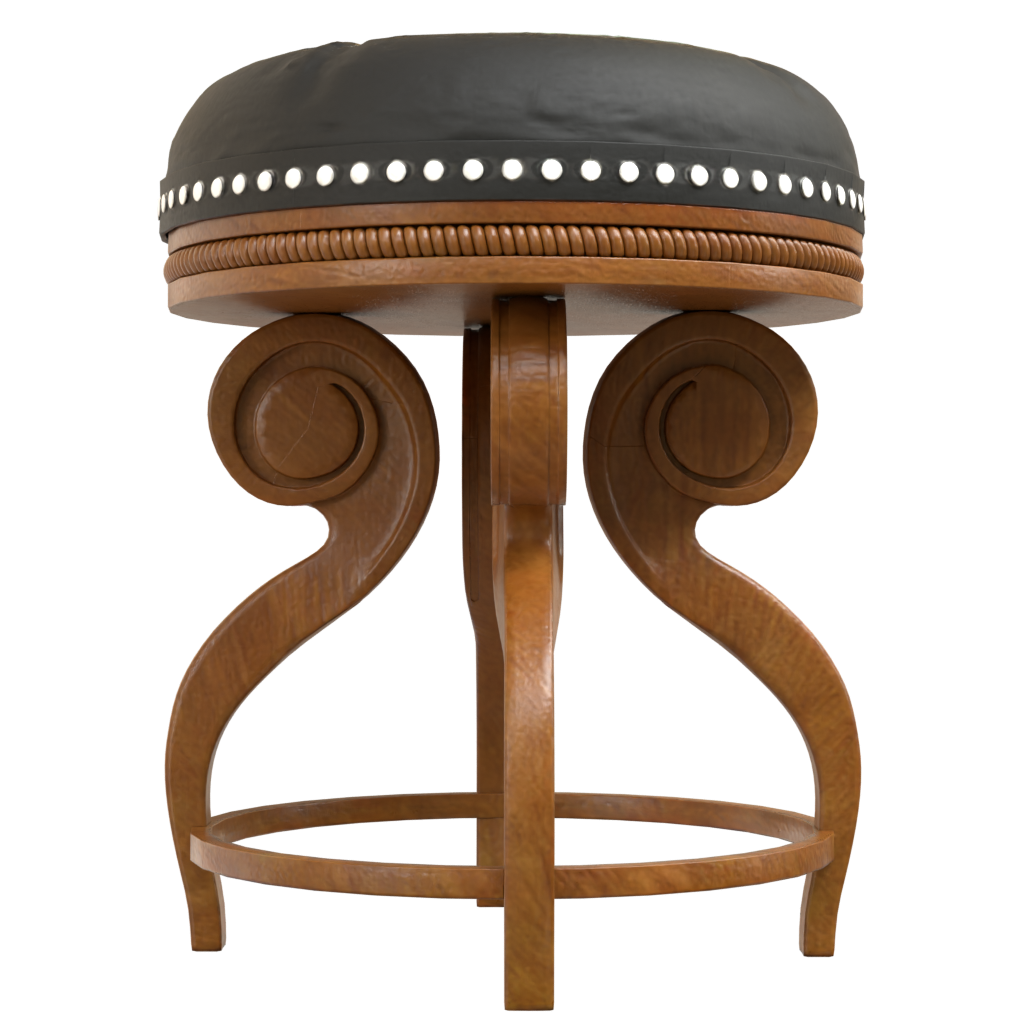}} &
       \raisebox{-0.5\height}{\includegraphics[width=0.16\textwidth]{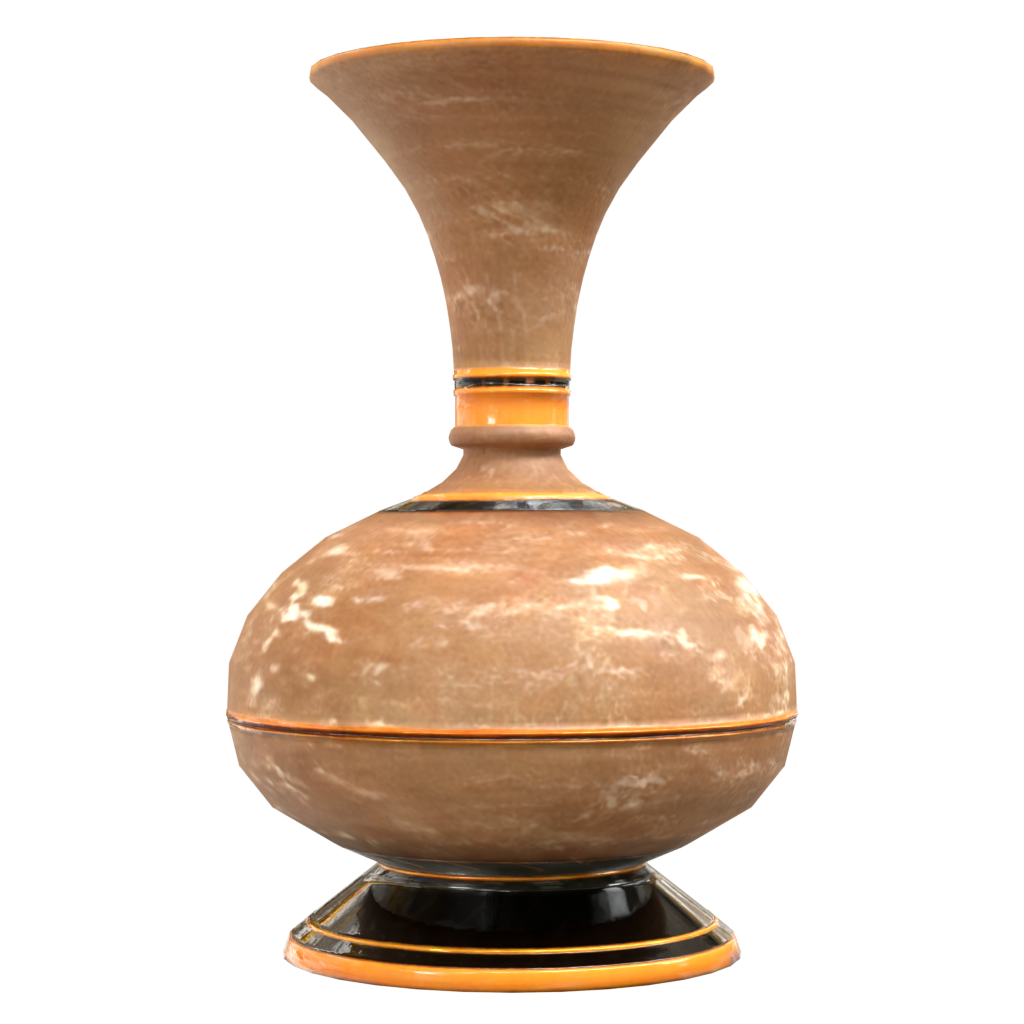}} &
       \raisebox{-0.5\height}{\includegraphics[width=0.16\textwidth]{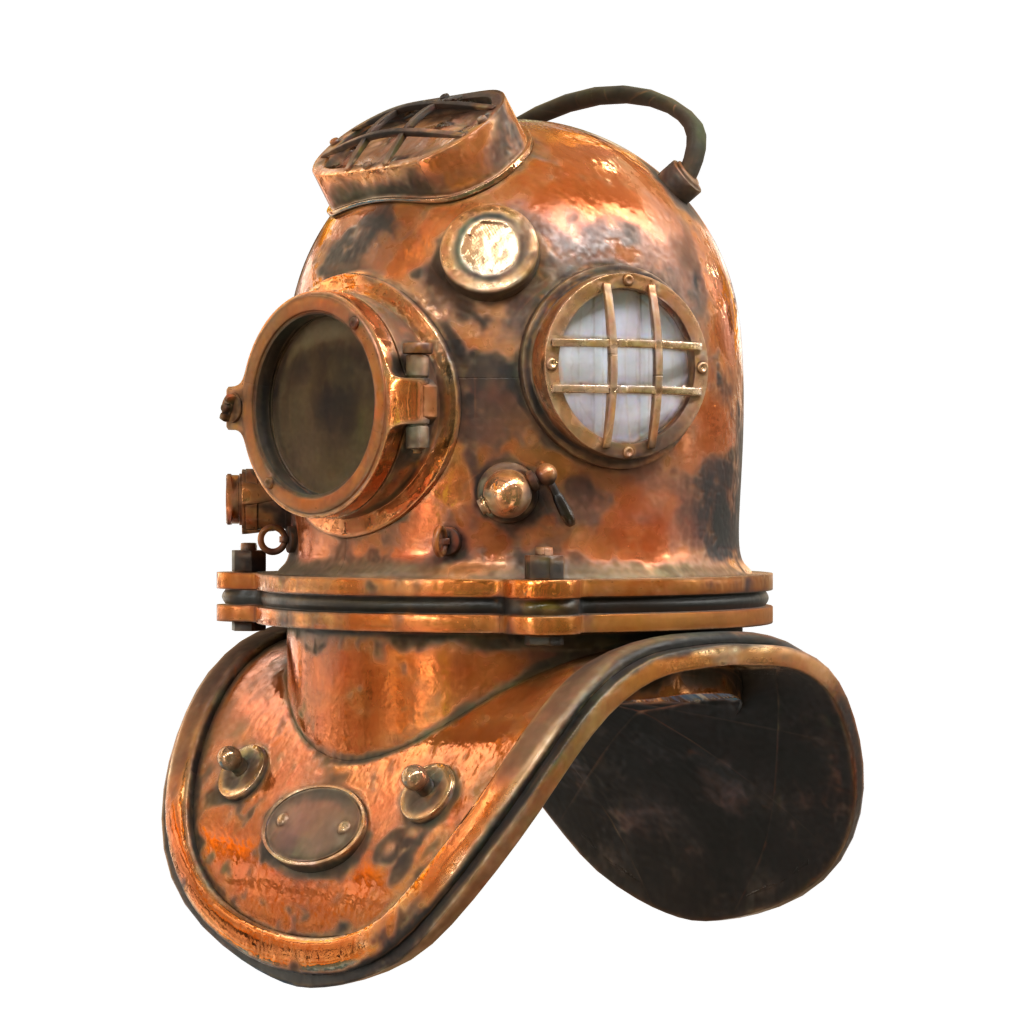}} &
       \raisebox{-0.5\height}{\includegraphics[width=0.16\textwidth]{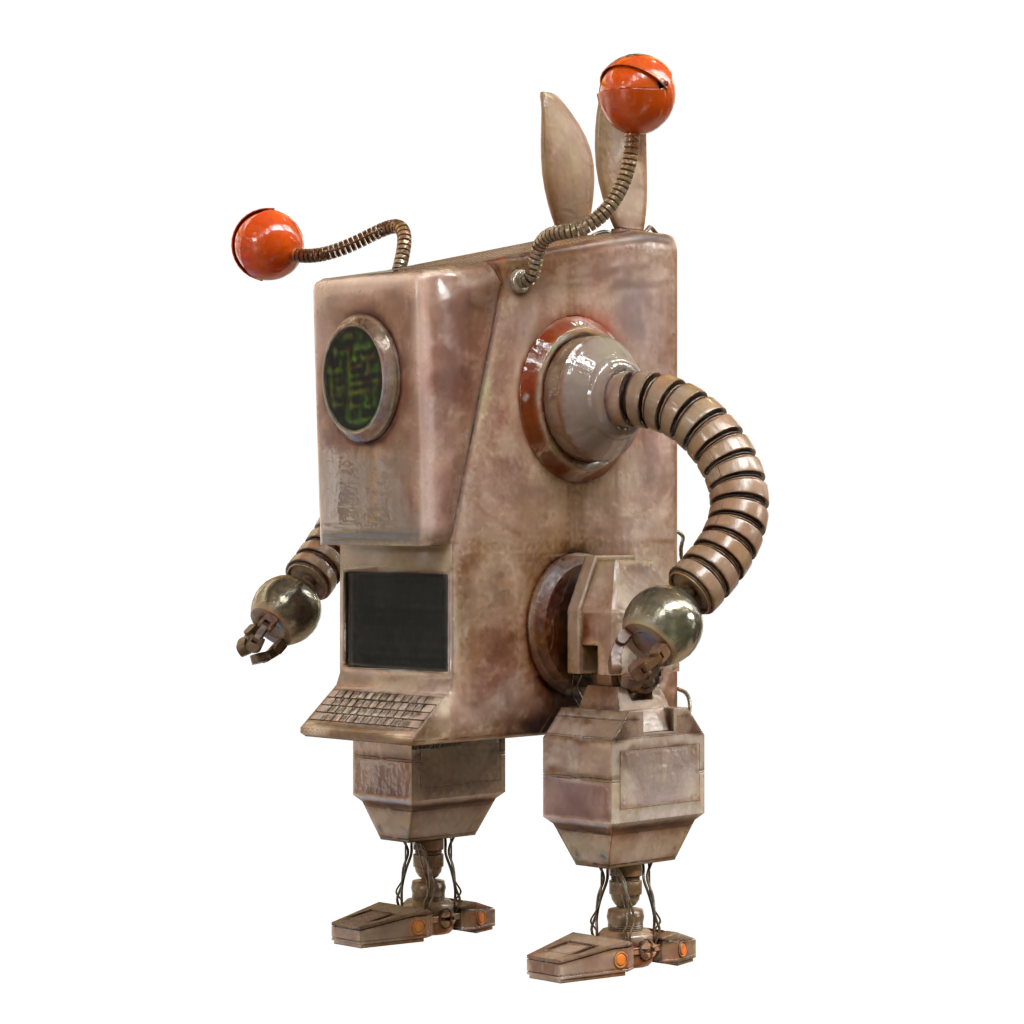}} \\
       \rotatebox[origin=c]{90}{Dataset entry} &
       \raisebox{-0.5\height}{\includegraphics[width=0.16\textwidth]{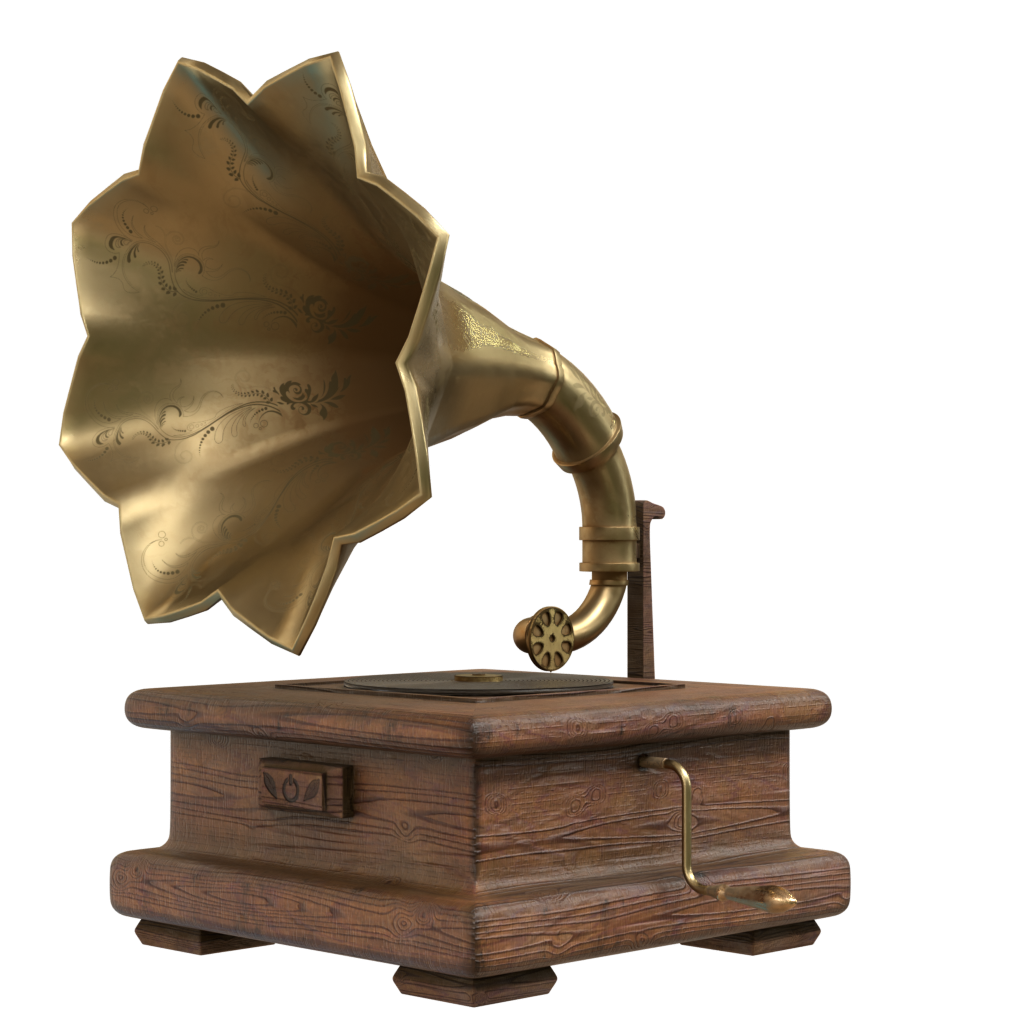}} &
       \raisebox{-0.5\height}{\includegraphics[width=0.16\textwidth]{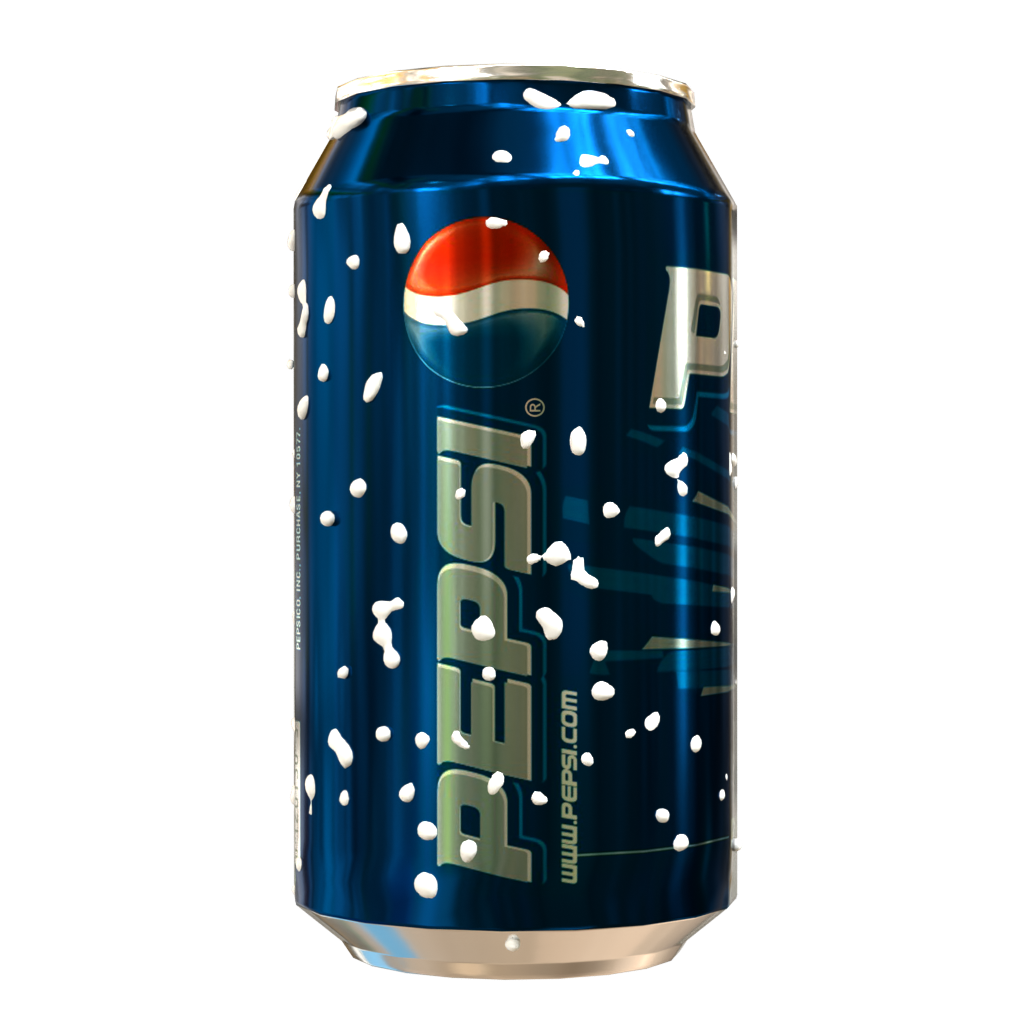}} &
       \raisebox{-0.5\height}{\includegraphics[width=0.16\textwidth]{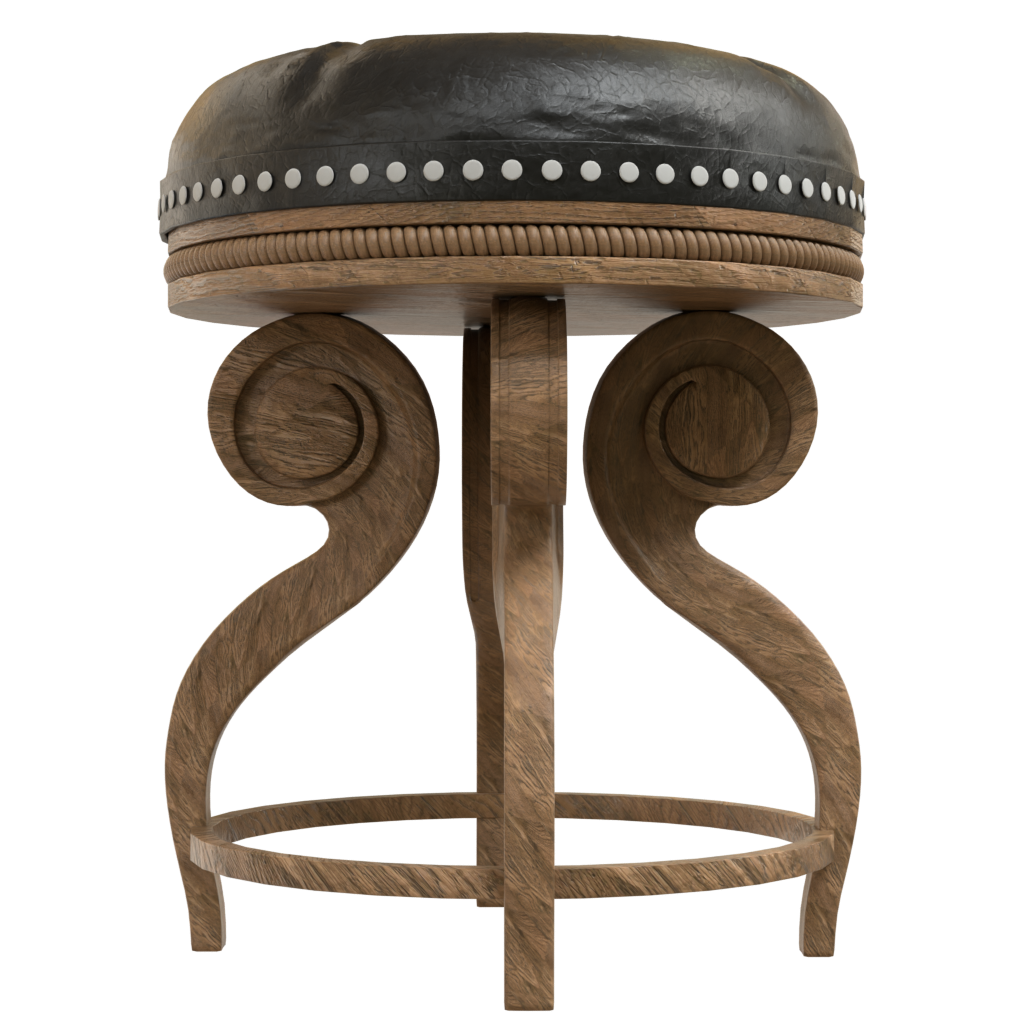}} &
       \raisebox{-0.5\height}{\includegraphics[width=0.16\textwidth]{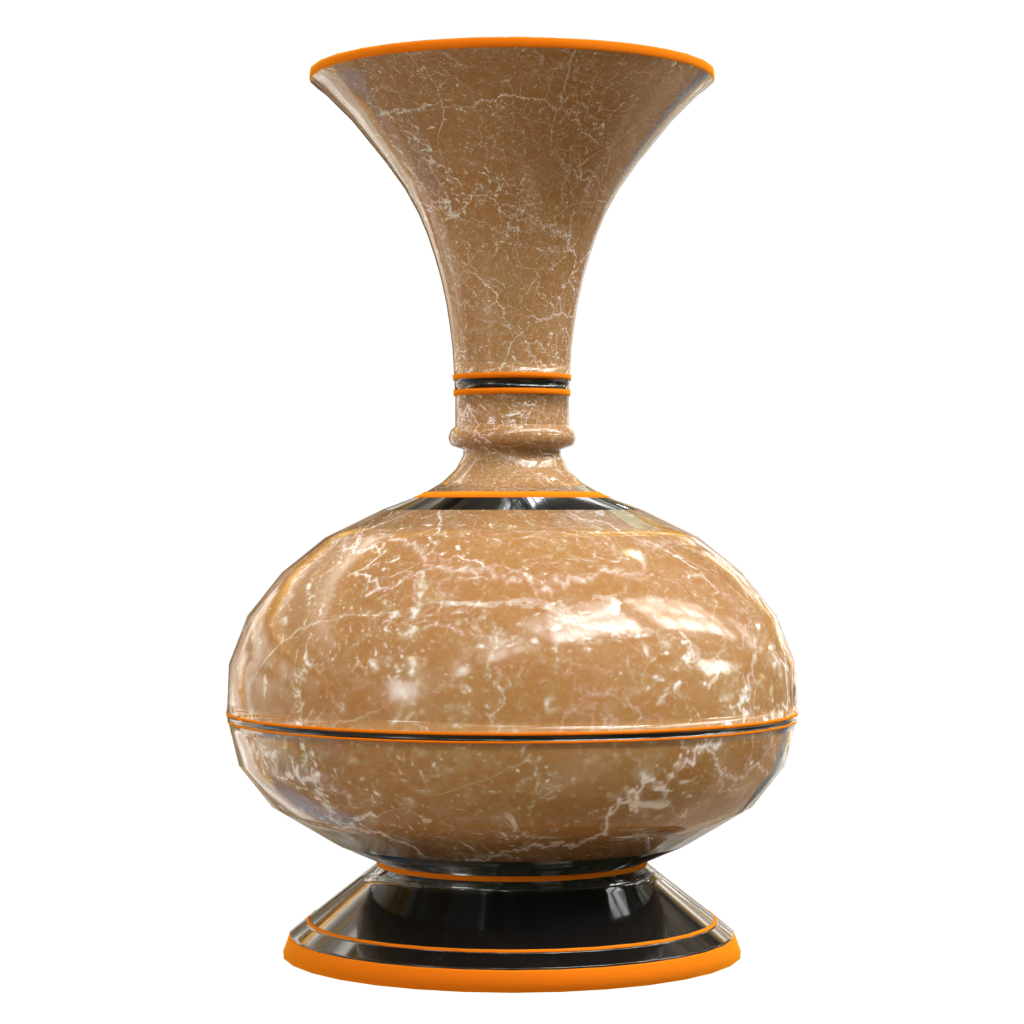}} &
       \raisebox{-0.5\height}{\includegraphics[width=0.16\textwidth]{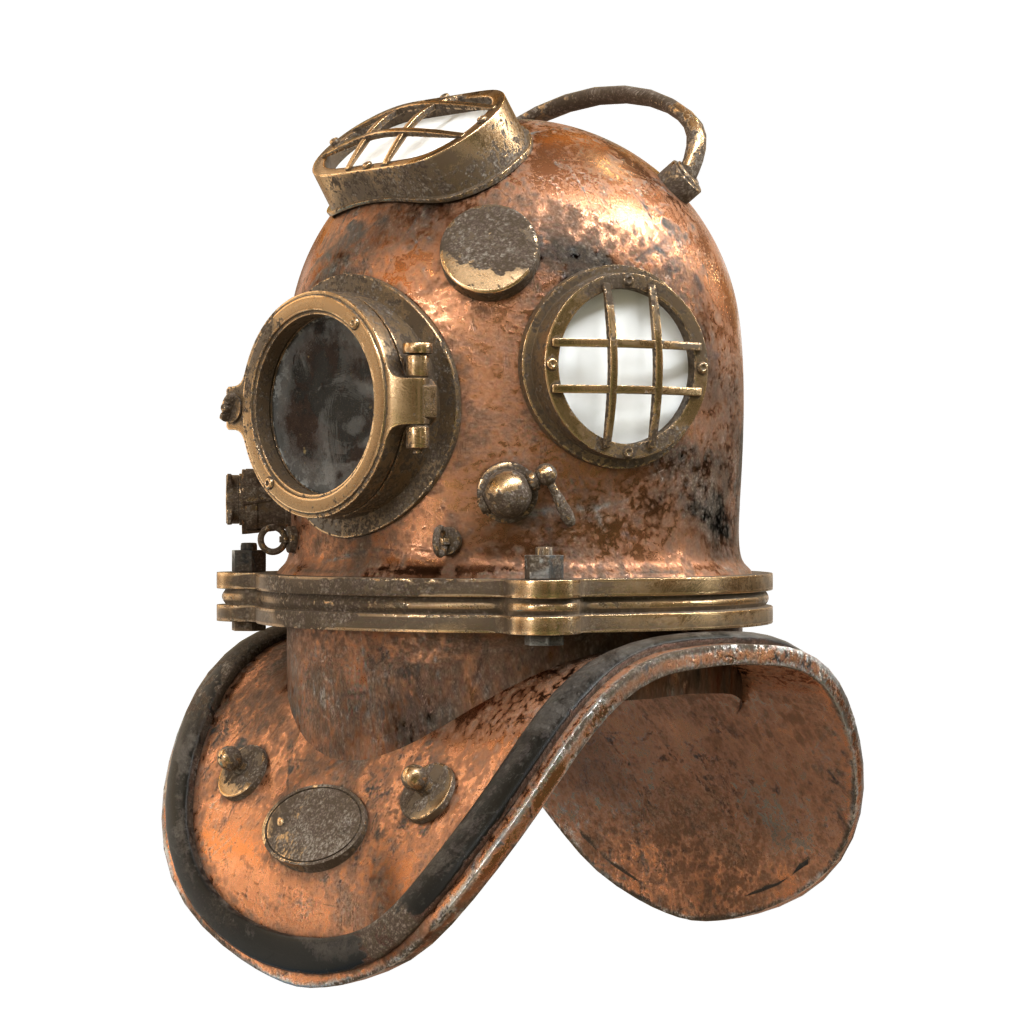}} &
       \raisebox{-0.5\height}{\includegraphics[width=0.16\textwidth]{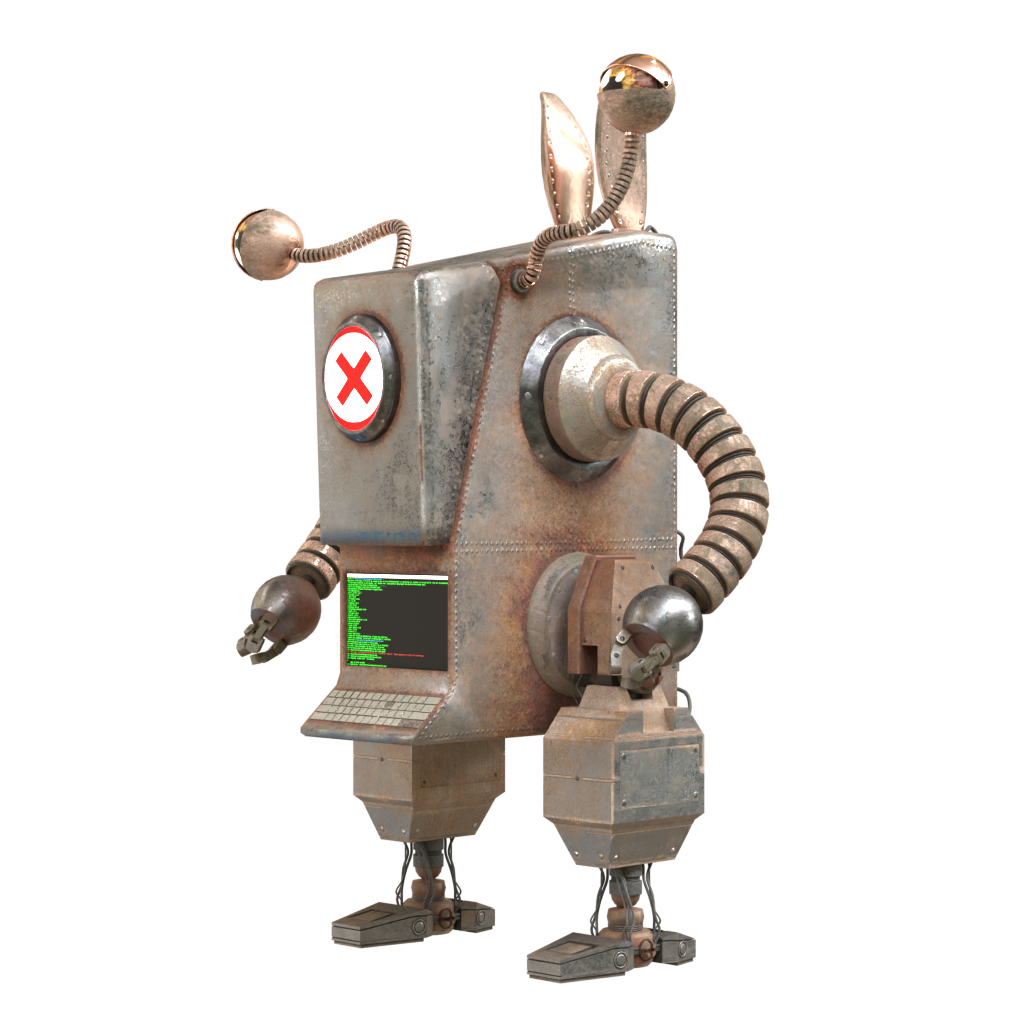}} \\
    \end{tabular}
    \caption{We compare our text- and image-guided models for six examples, and note that the image-guided version more closely resembles the materials of the dataset entry. We deliberately chose a view with $45^\circ$ rotation from the conditioning view.}
    \label{fig:imgcond}
 \end{figure*}
}


\newcommand{\figTrellis}{
\begin{figure}[t]
  \centering
  \small
  \setlength{\tabcolsep}{1pt}
  \newcommand{\trimbottom}{10pt}
  \newcommand{\trimtop}{30pt}
  \begin{tabular}{ccc}
        \includegraphics[trim=0 \trimbottom{} 0 \trimtop{}, clip, width=0.32\columnwidth]{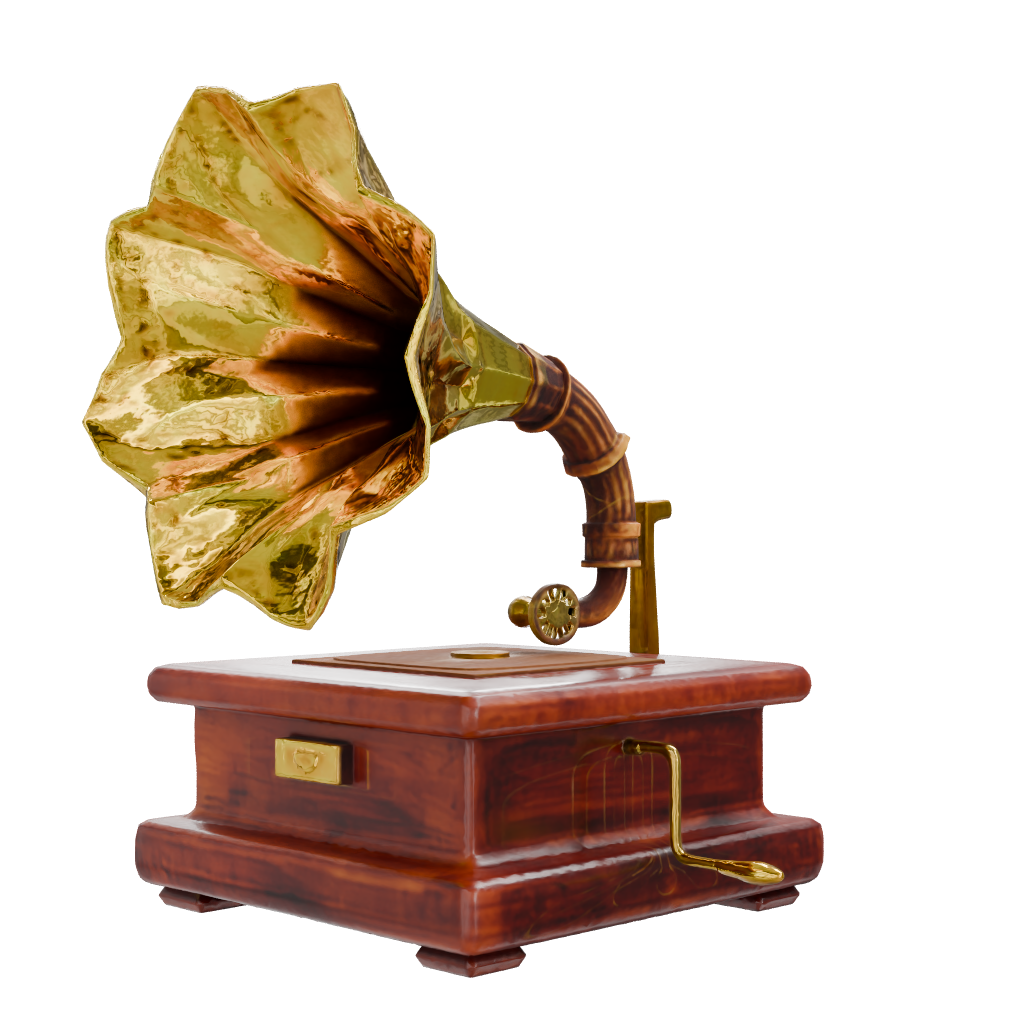} &
        \includegraphics[trim=0 \trimbottom{} 0 \trimtop{}, clip, width=0.32\columnwidth]{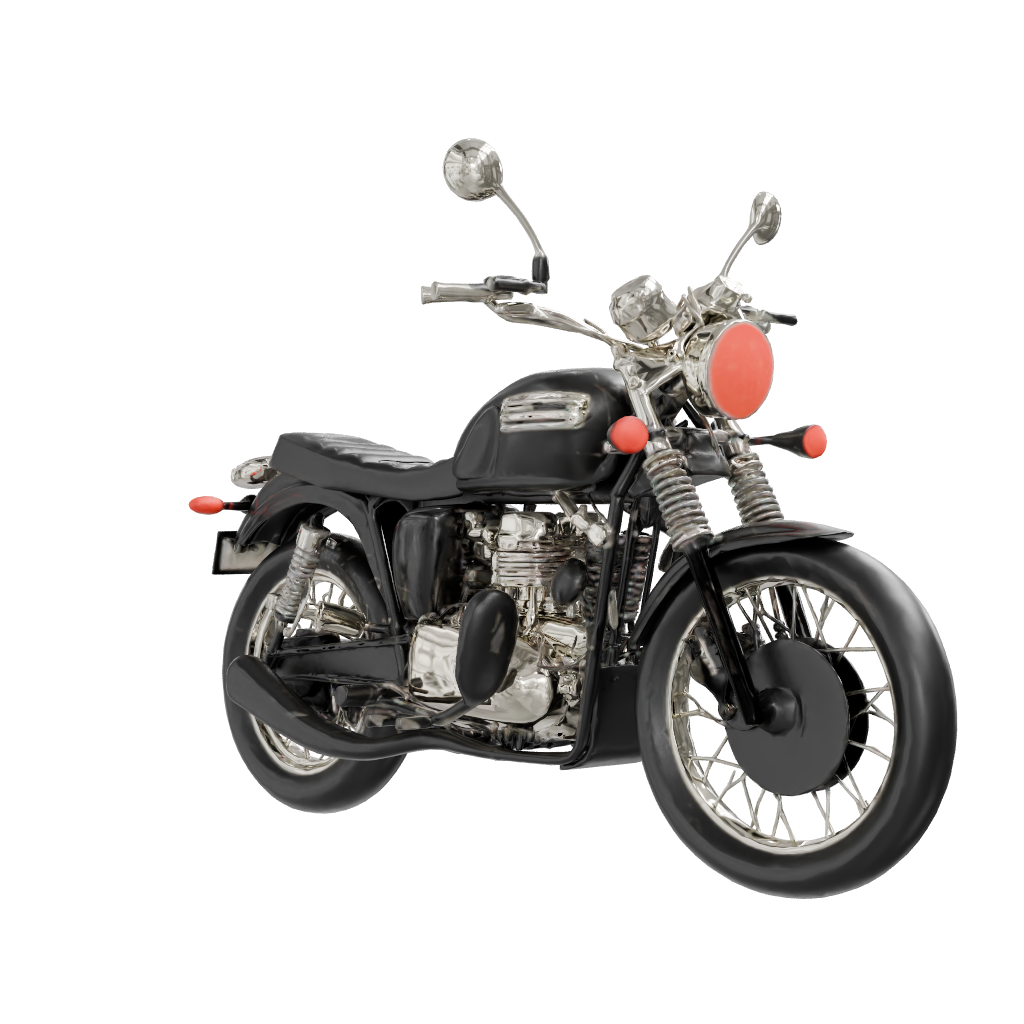} &
        \includegraphics[trim=0 \trimbottom{} 0 \trimtop{}, clip, width=0.32\columnwidth]{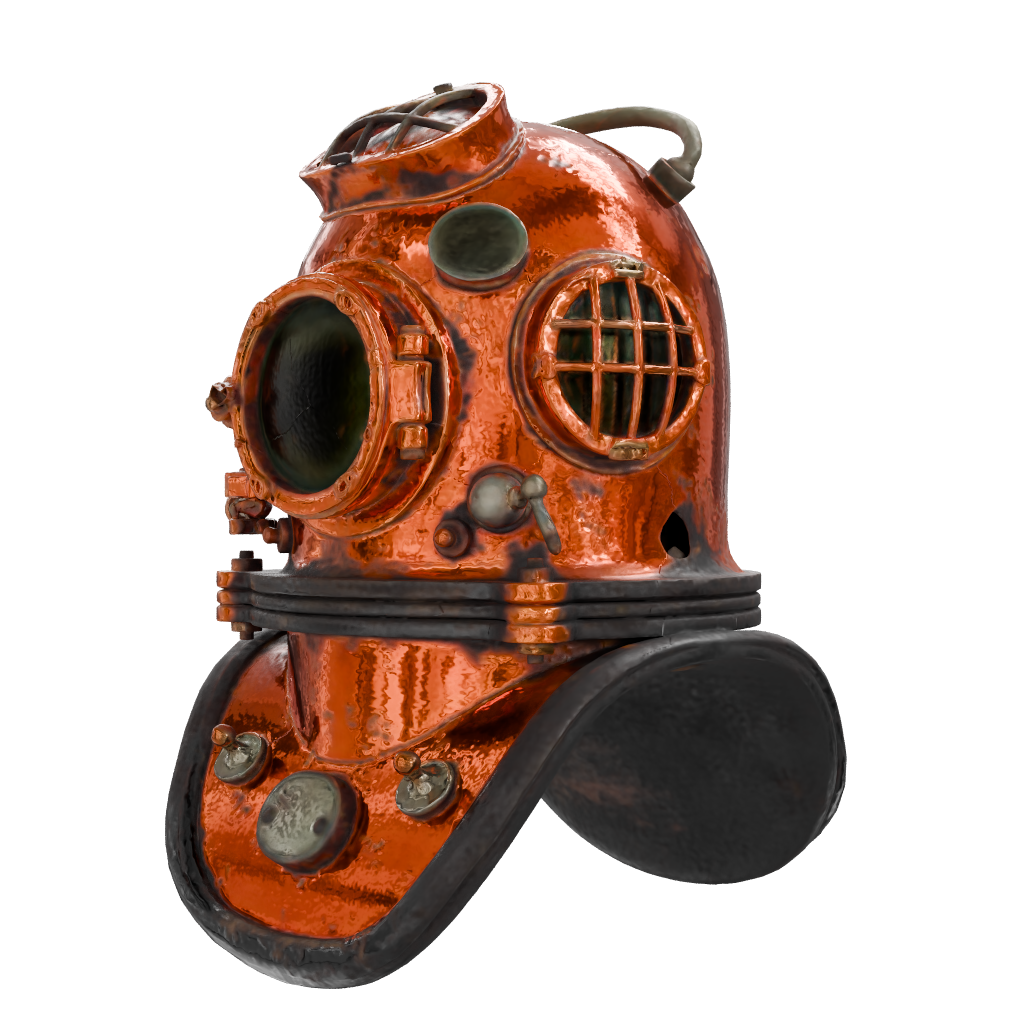} \\
        \includegraphics[trim=0 \trimbottom{} 0 \trimtop{}, clip, width=0.32\columnwidth]{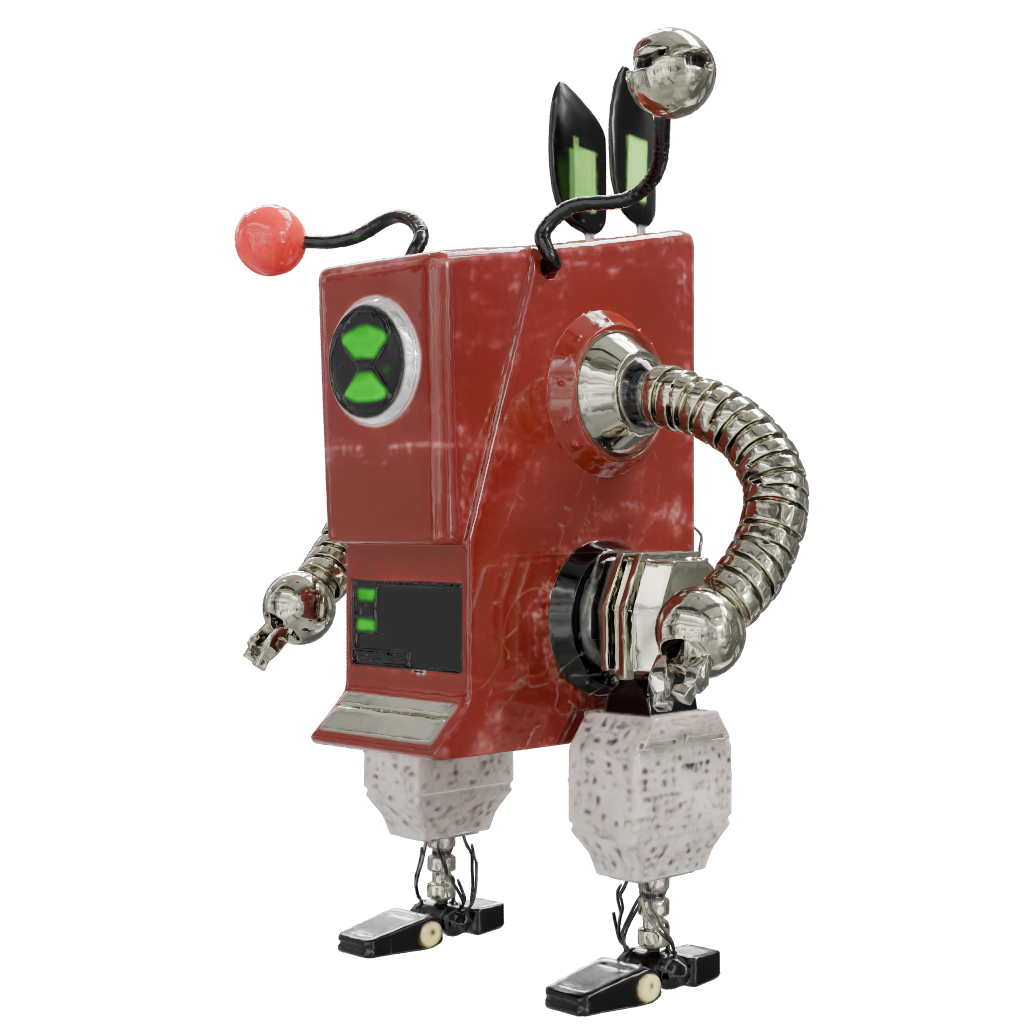} &
        \includegraphics[trim=0 \trimbottom{} 0 \trimtop{}, clip, width=0.32\columnwidth]{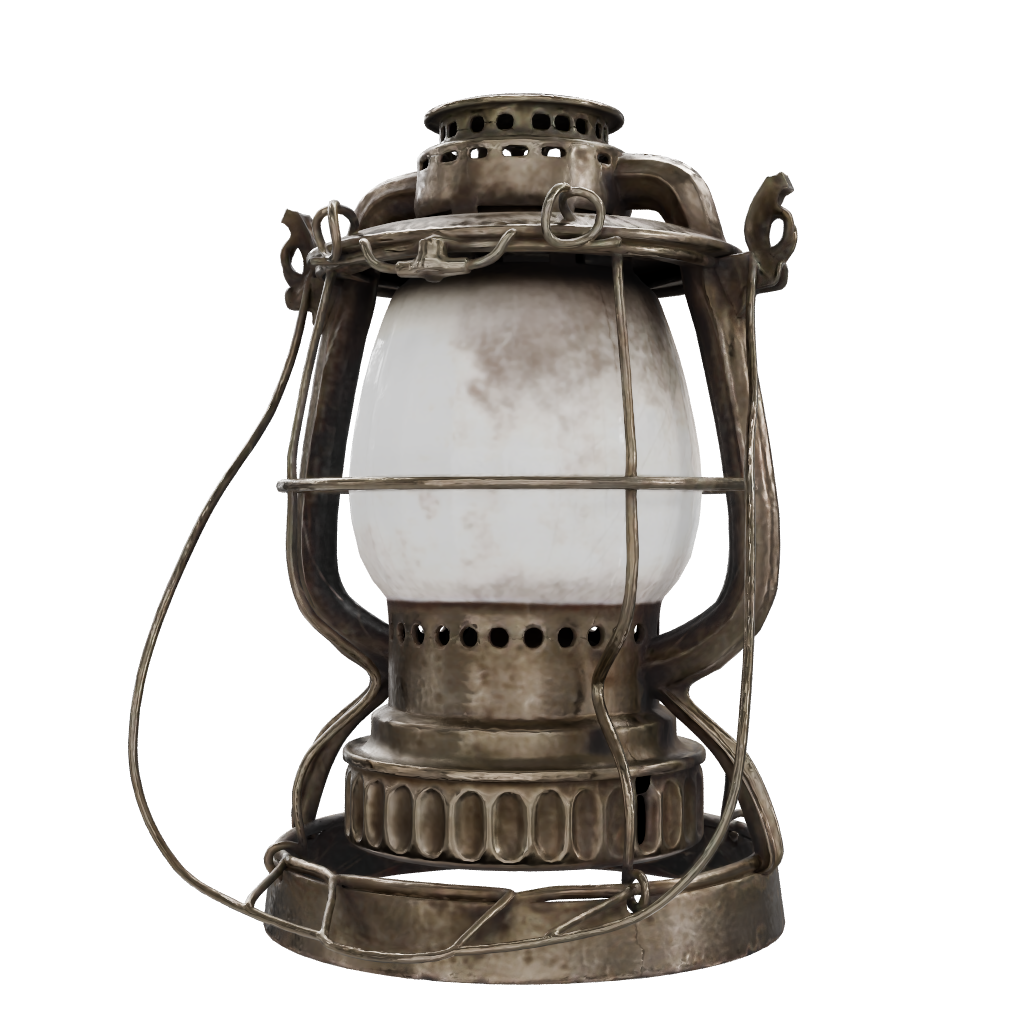} &
        \includegraphics[trim=0 \trimbottom{} 0 \trimtop{}, clip, width=0.32\columnwidth]{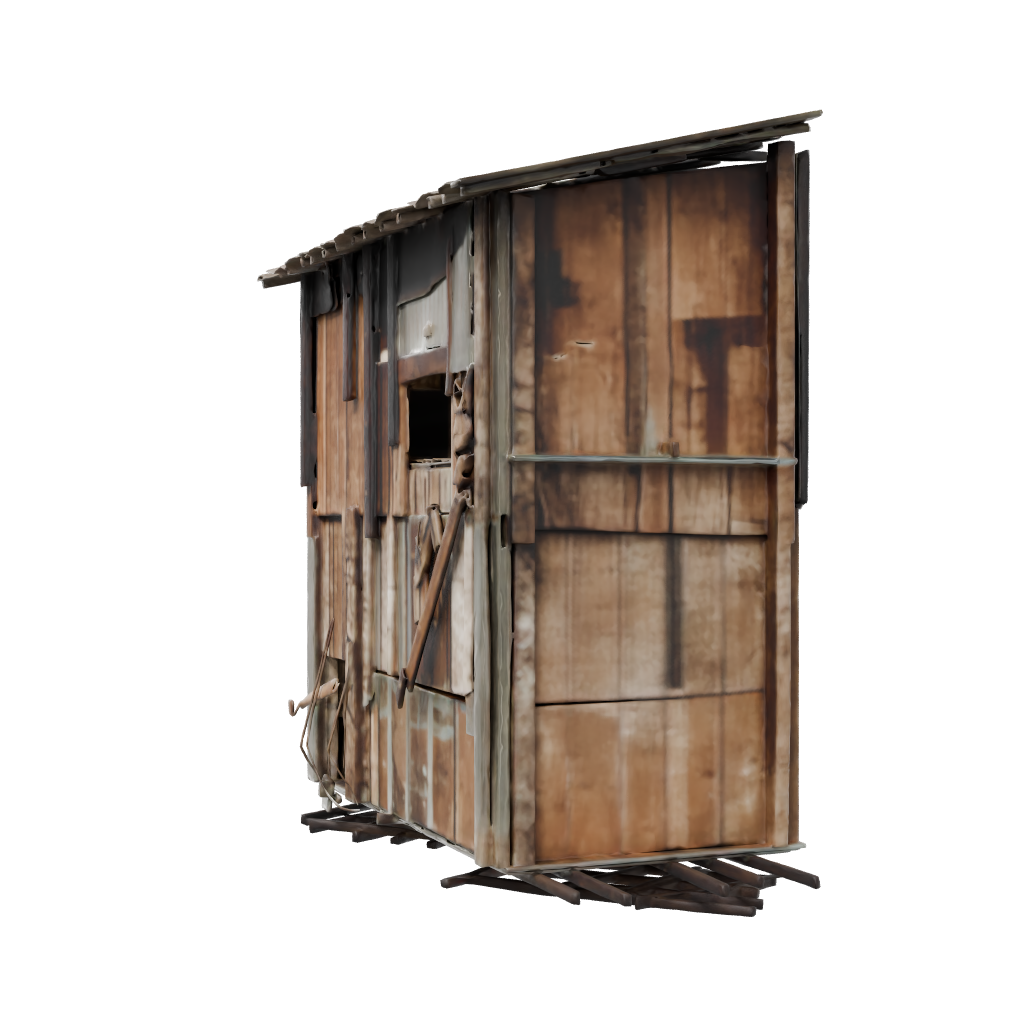}      
  \end{tabular}
  \vspace*{-2mm}
  \caption{Our text-to-PBR material generations (renderings in Blender) on a collection of TRELLIS.2-generated base geometry.}
  \label{fig:trellis}  
\end{figure}
}

\twocolumn[{%
\renewcommand\twocolumn[1][]{#1}%
\maketitle
   \includegraphics[width=\textwidth]{figures/teaser_small.jpg}
    \captionof{figure}{Given 3D models and text prompts, we generate unique high quality PBR materials for each 
    3D part using a finetuned video diffusion model. Our generated materials are directly applicable in 
    content creation applications. Here we show a Physical AI training application, applying the generated materials to a virtual factory setting. On the right, we show three variations of generated materials (from the same detailed text prompts and different random seeds) for an industrial robot asset with 19 parts.\vspace{1em}}
    \label{fig:teaser}
}]

\begin{abstract}
We present a method for generating physically-based materials for 3D shapes based 
on a video diffusion transformer architecture.
Our method is conditioned on input geometry and a text description, and jointly models multiple material properties (base color, roughness, metallicity, height map)
to form physically plausible materials. We further introduce a custom variational auto-encoder
which encodes multiple material modalities into a compact latent space, which 
enables joint generation of multiple modalities without increasing the number of tokens.  
Our pipeline generates high-quality materials for 3D shapes given a text prompt, compatible with common content creation tools.
\end{abstract}    
\section{Introduction}
\label{sec:intro}

Manually authoring 3D assets is time-consuming and requires expert skills; using generative models to produce 3D assets is a promising alternative. A new research field of leveraging 
diffusion models to generate 3D models from text prompts has recently emerged \cite{poole2022dreamfusion,zhang2024gslrm,gao2024cat3d,xiang2024trellis}. Another line of work assumes an input untextured 3D shape and generates texture through multi-view applications of image diffusion models~\cite{feng2025romantex,he2025materialmvp,shao2025mvpainter,yang2025pandora3d,engelhardt2025svim3d}. While the results look impressive for novel view synthesis, the methods bake final RGB colors (under some lighting) into the asset and cannot extract materials for physically-based rendering (PBR)~\cite{Burley12,Walter2007}, critical in more advanced content creation workflows.
Fitting these properties through differentiable rendering is possible, but in addition to unknown lighting, image diffusion models typically lack perfect view-consistency, introducing blur and 
washing out material details. This is particularly obvious when optimizing parameters for PBR material models, 
which rely on consistency of specular reflections.

Video diffusion models provide improved view consistency and exceed image models in handling specular highlights. 
This greatly helps when estimating per-pixel material parameters and for intrinsic decomposition~\cite{DiffusionRenderer}. This decomposition approach is utilized in recent work, VideoMat~\cite{munkberg2025videomat}, to generate a video orbit around a given 3D shape with synthesized final RGB appearance, and finally extract material parameters from this video using intrinsic decomposition. While the results are promising, their quality is limited by unnecessarily solving two hard problems that cancel each other out: synthesizing final appearance under natural lighting, and then removing the lighting to estimate clean material parameters. 

We present VideoMatGen, a video diffusion method for direct text-to-material generation.
Our work extends VideoMat~\cite{munkberg2025videomat} to generate higher quality materials
using a more efficient fused architecture based on joint generative modeling, without relying on an intermediate RGB appearance.
We start from a known untextured 3D geometry and a text prompt describing the desired material.
We condition a video diffusion model on multiple views of geometry guides (G-buffers): surface normals and world space positions.
By fine-tuning a recent video model, Cosmos Predict~1-7B~\cite{cosmos_short}, with  a custom dataset mapping these conditions and text prompts to material parameters,
we generate video sequences of synthesized intrinsic material channels (G-buffers): \emph{base color}, \emph{roughness}, 
\emph{metallicity}, and \emph{height}. 
Finally, the resulting views are projected into traditional texture maps, optionally turning height into normal variation (though the height could also be used as displacement).
As shown in \cref{fig:teaser} we produce spatially varying, detailed materials
that adapt to the underlying geometry. In comparison to related 
work, we show higher quality results and improved separation of lighting and materials.
Our main contributions are:
\begin{itemize}
    \item A video diffusion method for generating physically-based materials for 3D shapes based on text prompts, jointly predicting base color, roughness, metallicity, and height.
    \item A unified variational auto-encoder and latent space, jointly encoding base color, roughness, metallicity and height. This enables improved joint prediction without increasing the number of tokens.
\end{itemize}

\section{Related Work}

\paragraph{Diffusion Models.}
Image diffusion models add random noise to an image through a sequence of diffusion steps. They are trained to reverse this process, enabling sample generation by iterative denoising starting from Gaussian noise. 
Many generative models have been developed based on similar principles~\cite{sohl2015deep, ho2020denoising, dhariwal2021diffusion}.
Recently, video diffusion models~\cite{blattmann2023videoldm, blattmann2023svd, hong2023cogvideo, yang2024cogvideox, cosmos_short} extend image-based diffusion approaches to the temporal domain, enabling video generation from inputs such as text or an initial frame. Diffusion transformer (DiT) models have become the standard architecture of choice for both image and video diffusion \cite{peebles2022dit} due to their performance and flexible finetuning opportunities.
In this work, we build upon the Cosmos~\cite{cosmos_short} DiT-based video diffusion model.

\paragraph{Differentiable Rendering.}
In this paper, we focus on mesh-based surface geometry with PBR materials~\cite{Burley12}. 
Previous work includes differentiable rasterization~\cite{Laine2020diffrast}, 
which has low run-time cost and has been successfully applied to photogrammetry~\cite{Munkberg_2022_CVPR}. 
Differentiable path tracing~\cite{Zhang:2020:PSDR,Mitsuba3} approaches are considerably more costly, and introduce Monte-Carlo noise 
in the training process, which can make gradient-based optimization more challenging. However, path tracing accurately simulates global 
illumination effects, and has higher potential reconstruction quality. Fuzzy scene representations such as NeRFs~\cite{Mildenhall2020} and 
Gaussian splatting~\cite{kerbl3Dgaussians} are commonly used in optimization setups, 
and generate impressive novel-view synthesis results. However, disentangling materials and lighting remains non-trivial. We assume known mesh geometry, but our approach can be extended to generate materials on other geometry representations (e.g. Gaussians, SDFs, etc.).

\paragraph{Texture and material extraction using diffusion.} 
Various hybrid approaches combine image diffusion models with inpainting, or coarse-to-fine texture 
refinement, such as TEXTure~\cite{richardson2023texture}, Text2tex~\cite{chen2023text2tex}, and Paint3D \cite{zeng2024paint3d}. 
Paint-it~\cite{youwang2024paintit} proposes representing material texture maps with randomly initialized convolution-based neural kernels. This regularizes the optimization landscape, improving material quality.
TextureDreamer~\cite{yeh2024texturedreamer} finetunes the diffusion model using Dreambooth~\cite{ruiz2022dreambooth} with a few 
images of a 3D object, and uses variational score distillation \cite{wang2023prolificdreamer} to optimize the material maps.
DreamMat~\cite{Zhang2024dreammat} and FlashTex~\cite{deng2024flashtex} improve on light and material disentanglement by finetuning 
image diffusion models to condition on geometry and lighting, allowing for optimization over many known lighting conditions. 

\figSystem

MaPa~\cite{zhang2024mapa}, MatAtlas~\cite{ceylan2024matatlas}, and Make-it-Real~\cite{fang2024makeitreal} start from a database of known 
high-quality materials, and learn to project the input (image or text) onto the known representation. MaPa relies on \
material graphs and optimize parameters of known graphs, while Make-it-Real uses a database of PBR-textures, and 
MatAtlas a database of procedural materials. These methods are limited by the expressiveness of their material databases, but benefit from much improved regularization.

\paragraph{Diffusion-based 3D asset generation.} Many methods build on image diffusion models to produce full 3D assets, with either RGB colors or PBR material maps. DreamFusion~\cite{poole2022dreamfusion} introduces a \emph{score distillation sampling} (SDS) loss, and generates 3D assets from pre-trained text-to-image diffusion models. This approach has since been refined~\cite{zhu2023hifa,wang2023prolificdreamer,zhu2023hifa}. SDS-based methods require slow optimization, prompting the development of methods like Instant3D \cite{li2023instant3d} and GS-LRM \cite{zhang2024gslrm} that instead reconstruct in a forward pass using a single pretrained transformer model.

A common limitation for most image models is lack of view consistency, which may show up as 
blur in the extracted textures. SV3D~\cite{voleti2024sv3d} and Hi3D~\cite{yang2024hi3d} improve on this aspect by 
finetuning video models for object rotations, and extract 3D models from the generated views. However, these approaches 
have limited resolution and do not provide PBR materials. Trellis~\cite{xiang2024structured} and TEXGen~\cite{yu2024texgen} 
avoid the view consistency problem altogether by having the diffusion model operate directly in 3D space
and texture space respectively.
These methods show great promise, but they do not focus on material parameter generation.
CLAY~\cite{zhang2024clay} and SF3D~\cite{boss2025sf3d} also generate 3D geometry and materials from text or image inputs. CLAY's material generation models uses a finetuned multi-view image diffusion model~\cite{shi2023MVDream} conditioned on normal maps. 
The material model generates four canonical views of the PBR texture maps (base color, roughness, metallicity), which are then projected 
into texture space. Several recent methods~\cite{feng2025romantex,he2025materialmvp,shao2025mvpainter,yang2025pandora3d,engelhardt2025svim3d,seed3d} 
extends this approach with additional input conditioning (normal, depth and/or world space positions).
3DTopia-XL~\cite{chen2024primx} proposes a novel
3D representation, which encodes the 3D shape, textures, and materials in volumetric primitives anchored to the surface of the object. 
Their denoising process jointly generates shape and PBR materials.
 
\paragraph{Intrinsic decomposition of images/videos.} Another related line of research is intrinsic decomposition of images, which is closely related to per-pixel material parameter estimation. IntrinsicAnything~\cite{chen2024intrinsicanything}
decomposes images into diffuse and specular components, and leverages these components as priors using physically-based inverse
rendering to extract material maps. MaterialFusion~\cite{litman2025materialfusion} introduces
a 2D diffusion model prior to help estimate material parameters in an multi-view reconstruction pipeline.
RGB$\leftrightarrow$X~\cite{zeng2024rgb} uses finetuned diffusion models for both intrinsic decomposition of images into G-buffers and the neural rendering of images from G-buffers. DiffusionRenderer~\cite{DiffusionRenderer} extends RGB$\leftrightarrow$X
to videos, and also supports relighting. NeuralGaffer~\cite{jin2024neural_gaffer} and DiLightNet~\cite{zeng2024dilightnet} leverage 
diffusion models for relighting single views. 
IllumiNerf~\cite{zhao2024illuminerf} relights each view in a multi-view dataset, then reconstructs a NeRF model with these relit images.
IntrinsiX~\cite{kocsis2025intrinsix} combines intrinsic predictions for PBR G-buffers for a single view 
from text (using image diffusion models) with a rendering loss. 
MCMat~\cite{zhu2024mcmat} leverages Diffusion Transformers (DiT) to extract multi-view images of PBR material maps, combined with a second DiT to enhance details in UV space.

VideoMat~\cite{munkberg2025videomat}, the closest related work to ours, generates materials for 3D shapes by first generating an RGB video of a textured and lit 3D model conditioned on untextured geometry, and then extracting the material parameters by combining video intrinsic decomposition and differentiable rendering to project the material parameters into texture space.

\paragraph{Joint generative modeling} approaches enable diffusion models to predict multiple modalities. Matrix3D~\cite{lu2025matrix3d} 
predicts pose estimation, depth, and novel view synthesis using a single DiT~\cite{peebles2022dit} model. 
VideoJAM~\cite{hila2025videojam} extends this by predicting both generated pixels and their corresponding motion from a 
single DiT. UniRelight~\cite{he2025unirelight} leverages this approach to jointly predict relit and base color videos.

\section{Method}

Our pipeline, as shown in \cref{fig:system}, uses join generative modeling with video diffusion models 
to produce PBR material textures. We assume a given 3D model with a valid texture parameterization (but no textures) as input.
We generate multiple views of material intrinsics: G-buffers of \emph{base color}, \emph{roughness}, \emph{metallicity}, and \emph{height}
values, conditioned on corresponding input geometry (views of surface normals and world space positions). 
Finally, we project the intrinsic views into texture space to obtain standard PBR materials directly compatible 
with common 3D authoring tools: Blender, Unreal Engine, etc. Below, we describe each step in detail.

\subsection{Base Video Model Architecture}
\label{sec:video_model1}
In a first step, we produce a synthetic dataset consisting of multiple views of material intrinsics, 
conditioned on geometry (surface normals and world space positions for each view) 
and a text prompt describing each object's material. We use this data to finetune a recent Diffusion Transformer (DiT) video model, 
Cosmos~\cite{cosmos_short}, for this task. We use the \texttt{Cosmos-1.0-Diffusion-7BVideo2World}\footnote{https://github.com/NVIDIA/Cosmos} model which is trained in a latent space with 8$\times$ compression in the spatial and temporal domain. This model supports text- and image guided video generation at a resolution of 1280$\times$704 pixels and 121 frames.
The base model leverages the pretrained \texttt{Cosmos-1.0-Tokenizer-CV8x8x8}
to encode and decode RGB videos to and from latent space. We directly use this encoder
to encode our input conditions, but introduce a novel tokenizer to jointly compress the material modalities.

\subsection{Per-frame encoding}
The temporal compression of the Cosmos Tokenizer~\cite{cosmos_short} encoder, $\vaeEncoder$,  
introduces some motion blur in the reconstructed frames. To 
avoid this, we use the image (keyframe) mode, which encodes each frame individually, 
so our latents only have 8$\times$ spatial compression. In other words, we opted for encoded videos with fewer, but higher quality, frames. 
Specifically, we encode an input video with $F$ frames, $C=6$ channels, and spatial 
resolution $H \times W$, represented a tensor $F \times C \times H \times W$ into a latent space with dimensions
$F \times 16 \times H/8 \times W/8$.
Furthermore, a typical video VAE is trained on mostly coherent videos with limited motion between frames; we encode each frame individually, so we do not need to adhere to this constraint, and we pick a random camera view for each frame in each training example. 

\subsection{Joint generative modeling}

Our goal is to jointly predict spatially varying \emph{base color}, \emph{roughness}, \emph{metallicity}, and \emph{height} material parameters,
conditioned on positions and normals of the input 3D model. Unlike recent neural inverse renderers~\cite{zeng2024rgb, DiffusionRenderer}
which predict one modality at a time in separate inference passes, we instead follow the approach in recent joint generative modeling
approaches~\cite{lu2025matrix3d,hila2025videojam,he2025unirelight} to predict multiple modalities in a single inference pass.

UniRelight~\cite{he2025unirelight} jointly predicts a relit video and base color by concatenating latents 
for the two modalities along the \emph{frame} dimension. In contrast, we leverage a custom variational auto-encoder (VAE), which encodes all material modalities into a shared latent space.  
This way we obtain a VAE specialized for the material domain, while avoiding the  increased token length from frame concatenation.

Recent work in neural texture compression~\cite{vaidyanathan2023ntc} shows that multiple material maps can be efficiently compressed together as the maps often contain correlated details. We explore if this is also applicable to VAEs.
More precisely, we leverage the pretrained Cosmos Tokenizer~\cite{cosmos_short}, which bidirectionally maps between RGB images ($3 \times H \times W$ tensors) and a latent representation using an encoder-decoder pair, $(\vaeEncoder, \vaeDecoder)$. We use the image (keyframe) VAE encoding mode.
We make minimal changes to the base model, only updating the channel count for the input layer of the encoder and output layer of the decoder, and perform finetuning to create our $\vaepbr$ which maps a $6 \times H \times W$ tensor (\emph{base color}, \emph{roughness}, \emph{metallicity}, and \emph{height}) to latents of the same size as the basemodel.
We leverage the latent space produced by $\mathrm{VAE}_\mathrm{pbr}$ in the diffusion process to jointly predict frames of material parameters for all views, as is shown in \cref{fig:system}.



\subsection{Finetuning}

We finetune the embedding layer (extended from the base model to support our input conditions) and all DiT layers for 20k iterations on 64 A100 GPUs. 

Given an input video $\videoInput$ consisting of normals and world space positions for N views
of a 3D model, our goal is to train a model $\diffusionModelFn$ that {\it jointly} denoises views of PBR material maps conditioned on $\videoInput$.

The model comprises a VAE encoder-decoder pair (the Cosmos Tokenizer), $(\vaeEncoder, \vaeDecoder)$, and a transformer-based denoising function, $\diffusionModelFn$.
We use the encoder $\vaeEncoder$ to encode the input conditions, $\videoInput$, 
into a latent tensor, $\textbf{z}^{\videoInput}$.

Our model is finetuned on a synthetic video dataset. 
Each data sample consists of 16 random object-centric camera views of 
a 3D objects. Each view includes G-buffers of normals, depth, base color, roughness, metallicity, height values, 
and the camera pose. We use the depth and camera pose to compute a world space position buffer in the data loader.  

The target latent variable, $\textbf{z}_0^{\matmap}$, for this dataset is constructed by encoding the base color, roughness, 
metallicity, and height values \emph{jointly} (six channels) using our $\vaepbr$ encoder,  $\mathcal{E}_{\mathrm{pbr}}$.
Noise, $\diffusionNoise$, is introduced to our latent, $\textbf{z}_0^{\matmap}$, representing the material parameters, to produce $\textbf{z}_\tau^{\matmap}$. 
The model parameters, $\theta$, of the diffusion model, $\diffusionModelFn$, are optimized by minimizing the objective function: 
\begin{eqnarray}
\hat{\textbf{z}}^{\matmap}(\theta) &=&  \diffusionModelFn ([\textbf{z}_\tau^{\matmap},\textbf{z}^{\videoInput}]; \typeEmb, \tau) \label{eq:denoiser} \nonumber \\
\mathcal{L}(\theta) &=& \mathbb{E}_{\textbf{z}_0^{\matmap}\sim\dataDistribution,\diffusionNoise\sim \mathcal{N} (0,\sigma^2 I)} \left\| \hat{\textbf{z}}^{\matmap}(\theta) - \textbf{z}_0^{\matmap} \right\|_2^2
\nonumber \label{eq:objective},
\end{eqnarray}
where $[\cdot]$ denotes concatenation in the channel dimension and $\typeEmb$ is the encoded text prompt (encoded using T5-XXL~\cite{raffel2023t5}).
We increase the input feature count of the input embedding layer of $\diffusionModelFn$ to account for our additional input conditions, $\textbf{z}^{\videoInput}$.

We use the denoising score matching loss from Cosmos~\cite{cosmos_short} unmodified, applied to
the predicted latent $\hat{\textbf{z}}^{\matmap}(\theta)$ and the corresponding 
target latent $\textbf{z}_0^{\matmap}$.

\paragraph{Dataset}

Our dataset consists of 60k videos of object-centric renderings of 3D models from 
Objaverse~\cite{objaverse}, BlenderVault~\cite{litman2025materialfusion},
ABO~\cite{collins2022abo}, and HSSD~\cite{khanna2023hssd}. 
For each object, we render a video with 16 frames at a resolution of 1024$\times$1024, using a path tracer with three bounces and Blender AgX tonemapping. We use black background, and for each frame the view is randomized. For lighting, we use the ``BoilerRoom'' light probe from Poly Haven~\cite{polyhaven}, providing constant, neutral lighting for all objects. We only use the shaded video to automatically generate captions using Qwen2.5-VL-7B~\cite{qwen25}, and want to avoid prompt noise due to variation in lighting. 
We also render intrinsic maps (normals, world space positions, base color, roughness, metallicity, height). 
The height map is not available for most assets, and we reconstruct it from the normal map using standard conversion tools when available.
We augment the dataset by randomly reversing the video, and randomly offsetting the video start frame in each training iteration.

We additionally use this dataset to finetune our VAE. To avoid biasing too heavily towards objects on a black background, we additionally use the MatSynth~\cite{vecchio2024matsynth} training set (which contains all material channels expected by our model) and randomly pick samples using a $60/40$ distribution.

\subsection{Transfer multi-view intrinsics to texture space}

At inference, we generate 16 views of the material intrinsics from known cameras. 
To extract material maps in texture space, which is the standard format in 
content creation tools, we project the intrinsic 
views into texture space using a splatting approach. 
We upscale the generated views to a resolution of 16k $\times$ 16k pixels and render a texture coordinate guide using the 3D asset, assuming a known, non-overlapping UV-mapping. Each pixel is splatted to the corresponding (nearest neighbor) texel of a 2048 $\times$ 2048 texture with a weight inversely proportional to the screen space texture derivatives~\cite{heckbert1989texmap}
to suppress areas with high perspective distortion. More formally, given texture coordinates $(u,v)$ for a pixel $(x,y)$ the weight is computed as:
\begin{equation*}
    w = \frac{1}{\max\left(\left|\left(\partial u / \partial x, \partial v / \partial x\right)\right|, \left|\left(\partial u / \partial y, \partial v / \partial y\right)\right| \right)}.
\end{equation*}
We normalize the final texture by the total weight per texel, and perform basic inpainting~\cite{Telea2004} for all texels with zero weight to reduce texture atlas seams.

\subsection{Image-conditioned video generation}
\label{sec:image_conditioned}

While our primary focus is on material generation from text, our pipeline can be straightforwardly extended to add image conditioning. We adopt an approach similar to Gen3C~\cite{ren2025gen3c}
where the video model is conditioned on a single shaded input image, which is warped (using a provided depth buffer) according to the known camera matrix and intrinsics for each view. As in our text-to-video setting, we condition the video model on normals and world space positions, and simply concatenate the warped shaded images to the condition, $\videoInput$, with no further changes needed.  We argue that both forms of conditioning are useful in production workflows, as reference images are not always available.

\section{Results}

\figMainQualityResults

We evaluate our method against VideoMat~\cite{munkberg2025videomat}, 
a material generation method which also leverages DiT video models. To make comparisons easier,
we use the same pretrained base video model as VideoMat throughout this paper. However, we note that our method
will directly benefit from a stronger base model.
As a representative example of recent multi-view diffusion material generation techniques, 
we chose Hunyuan3D-Paint 2.1~\cite{he2025materialmvp} and MVPainter~\cite{shao2025mvpainter}, which both are image-guided material generation methods. 
We also note that image-conditioned models can be repurposed for text conditioning by an additional text-to-image step. Therefore, we also constructed a text-guided version of Hunyuan3D-Paint and MVPainter by first generating an image from the text prompt using a depth-guided Flux ControlNet~\cite{fluxcontrol}, and feeding it as an image condition into Hunyuan3D-Paint and MVPainter.
We include TRELLIS.2~\cite{xiang2025trellis2} as an image-conditioned method generating materials directly in 3D space
(using their PBR texture generation mode with known geometry).
There is a plethora of recent multi-view diffusion methods, 
and we refer the reader to the concurrent commercial approach Seed3D~\cite{seed3d} for extensive comparisons; however, their model has not been released.

\subsection{Quantitative evaluation}
\label{sec:quantiative_eval}
We report quantitative results on material generation in \cref{tab:matgen}. There are no established metrics for text-to-material generation quality; therefore, we repurposed image-based metrics as follows. We choose 32 test assets, render images of the assets with their original material assignments, and annotate them with text captions using Qwen2.5-VL-7B (similarly to training assets). We generate materials using the estimated text prompts using all methods for these 32 test models. For the image-guided methods, we used a reference rendering per assets with original material assignments as guidance. Finally, we render four views, each in four different lighting conditions (four different HDR probes), resulting in 512 images each for both original and generated materials. The resulting renderings can be compared using image metrics. Note that this comparison goes through a "text bottleneck": the achievable similarity of the corresponding image pairs is limited by this, and the resulting numbers are not directly comparable to image-conditioned models.

We report CLIP-based Fr\'echet Inception Distance
(CLIP-FID)~\cite{Kynkaanniemi2022}, Learned Perceptual Image Patch Similarity (LPIPS)~\cite{zhang2018perceptual}, and CLIP Maximum-Mean Discrepancy (CMMD)~\cite{jayasumana2024cmmd}.
We refer to VideoMat~\cite{munkberg2025videomat}
for additional comparisons against Paint-it~\cite{youwang2024paintit}, DreamMat~\cite{Zhang2024dreammat}, and Make-it-Real~\cite{fang2024makeitreal}.

Among the text-guided variants, our method has the best scores. We encourage the reader to closely inspect the visual results,
where we argue that VideoMatGen produces sharper results, with more definition, particularly in the roughness 
and metallicity maps; furthermore, VideoMatGen is the only method producing a height map. 
For completeness, we also report image-guided results where we have extended our model to accept both a prompt and a single image as guides. While not our primary design goal, we note that our method still performs competitively compared to the state of the art.

\begin{table}
  \caption{Quantitative metrics for material generation. The mode column indicates
  if the method is image or text guided.}
  \label{tab:matgen}
  \centering
  \setlength{\tabcolsep}{2pt} 
  \renewcommand{\arraystretch}{0.9} 
  \begin{small}
  \begin{tabular}{@{}lccccc@{}}
    \toprule
    Method & Mode & CLIP-FID~($\downarrow$)  & CMMD~($\downarrow$) & LPIPS~($\downarrow$) \\
    \midrule
    TRELLIS.2~\cite{xiang2025trellis2}       & image & \sota{3.913} & \subsota{0.030} & \sota{0.101} \\
    Hunyuan3D~2.1~\cite{he2025materialmvp}   & image & 4.197 & 0.039 & \subsota{0.102} \\
    MVPainter~\cite{shao2025mvpainter}       & image & 6.583 & 0.112 & 0.136 \\
    VideoMatGen (ours)                       & image & \subsota{4.032} & \sota{0.028} & 0.109 \\
    \midrule
    Hunyuan3D~2.1~\cite{he2025materialmvp}   & text & 6.694 & \subsota{0.046} & 0.137 \\    
    MVPainter~\cite{shao2025mvpainter}       & text & 7.313 & 0.096 & 0.149 \\
    VideoMat~\cite{munkberg2025videomat}     & text & \subsota{5.640} & 0.070 & \subsota{0.130} \\
    VideoMatGen (ours)                       & text & \sota{5.575} & \sota{0.035} & \sota{0.124} \\
    \bottomrule
  \end{tabular}
  \end{small}
\end{table}

\subsection{Qualitative evaluation}

\figBumpmap

\figSeed

\figRelighting

In \cref{fig:main_quality_results}, we show visual comparisons against VideoMat~\cite{munkberg2025videomat} and two variants of Hunyuan3D-Paint~\cite{he2025materialmvp} (image- and text-guided). Overall, the visual results are compelling for all methods, 
but we notice that the strong prior from the video model helps us generate fine scale detail,
and more interesting spatial texture variations,  which are coherent across the different material maps thanks to our joint modeling. 
Unlike the competing methods, we predict a height map, which improves fine scale material detail, as highlighted in \cref{fig:bump}. 
We can create subtle material variations from a single prompt by changing the seed, as shown in \cref{fig:teaser,fig:seed}. This 
can be a helpful artistic tool in creating unique instances for the same base geometry in larger scenes.
In \cref{fig:relighting} we show our generated materials rendered with three different lighting conditions. 
Finally, our image conditioned pipeline generates materials which are visually more similar to the test set examples, as shown in \cref{fig:imgcond}. 

\figImgCond

\subsection{Evaluation of joint prediction}

\paragraph{VAE Quality}
We compare our finetuned VAE with the Cosmos Tokenizer (Cosmos-0.1-Tokenizer-CV8x8x8, applied to single frames). Quality is evaluated using image metrics after encoding and decoding each image. For the Cosmos Tokenizer, base color and HRM (a packed 3-triplet with height, roughness, metallicity) are encoded separately as RGB images, while we jointly encode all six channels using 
$\mathrm{VAE}_\mathrm{pbr}$.
Our test set consists of 4 views of each of our 32 test assets (128 samples), with their original material assignments. For each view, we render material intrinsics maps for base color, height, roughness and 
  metallicity. Additionally, we use the material textures from the MatSynth~\cite{vecchio2024matsynth} test set (89 samples). 
As shown in \cref{tab:vae}, when applying our VAE on material maps, 
we have similar quality as the Cosmos Tokenizer, while achieving $2\times$ higher compression rate in latent space.

\begin{table}
  \caption{VAE finetuning evaluation on material maps from our test set and the MatSynth test set.
  We report PSNR~(dB) and LPIPS scores for base color and only PSNR~(dB) scores for HRM (height, roughness, metallicity), as perceptual metrics are not applicable. The $\mathrm{VAE}_\mathrm{pbr}$ 
  offers 2$\times$ additional compression but has similar visual quality as the Cosmos Tokenizer.}
  \label{tab:vae}
  \centering
  \begin{tabular}{@{}lccc@{}}
    \toprule
    & \multicolumn{2}{c}{Base color} & HRM \\
    Method & PSNR~($\uparrow$) & LPIPS~($\downarrow$) & PSNR~($\uparrow$) \\
    \midrule
    Cosmos Tokenizer & \best{38.8} & 0.046 & \best{35.1} \\
    $\mathrm{VAE}_\mathrm{pbr}$ & 38.3 & \best{0.043} & 33.8 \\
    \bottomrule
  \end{tabular}
\end{table}

\section{Limitations and Future Work}

Our method is currently unoptimized and made with no regards to runtime performance. Inference is costly,  approximately 2-3 minutes for a single asset on 8$\times$A100 GPUs. We see large potential for optimizing inference with recent video model acceleration and distillation techniques.

Our image (keyframe) VAE approach allows for random camera views at inference time, but we still note that the video model produces best results with a reasonably coherent camera trajectory. Incoherent view-patterns can lead to ghosting or blurring due to misaligned details, and for this reason we chose a object-centric $360^\circ$ camera orbit during inference. In future work we hope consistency can be improved by better image guides or 3D positional encoding. 

While not the primary focus of this paper, our texture baking step is a relatively simple projection of the generated video frames. Recent works have shown that quality can be improved by applying image diffusion models in texture space~\cite{seed3d,zhu2024mcmat} to in-paint or sharpen details. 

We would also like to upgrade our base model to a more recent video diffusion model. In this paper, we deliberately used Cosmos-1.0 for fair comparison with VideoMat, but more recent models can likely improve quality.

\section{Conclusion}
We present a video diffusion method for joint prediction of material parameters for 3D shapes. We also show the benefits of our new joint material modeling VAE.
Our model produces high-quality PBR materials with coherent detail between the material channels and meaningful correlation to geometry parts, and outperforms previous text-to-material approaches.
We believe that our text-based material generation can be a useful tool for artists 
to quickly prototype materials for large sets of 3D objects. Unique material variation for instances
of the same geometry can be obtained by simply changing the seed of the noise passed to the diffusion process.

{
   \small
   \bibliographystyle{ieeenat_fullname}
   \bibliography{main}
}

\clearpage
\setcounter{page}{1}

\maketitlesupplementary

\section{Supplemental material}

\figTrellis

\subsection{Results on AI generated geometry} 

To illustrate that our method can complement an image$\rightarrow$3D pipeline with high quality PBR material generation,
we create 3D objects using TRELLIS.2~\cite{xiang2025trellis2} (conditioned on images from our test set), and use the unmodified GLTF meshes in our pipeline. There are no significant mapping issues as shown in \cref{fig:trellis}. Our method does not rely on the UV mapping except in the final projection to UVs, which is optional.

\subsection{Frame concatenation vs. Compressed VAE}

\figFrameConcatSup

\paragraph{Implementation}
Following UniRelight~\cite{he2025unirelight}, in our experiment with frame concatenation, we concatenate two encoded latent 
videos along the \emph{frame} dimension of the input tensor of the DiT. The first video represents
sixteen views of base color, the second video represents the corresponding
sixteen views of height, roughness, and metallicity, packed into RGB images. 
Both videos are encoded with the Cosmos Tokenizer, $\vaeEncoder$, using the image 
(keyframe) mode. To distinguish the two video segments, 
we leverage the \emph{view} encoding used in the multi-view post training example of Cosmos~\cite{cosmos_short}. Note that frame concatenation 
doubles the number of tokens (and inference time), which limits us
to train with examples with 16 frames in a resolution of 768$\times$768 pixels.
In contrast, our proposed architecture using $\mathrm{VAE}_\mathrm{pbr}$ is more memory-efficient,
and we can train with videos with a spatial resolution of 1024$\times$1024 pixels.
We implemented both variants of joint prediction for our material prediction task to evaluate their quality. For positional encoding in the frame concatenation version, we leverage the \emph{view} encoding used in the multi-view post training example of Cosmos~\cite{cosmos_short}, to distinguish the two video segments.

\paragraph{Results}
\label{sec:frameconcat}
In \cref{fig:frame_concat_sup} we show examples of the generated materials for a frame-concatenation variant (which doubles the number of tokens and inference time) vs. our proposed $\mathrm{VAE}_\mathrm{pbr}$. In \cref{tab:fcat} we present metrics comparing the two series, 
using the same evaluation protocol from \cref{sec:quantiative_eval}.

\begin{table}[h]
  \caption{Quantitative metrics for text-to-material generation. We compare two variants of joint prediction, frame concatenation (FCat) which doubles the number of video frames/latent tokens, 
  and our version with $\mathrm{VAE}_\mathrm{pbr}$.}
  \label{tab:fcat}
  \centering
  \setlength{\tabcolsep}{2pt} 
  \renewcommand{\arraystretch}{0.9} 
  \begin{small}
  \begin{tabular}{@{}lcccc@{}}
    \toprule
    Method & CLIP-FID~($\downarrow$)  & CMMD~($\downarrow$) & LPIPS~($\downarrow$) \\
    \midrule
    VideoMatGen ($\mathrm{VAE}_\mathrm{pbr}$) & \best{5.638} & \best{0.035} & \best{0.126} \\
    VideoMatGen (FCat)                        & 5.712 & 0.039 & 0.129 \\
    
    \bottomrule
  \end{tabular}
  \end{small}
\end{table}

\subsection{Text-alignment metric} 

Our prompts (generated by Qwen2.5-VL-7B) exceeds CLIP's limitation of 77 tokens. Therefore, we report
BLIP scores~\cite{li2022blip} below (Using the \texttt{Salesforce/blip-itm-base-coco} model). We use 16 views of each of our 32 test examples. The score is a binary classification probability ($\in [0,1]$). Here, the reference means views of the test set materials with captions from Qwen2.5-VL-7B.
\\
\\
\begin{tabular}{@{}lc@{}}
  \toprule
  Method & BLIP Mean Image-Text Matching scores\\
  \midrule
  VideoMat    & 0.9015 $\pm$~0.2474 \\
  VideoMatGen & 0.9271 $\pm$~0.2025 \\
  Reference   & 0.9339 $\pm$~0.1944 \\
  \bottomrule
\end{tabular}
\\

\subsection{Rendering loss experiment}
We experimented with including a \emph{rendering loss} when finetuning the diffusion model, where we leverage a \emph{differentiable} version of split sum shading~\cite{Munkberg_2022_CVPR}. 
In each training iteration, we load a random HDR probe, and evaluate
image-based shading for each view using a Lambertian term and a Cook-Torrance microfacet specular shading approximated by split sum. However, we did not see improved results compared to only training with the denoising score matching loss. 
In our setting, given that our predicted latent includes all material modalities, a rendering loss is only 
representing a different weighting factor of each modality. This is in contrast to methods
which predict each material modality in separate networks~\cite{kocsis2025intrinsix}, where a rendering loss is critical to align the modalities. 
Note also that an image space loss term requires more memory during training, as it is computed \emph{after} VAE decoding, and hence, requires backpropagation through the VAE decoder, $\vaeDecoder_\mathrm{pbr}$, to update the DiT weights. 

\subsection{Prompts}

In this subsection, we include the text prompts for the examples shown in the main paper. 
For the full set of 32 prompts used in our test set, please refer to the 
image viewer. 

\textbf{Diver}: "A vintage diving helmet with a worn, copper-colored finish rotates against a black background. The helmet features multiple circular windows for visibility, with one prominently positioned on the front. It has a sturdy, metallic construction with visible bolts and rivets, giving it a rugged and industrial appearance. The helmet's design includes a curved neck guard and a handle on top, suggesting it was used for deep-sea exploration. The surface shows signs of age and wear, with patches of rust and discoloration, adding to its historical charm. A close-up shot from various angles highlights the intricate details and craftsmanship of this classic diving gear."

\textbf{Robot}: "A quirky, retro-futuristic robot with a boxy head and a small screen displaying green code. It has two large, striped arms and legs, each ending in simple, rounded feet. The robot's body is adorned with various mechanical components, including a circular antenna on top and a small, round sensor on one side. It moves slowly, swaying slightly as it walks, giving off a playful and endearing vibe. The background is plain black, emphasizing the robot's unique design and movements. A medium shot capturing the robot's full body as it navigates through space."

\textbf{Shed}: "A small wooden house model rotates against a bright background. Light, bright, vivid colors. The structure is made of weathered wooden planks, with a sloped roof covered in corrugated metal sheets. The house features two small windows, one on each side, and a small door with a window above it. A small awning with a striped pattern hangs over the entrance. The model is detailed with visible joints and supports, giving it a rustic and handmade appearance. The camera pans around the house, showcasing its various angles and features."

\textbf{Lantern}: "A vintage-style lantern rotates against a black background. The lantern is made of metal with a weathered, rustic appearance, featuring a white glass globe protected by a wire cage. The handle is coiled and attached to the side, and the base has a textured, ribbed design. The lantern's intricate details and sturdy construction suggest it is designed for practical use, possibly for camping or outdoor activities. A medium shot captures the lantern from various angles as it spins slowly."

\textbf{Motorcycle}: "A sleek black motorcycle rotates against a bright background, Light, bright, vivid colors, showcasing its intricate design and polished chrome details. The bike features a classic retro aesthetic with a rounded front fender, a prominent headlight, and a comfortable-looking black seat. The handlebars are equipped with round mirrors, and the engine is exposed, revealing a robust and powerful build. The wheels have spoked rims, adding to its vintage charm. The motorcycle's design is highlighted from multiple angles as it spins, emphasizing its elegant lines and craftsmanship. A close-up shot from various perspectives."

\end{document}